%% file: main.tex
\documentclass[conference]{IEEEtran}

\usepackage{times}
\usepackage{amsmath,amsfonts}
\usepackage{stix}

\usepackage{array}

\usepackage[numbers]{natbib}
\usepackage{multicol}
\usepackage[bookmarks=true]{hyperref}

\usepackage[table]{xcolor}
\usepackage{booktabs}
\let\labelindent\relax
\usepackage{enumitem}
\usepackage{multirow}
\usepackage{graphicx}
\renewcommand{\baselinestretch}{0.982}

\usepackage[font=scriptsize]{caption}

\usepackage{float}
\usepackage{comment}

\usepackage{algorithm}
\usepackage{algpseudocode}
\usepackage{etoolbox}

\usepackage{subcaption}

\newcommand{\teaser}{%
  \vspace{-0mm}
  \begin{center}
    \includegraphics[width=0.9999\textwidth]{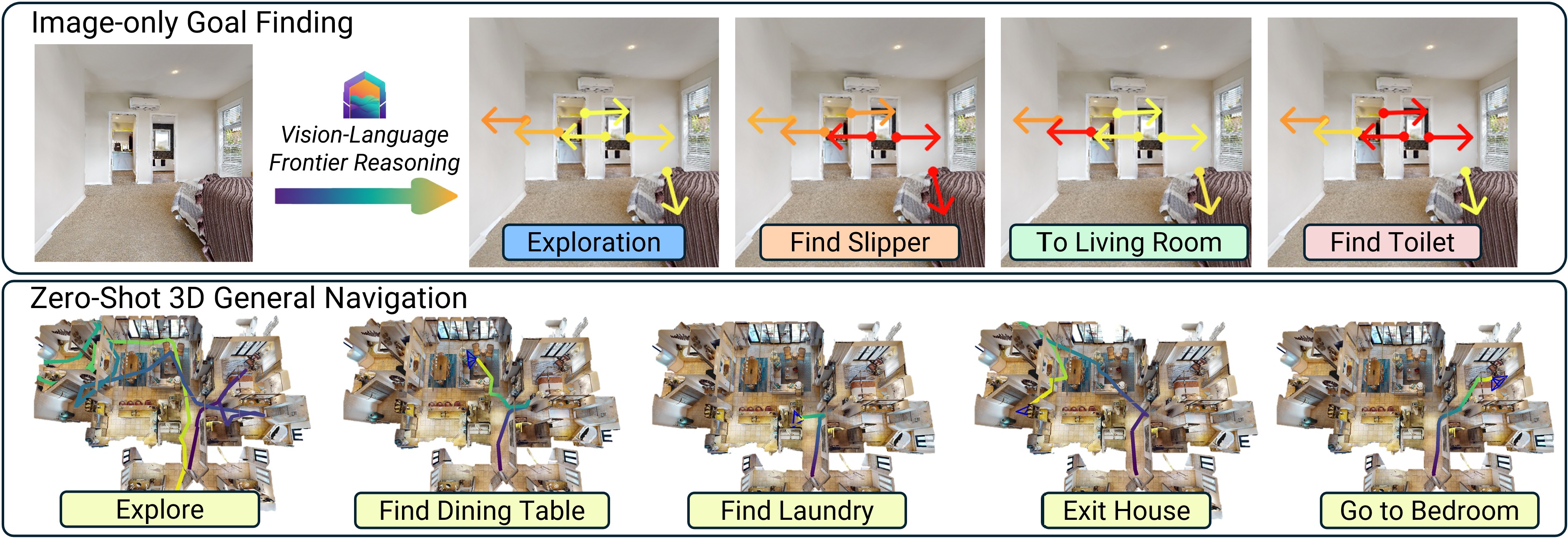}
    \vspace{-5mm}
    \captionof{figure}{
    We present \textbf{OpenFrontier}.
    \textbf{Top:} OpenFrontier detects and semantically evaluates visual frontiers directly in the image space, enabling language-conditioned goal reasoning without dense 3D reconstruction. Arrow color encodes goal relevance, from \textcolor{red}{red} (high) to \textcolor[rgb]{0.75,0.6,0}{yellow} (low).
    \textbf{Bottom:} Visual frontiers from individual observations are sequentially grounded into the 3D metric space, driving long-horizon, natural-language–conditioned navigation across diverse goals in a fully zero-shot manner, without task-specific fine-tuning.
    }
    \label{fig:teaser}
  \end{center}
  \vspace{-2mm}
}

\makeatletter
\apptocmd{\@maketitle}{\teaser}{}{\GenericWarning{}{Failed to patch \string\@maketitle}}
\makeatother

\begin{document}

\title{OpenFrontier: General Navigation with Visual-Language Grounded Frontiers}

\author{
Esteban Padilla-Cerdio\textsuperscript{*,1},
Boyang Sun\textsuperscript{*,1},
Marc Pollefeys\textsuperscript{1,2},
Hermann Blum\textsuperscript{3}\\
\textsuperscript{1}ETH Zurich \quad
\textsuperscript{2}Microsoft Spatial AI Lab \quad
\textsuperscript{3}University of Bonn\\
\textsuperscript{*}Equal Contribution\\
{\small\href{https://boysun045.github.io/OpenFrontier-Project/}{\textcolor{blue}{\underline{https://boysun045.github.io/OpenFrontier-Project/}}}}
}

\maketitle
\begin{abstract}
Open-world navigation requires robots to make decisions in complex everyday environments while adapting to flexible task requirements. Conventional navigation approaches often rely on dense 3D reconstruction and hand-crafted goal metrics, which limits their generalization across tasks and environments. Recent advances in vision--language navigation (VLN) and vision--language--action (VLA) models enable end-to-end policies conditioned on natural language, but typically require interactive training, large-scale data collection, or task-specific fine-tuning with a mobile agent. We formulate navigation as a sparse subgoal identification and reaching problem and observe that providing visual anchoring targets for high-level semantic priors enables highly efficient goal-conditioned navigation. Based on this insight, we select visual frontiers as semantic anchors and propose \emph{OpenFrontier}, a navigation framework that requires no task-specific training or fine-tuning and seamlessly integrates diverse vision--language prior models. OpenFrontier enables efficient navigation with a lightweight system design, without dense 3D semantic mapping, task-specific policy training, or model fine-tuning. We evaluate OpenFrontier across multiple navigation benchmarks and demonstrate strong zero-shot performance, as well as effective real-world deployment on a mobile robot.
\end{abstract}

\IEEEpeerreviewmaketitle
\setcounter{figure}{1}

\input{intro}

\input{related}
\input{method}
\input{experiment}

\input{conclusion}


\section*{Acknowledgement}
This work was partially supported by the Lamarr Institute and by the Robotics Institute Germany.

\bibliographystyle{plainnat}
\bibliography{references}

\clearpage
\appendix

\addcontentsline{toc}{section}{Supplementary Material}

\setcounter{figure}{0}
\setcounter{table}{0}
\renewcommand{\thefigure}{S\arabic{figure}}
\renewcommand{\thetable}{S\arabic{table}}
\renewcommand{\thesection}{S\arabic{section}}

\input{sup_tech}
\input{sup_qualitative}

\end{document}

%% file: intro.tex
\section{Introduction}
\label{sec:intro}

\begin{figure*}[t]
\centering
\includegraphics[width=.99\linewidth]{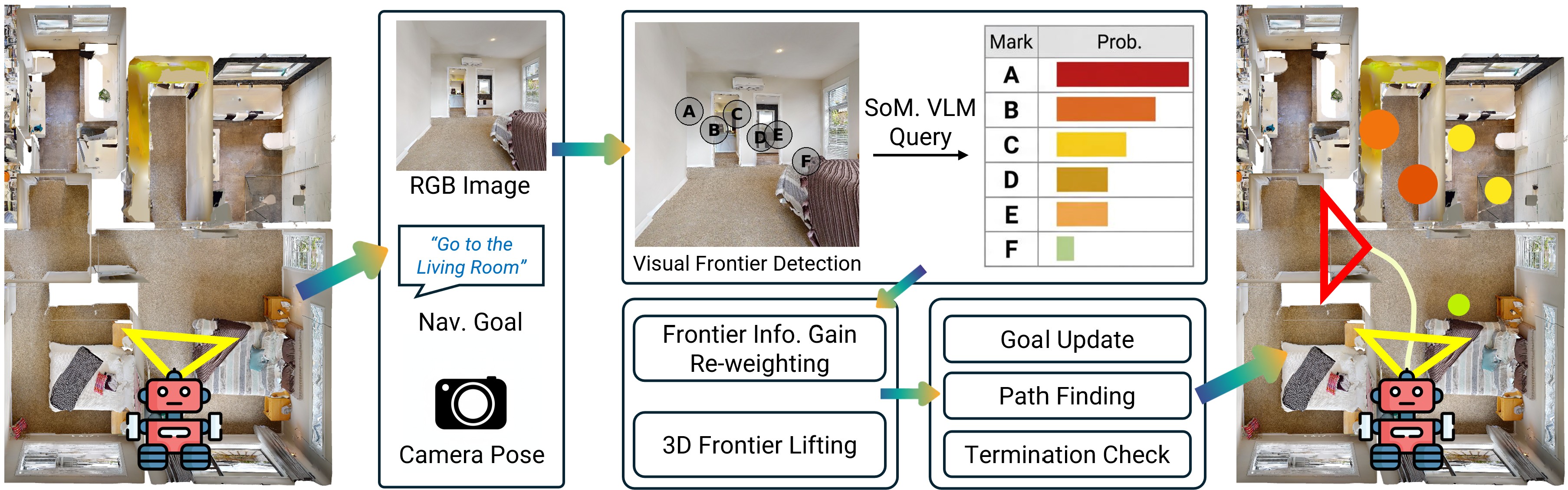}
\caption{\textbf{System Overview}. 
Given a posed RGB observation and a natural-language navigation goal, OpenFrontier detects visual frontiers in the image and directly queries a vision--language model to evaluate their relevance (probability to the target) using in-image context.
The resulting frontiers are then lifted into the 3D metric space with the updated information gain as goal-conditioned candidates and globally managed to update navigation targets, perform path planning, and determine termination.}
\label{fig:overview}
\vspace{-1mm}
\end{figure*}

Navigation in open-world environments requires robots to reason over both high-level semantics and low-level geometry while operating under partial observability. Classical object-goal navigation methods typically rely on dense 3D reconstruction followed by object detection and localization in a global map~\cite{zhang20233d,blomqvist2020go,hughes2024foundations}, which demands accurate mapping and struggles with cluttered scenes or small, ambiguous objects. Previous learning-based approaches attempt to jointly solve high-level decision making and low-level control through reinforcement learning~\cite{chaplot2020object,chaplot2021seal,yu2023frontier,wang2025matters,zeng2025poliformer}, but are commonly limited to closed-set object categories and show poor generalization beyond training distributions~\cite{gervet2023navigating}. More recently, large language models (LLMs) and vision-language models (VLMs) have been explored as sources of semantic priors for navigation, either via end-to-end vision-language-action training~\cite{xue2025omninav,zhang2024uni,cheng2024navila} or by directly prompting models for action cues~\cite{melnik2024cognitive,goetting2024end}. However, such approaches often require large-scale interactive training, suffer from real-time constraints, or face fundamental difficulties in grounding high-level semantic reasoning into metric navigation decisions.

While stronger semantic priors, larger datasets, and more versatile policies continue to advance language-conditioned navigation, an equally important challenge lies in identifying an \emph{effective interface} that grounds semantic reasoning into physically meaningful navigation decisions. In this work, we revisit the general robot navigation task and propose \textbf{OpenFrontier}, a zero-shot navigation framework that uses visual navigation frontiers as sparse, interpretable, and physically grounded semantic anchors. Frontiers correspond to navigable regions in metric space and naturally encode exploration while remaining directly actionable, making them an ideal substrate for language-conditioned reasoning. OpenFrontier presents detected frontiers to a VLM using a set-of-marks formulation \cite{yang2023set} over the individual RGB observation, enabling the model to assign semantic relevance and priority to each candidate. It maintains a global updating mechanism to keep track of sparse frontiers. The robot then navigates by iteratively clearing frontiers according to these grounded priors, balancing exploration and exploitation while continuously assessing goal satisfaction. 

OpenFrontier requires no dense semantic mapping, no task-specific policy training, and no fine-tuning, is agnostic to the choice of VLM for reasoning, and transfers seamlessly to unseen environments, open-set goals and real-world scenarios. We evaluate OpenFrontier across multiple indoor navigation benchmarks, including both closed-set and open-set object-goal navigation tasks, and demonstrate strong zero-shot generalization compared to prior methods. Despite its simplicity, OpenFrontier shows competitive performance against recent baselines while requiring the minimum mapping or fine-tuning effort. Finally, we deploy our system on a mobile legged robot and demonstrate robust real-world navigation in a large indoor environment. As a summary, our main contributions are:

\begin{itemize}[leftmargin=*]
    \item We propose \textbf{OpenFrontier}, a navigation framework that uses \emph{visual navigation frontiers} as an interface for grounding vision--language priors into actionable navigation goals for language-conditioned object-goal navigation.
    
    \item We introduce a visual-target reasoning formulation that evaluates candidate frontiers using vision--language models and integrates semantic relevance with exploration-driven information gain, without requiring dense 3D semantic maps or task-specific policy training.
    
    \item We demonstrate strong zero-shot performance across multiple navigation benchmarks and validate the approach in real-world robotic deployment.
\end{itemize}

%% file: related.tex
\section{Related Work}

\subsection{Map-based Exploration and Navigation}

Building explicit maps for navigation is one of the most classical paradigms in mobile robotics~\cite{10475904,batra2020objectnav}.
A wide range of map representations have been investigated.
Among them, frontier-based exploration has long been a core approach, where frontiers represent the boundary between known and unknown space and serve as natural candidates for exploration-driven motion~\cite{yamauchi1997frontier}.
Classical methods extract frontiers from volumetric occupancy maps and select navigation goals based on hand-crafted heuristics such as distance or information gain~\cite{keidar2014efficient,tao2023seer,dai2020fast,ho2025mapex}.
While effective, these approaches depend on reliable mapping and typically operate purely on geometric criteria.

Recent works extend frontier-based navigation by incorporating learning-based perception and semantic reasoning.
FrontierNet~\cite{boysun2025frontiernet} demonstrates that frontiers can be detected and evaluated directly from RGB images without constructing dense maps, enabling lightweight exploration driven by learned visual cues.
Vision-Language Frontier Maps (VLFM)~\cite{yokoyama2024vlfm} integrate vision-language priors into frontier-based navigation by constructing dense semantic frontier maps, allowing semantic relevance to guide frontier selection.

Semantic-enabled frontier representations are particularly natural for object-goal navigation.
Object-goal navigation has been widely studied in the context of embodied AI, where agents are tasked with navigating to objects specified by category labels or semantic descriptions.
Several benchmarks have been proposed, including HM3D ObjectNav~\cite{habitatchallenge2022,habitatchallenge2023} and OVON~\cite{ovon}, with OVON explicitly targeting open-vocabulary settings, thereby increasing the complexity of required semantic representations.
GOAT-Bench~\cite{khanna2024goat} further evaluates open-world object navigation with diverse object categories and environments. 

Early approaches rely on dense 3D reconstruction and semantic mapping pipelines, combining object detection with explicit spatial representations to guide navigation~\cite{zhang20233d,chaplot2020object}.
BeliefMapNav~\cite{beliefmapnav} constructs voxel-based belief maps to represent object likelihoods in 3D space, enabling zero-shot navigation at the cost of maintaining dense semantic state.
Similarly, GOAT~\cite{chang2023goat} maintains a dense semantic map from which global and local policies derive actions.
Another line of work uses map representations as inputs to reinforcement learning models to learn spatial information and navigation policies~\cite{xie2025naviformer,chaplot2021seal,chaplot2020object,ramakrishnan2022poni}.

Another line of work focuses on zero-shot navigation, particularly enabled by recent advances in large-scale foundation models.
CoWs~\cite{gadre2023cows} guides frontier-based exploration using CLIP-based semantic similarity.
OpenFMNav~\cite{openfmnav} leverages vision-language foundation models to support open-set object-goal navigation, while History-Augmented VLM~\cite{historyaugmented} incorporates memory mechanisms to improve zero-shot exploration.
ForesightNav~\cite{shah2025foresightnav} constructs CLIP-based semantic maps and trains a model to predict vision-language features for unexplored regions to guide navigation decisions.
IPPON~\cite{qu2024ippon} builds volumetric occupancy maps augmented with language embeddings during exploration and uses them for open-vocabulary goal reasoning.
UniGoal~\cite{yin2025unigoal} proposes a universal framework for zero-shot goal-oriented navigation by combining dense mapping, scene graph reasoning, and both LLMs and VLMs.
Despite the different approaches and design choices, these methods often require persistent semantic maps, scene graphs, or learned belief representations, and some further rely on interactive learning to acquire navigation knowledge.
\subsection{Vision-Language Navigation and Action Models}

The emergence of large language models (LLMs) and vision-language models (VLMs) has led to significant progress in vision-language navigation (VLN) and vision-language-action (VLA) systems~\cite{zhang2024vision}.
Many approaches train end-to-end policies conditioned on language instructions using reinforcement learning or imitation learning, often requiring large-scale interactive data collection and task-specific fine-tuning~\cite{zhang2024navid,zeng2025poliformer,wei2025streamvln}.
While effective in constrained settings, these methods typically struggle to generalize beyond training distributions and incur high computational cost at inference time.

Several recent works explore general navigation by directly leveraging pretrained foundation models.
InstructNav~\cite{instructnav} relies on large-scale, closed-source vision-language and language models to infer navigation actions directly from visual observations and instructions.
NAVILA~\cite{cheng2024navila} fine-tunes vision-language models using datasets extracted from real-world videos and maps them to robot action spaces through a hierarchical structure.
StreamingVLN~\cite{wei2025streamvln} designs a sliding-window key--value cache management architecture that enables navigation through multi-round dialogue.
Uni-NaVid~\cite{zhang2024uni} further fine-tunes vision-language navigation models on specific navigation benchmarks to achieve strong performance.

These approaches generally require large-scale datasets for fine-tuning and often rely on frequent model queries, such as direct action inference at every step.
Such designs can be computationally expensive and difficult to ground in metric navigation space.
Moreover, despite the significant training resources involved, many VLN systems primarily focus on target reaching or instruction following, which does not fully exploit the rich spatial and semantic information inherently captured by large-scale vision-language models.

%% file: method.tex
\section{Method}

OpenFrontier adopts a minimal but effective principle: \textit{detect and evaluate navigation targets directly in the 2D image domain whenever possible}. 
Following this principle, both goal identification and goal evaluation are carried out directly on RGB observations 
$\mathbf{I}_t \in \mathbb{R}^{H \times W \times 3}$, together with the associated camera extrinsic parameters 
$\mathbf{T}_t \in SE(3)$ and the camera intrinsic matrix $\mathbf{K} \in \mathbb{R}^{3 \times 3}$ at time step $t$. OpenFrontier outputs a goal pose, which can be further translated into control actions for the robot. Given a natural-language navigation goal, OpenFrontier processes the current observation $\mathbf{I}_t$ to identify a set of \emph{visual frontiers}. Each frontier is first represented as $\mathbf{F} = \{\mathbf{X}, g\}$, where $\mathbf{X} = \{\mathbf{p}, \mathbf{q}\}$ denotes the 3D position and orientation of the frontier in the world frame, and $g$ denotes its information gain. Importantly, no dense 3D semantic reconstruction is required for goal identification. The selected frontier is grounded in 3D space and used to generate a goal pose, which is then passed to a low-level motion planner for execution. The system iteratively updates frontier information gain and replans as new observations arrive, until the navigation goal is reached. An overview of the full system of OpenFrontier is shown in Fig. \ref{fig:overview}.

\begin{figure}
\centering
  \includegraphics[scale=0.21]{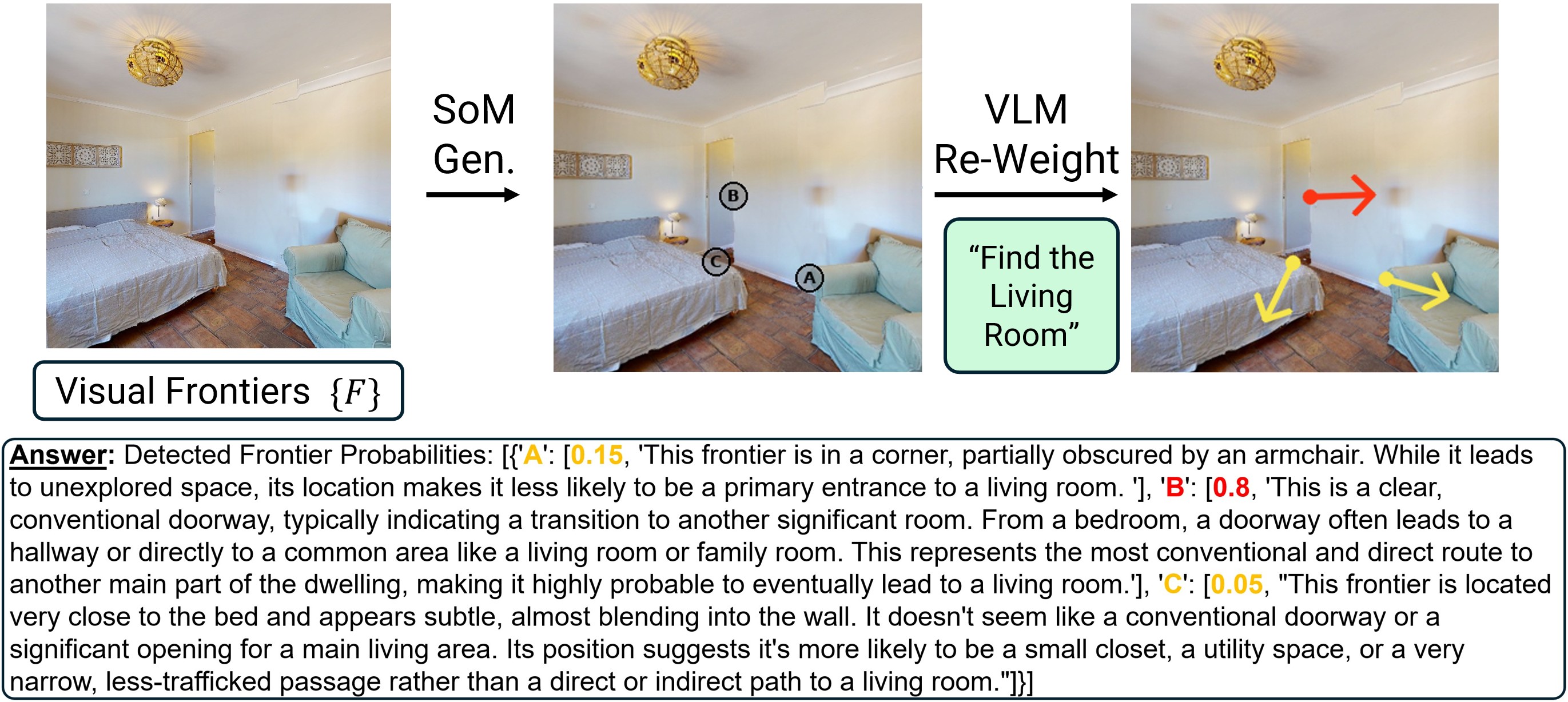}
\caption{The detected visual frontiers are jointly queried with the corresponding RGB image using a set-of-marks prompting strategy.
Each frontier is marked in the image, enabling the VLM to evaluate its relevance to the given navigation instruction within the local visual context.
The resulting relevance probabilities are used to re-weight its exploration-driven information gain from FrontierNet, effectively integrating task-specific semantic priors with exploration.
}
\label{fig:som}
\vspace{-3mm}
\end{figure}

\subsection{Visual-Frontier Identification}
A large body of robot navigation follows the convention of using geometry frontiers \cite{yamauchi1997frontier,Placed2023AFrontiers,keidar2014efficient,yokoyama2024vlfm,zhou2021fuel} extracted from volumetric maps as goal candidates for robot motion. Frontiers provide an effective and natural representation for encouraging exploration toward uncertain regions of the environment. Inspired by FrontierNet~\cite{boysun2025frontiernet}, we observe that frontiers can be detected and evaluated directly from a single 2D image, which we refer to as \emph{visual frontiers}, without relying on dense 3D mapping or separating detection from evaluation. We use \emph{visual frontier} and \emph{frontier cluster} interchangeably in the remainder of the paper.

Following FrontierNet, we apply a similar pipeline to a keyframe RGB observation $\mathbf{I}_t$ and, after post-processing, returns a set of frontier clusters $\mathbf{F}_i$, $i \in \mathbb{N}$. Each frontier cluster is represented in the image domain and then back-projected into 3D space to obtain its pose. The frontier identification pipeline also provides an information gain estimate $\hat{g}_i$, which captures a pure exploration prior by predicting the volume of unknown space associated with the frontier \cite{boysun2025frontiernet,cao2021tare,dai2020fast}. To incorporate task-specific semantic priors, we directly ground a VLM onto the frontier clusters in the image $\mathbf{I}_t$. Given a natural-language goal $\mathbf{L}$, the VLM assigns each frontier cluster a probability $p_i \in [0,1]$ indicating the likelihood that navigating toward that frontier will lead to accomplishing the goal. The final utility of each frontier is computed as:
\begin{equation}
     g_i = p_i \cdot \hat{g}_i,
\end{equation}
\begin{equation}
     p_i = \text{VLM}(\mathbf{I}_t, \mathbf{F}_i, \mathbf{L}),
\end{equation}
where $\hat{g}_i$ represents exploration-driven information gain and $p_i$ injects task-conditioned semantic guidance, which is obtained via the set-of-marks VLM query described below. This formulation naturally balances general exploration with goal-directed exploitation. 

Visual frontiers are detected and clustered based on visual cues in the input image, resulting in a set of frontier clusters localized by their centroid pixel coordinates in the image plane. These 2D locations provide a natural and explicit mechanism for guiding the attention of a vision-language model. To this end, we adopt a set-of-marks querying strategy that encourages the VLM to evaluate the correspondence between each frontier cluster and the language-conditioned goal using the surrounding image context. As illustrated in Fig. \ref{fig:som}, we overlay a visual marker \cite{yang2023set,shah2025bumble} at the 2D centroid of each frontier cluster on the RGB observation. The marked image, together with the natural-language goal, is jointly fed into the VLM using an additional prompt that instructs the model to output a probability score for each marker. An example prompt template is provided in the Appendix. This querying strategy enables the VLM to jointly reason about all candidate frontiers within a single forward pass, while leveraging fine-grained 2D visual context directly tied to actionable navigation locations. Moreover, by anchoring reasoning in the image plane, this formulation avoids requiring the VLM to perform explicit 3D spatial reasoning, where current models are known to be less reliable~\cite{feng2025seeing}.

\begin{table*}[t]
\vspace{2mm}
\centering
\small 
\setlength{\tabcolsep}{3pt} 
\renewcommand{\arraystretch}{1.15} 
\begin{tabular}{lc@{\hspace{14pt}}c@{\hspace{14pt}}|cc|cc|cc}
\toprule
\multirow{2}{*}{Method}
 & \multirow{2}{*}{Zero-shot}
 & \multirow{2}{*}{Spatial Representation}
 & \multicolumn{2}{c}{HM3D Val}
 & \multicolumn{2}{c}{MP3D Val}
 & \multicolumn{2}{c}{OVON Val Unseen} \\
\cmidrule(lr){4-5} \cmidrule(lr){6-7} \cmidrule(lr){8-9}
&
&
& SR (\%)$\uparrow$ & SPL (\%)$\uparrow$
& SR (\%)$\uparrow$ & SPL (\%)$\uparrow$
& SR (\%)$\uparrow$ & SPL (\%)$\uparrow$ \\
\midrule
History-Aug. VLM \cite{historyaugmented}
& \checkmark & 2D semantic \& textual history
& 46.0 & 24.8 & -- & -- & -- & -- \\

OpenFMNav \cite{openfmnav}
& \checkmark & dense 2D semantic
& 52.5 & 24.1 & 37.2 & 15.7 & -- & -- \\

InstructNav \cite{instructnav}
& \checkmark & dense 2D value map
& 58.0 & 20.9 & -- & -- & -- & -- \\

BeliefMapNav \cite{beliefmapnav}
& \checkmark & 3D voxel belief map
& 61.4 & 30.6 & 37.3 & 17.6 & -- & -- \\

VLFM \cite{yokoyama2024vlfm}
& \checkmark & dense 2D value map
& 52.5 & 30.4 & 36.4 & 17.5 & 35.2 & 19.6 \\

DAgRL+OD \cite{ovon}
& $\times$ & none
& -- & -- & -- & -- & 37.1 & 19.9 \\

UniGoal \cite{yin2025unigoal}
& \checkmark & 3D scene graph
& 54.5 & 25.1 & \textbf{41.0} & 16.4 & -- & -- \\

Uni-NaVid \cite{zhang2024uni}
& $\times$ & none
& 73.7 & \textbf{37.1} & -- & -- & \textbf{39.5} & 19.8 \\

\midrule
\textbf{OpenFrontier (Ours)}
& \checkmark & sparse visual frontiers
& \textbf{77.3} & 35.6
& 40.7 & \textbf{17.8}
& 39.0 & \textbf{20.1} \\
\bottomrule
\end{tabular}
\caption{\textbf{Performance Comparison} on Habitat ObjectNav benchmarks.
\emph{Zero-shot} indicates no fine-tuning on navigation tasks.
\emph{Spatial Representation} denotes the primary global representation each method uses for goal reasoning and planning.}
\label{table:objnav}
\vspace{-1.5mm}
\end{table*}

\begin{algorithm}[t]
\caption{Global Target Management}
\label{alg:manager_compact}
\begin{algorithmic}[1]
\Require Global frontier set $\mathcal{F}$, cleared set $\mathcal{C}$, proposals $\tilde{\mathcal{F}}_t$
\Require Robot position $\mathbf{p}_r$, PointNav $\pi$, goal $\mathbf{L}$
\Require Thresholds $\tau_{\text{merge}}, r_{\text{clear}}, r_{\text{near}}, r_{\text{goal}}$
\Require Progress parameters $(T_{\text{stall}}, \epsilon_{\text{stall}})$
\State $\texttt{hasTarget}\leftarrow 0$, $\mathbf{x}^*\leftarrow \mathbf{0}$
\State $\texttt{stall}\leftarrow 0$, $d_{\text{prev}}\leftarrow 10^9$

\While{true}
    \State $\mathcal{F}\leftarrow \textsc{MergeUpdate}(\mathcal{F},\tilde{\mathcal{F}}_t,\tau_{\text{merge}})$
    \State $\mathcal{F}\leftarrow \textsc{Prune}(\mathcal{F},\mathcal{C},r_{\text{clear}})$
    \State $\mathcal{F}\leftarrow \textsc{InsertViewpointIfAny}(\mathcal{F},\mathbf{L},r_{\text{near}})$

    \State $(\texttt{ok},\mathbf{F}^\star)\leftarrow \textsc{SelectBest}(\mathcal{F},\mathbf{p}_r)$
    \If{$\texttt{ok}=0$}
        \State \textbf{continue}
    \EndIf

    \State $(\mathbf{p}_r,d)\leftarrow \textsc{StepTo}(\pi,\mathbf{F}^\star)$
    \State $(\texttt{drop},\texttt{stall},d_{\text{prev}})\leftarrow$
    \Statex $\hspace{\algorithmicindent}\textsc{HandleProgress}(d,d_{\text{prev}},\texttt{stall},T_{\text{stall}},\epsilon_{\text{stall}})$

    \If{$\texttt{drop}=1$}
        \State $\mathcal{C}\leftarrow \mathcal{C}\cup\{\textsc{Pos}(\mathbf{F}^\star)\}$
        \State $\mathcal{F}\leftarrow \mathcal{F}\setminus\{\mathbf{F}^\star\}$
        \State \textbf{continue}
    \EndIf

    \If{$d<r_{\text{near}}$}
        \State $\mathcal{C}\leftarrow \mathcal{C}\cup\{\textsc{Pos}(\mathbf{F}^\star)\}$
        \State $\mathcal{F}\leftarrow \mathcal{F}\setminus\{\mathbf{F}^\star\}$
    \EndIf

    \State $(\texttt{hasTarget},\mathbf{x}^*)\leftarrow$
    \Statex $\hspace{\algorithmicindent}\textsc{VerifyAndSetTarget}(\mathbf{F}^\star,\mathbf{L},d,r_{\text{near}})$

    \If{$\texttt{hasTarget}=1$}
        \State $(\mathbf{p}_r,\_)\leftarrow \textsc{StepToPos}(\pi,\mathbf{x}^*)$
        \If{$\|\mathbf{p}_r-\mathbf{x}^*\|<r_{\text{goal}}$}
            \State \textbf{break}
        \EndIf
    \EndIf
\EndWhile
\end{algorithmic}
\end{algorithm}

\subsection{Frontier Management}
Long-horizon navigation requires maintaining and updating goals at a global level. We maintain a set of active frontier goals and manage their selection and update throughout the navigation process. Similar to classical exploration strategies, the utility of each frontier is periodically updated based on the current robot state. Specifically, for each frontier $i$, we compute a global utility
\begin{equation}
    u_i = \frac{g_i}{\|\mathbf{p}_r - \mathbf{p}_i\|},
\end{equation}
where $\mathbf{p}_r$ denotes the current position of the robot. This formulation favors frontiers that are both semantically relevant and spatially efficient to reach. The frontier with the highest utility is selected as the next navigation goal and passed to a low-level planner. The choice of low-level planner is flexible.
When no global map is available, a map-free PointGoal navigation policy~\cite{Wijmans2020DD-PPO,Yang-RSS-23,NavRL,yang2025spatially,sightoversite2025} can be used.
Alternatively, if a 3D volumetric representation is maintained, a map-based collision-free planner can be applied without modifying the frontier management logic.

While the robot navigates through the environment, the RGB observations are processed by an open-vocabulary segmentation model~\cite{carion2025sam3} to generate segmentation masks corresponding to the target. When a valid mask is detected, depth and camera pose are used to estimate the 3D centroid of the object. A new frontier with high (effectively infinite) utility is then instantiated at a fixed offset from the object, oriented to face it directly. This viewpoint frontier is inserted into the active frontier set and prioritized by the frontier manager, encouraging the robot to move to a configuration that provides direct visual access to the object.

After the robot reaches the viewpoint frontier, the corresponding RGB observation is queried by the same vision--language model to verify the presence of the target.
If the object is not confirmed, the viewpoint frontier is removed and the object hypothesis is discarded.
If the object is confirmed, the estimated centroid is designated as the final target position, and the robot is instructed to approach it as closely as possible.
Algorithm~\ref{alg:manager_compact} summarizes the global frontier management process from goal selection to termination.
For simplicity, the algorithm presents a compact formulation; additional implementation details for the helper functions are provided in the Appendix.

\subsection{Flexible Framework Design and Open-Context Query}

OpenFrontier is structured around a clear separation between visual-frontier perception/reasoning and global goal management. This separation is enabled by the use of navigation frontiers as the interface between the two components. We construct frontiers as a lightweight and interpretable abstraction that bridges visual-frontier reasoning with global decision making. As a result, each component of the pipeline can be modified or replaced independently, providing a high degree of flexibility in system design.

Importantly, OpenFrontier does not assume navigation-specific pre-training, fine-tuning, or in-context learning of the vision-language model. The VLM is queried in an open-context manner and can be replaced directly without modifying the rest of the system. Likewise, the framework does not require dense 3D reconstruction or semantic mapping. However, when available, simple geometric information can be incorporated to improve robustness. In our implementation, an optional lightweight volumetric map built from monocular depth priors is used only for basic filtering: occupied space is used to discard unreachable or unsafe frontiers, while free space is used to suppress frontiers whose exploration information gain has been exhausted during navigation. These geometric cues operate asynchronously and are not required for the reasoning steps, but consistently improve navigation performance in practice.

%% file: experiment.tex
\section{Experiment and Results}

\subsection{Experimental Setup}
We evaluate OpenFrontier along three dimensions: (1) performance on object-goal navigation tasks, (2) flexibility and generalization across environments, VLMs, and task settings, and (3) robustness to the sim-to-real gap. To measure the overall quantitative performance, we benchmark OpenFrontier on three object-goal navigation datasets: HM3D ObjectNav, MP3D ObjectNav \cite{habitatchallenge2022,habitatchallenge2023}, and OVON \cite{ovon}. All benchmarks follow the Habitat Navigation Challenge formulation. HM3D and MP3D consider closed-vocabulary object goals, while OVON extends the task to open-vocabulary settings with more diverse language descriptions.

We implement OpenFrontier within the Habitat framework and evaluate navigation performance using standard metrics: Success Rate (SR, \%) and Success weighted by Path Length (SPL, \%). We compare against a set of baseline methods spanning dense 3D mapping approaches, belief-based planners, and recent vision-language navigation methods. For object-goal benchmark experiments, we use Gemini-2.5-flash as the VLM model. Low-level navigation is executed using a DD-PPO point-goal policy \cite{Wijmans2020DD-PPO} with pretrained weights from VLFM \cite{yokoyama2024vlfm}, which matches the Habitat action space. Visual frontier detection is performed using FrontierNet with pretrained weights from \cite{boysun2025frontiernet}. We use the same set of system parameters across all benchmarks. Empirically, we run frontier detection and reasoning every six steps. All experiments are conducted on a machine equipped with a single RTX 4090 GPU (24GB). Details about parameter settings can be found in the Appendix.

\begin{figure}
\centering
  \includegraphics[scale=0.06]{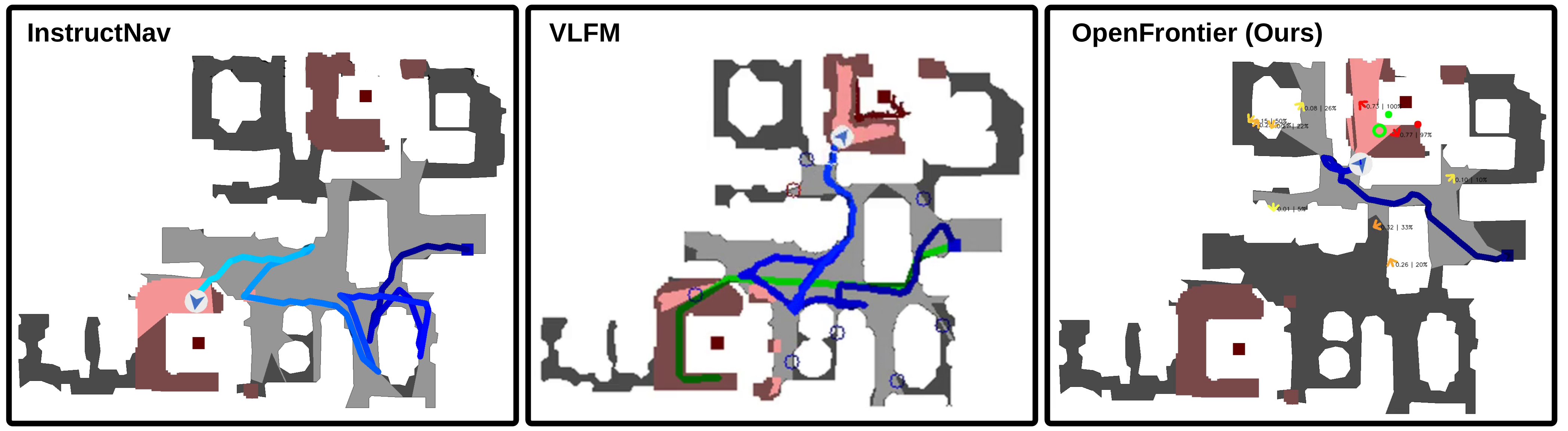}
\caption{\textbf{Navigation Results} across two representative baseline methods and OpenFrontier (HM3D: \texttt{5cdEh9F2hJL}, goal: bed).
The red square and shaded region indicate the ground-truth target location and its success region.
OpenFrontier shows stronger decision-making ability, e.g.\ at multi-choice intersections, navigating directly toward the target region while avoiding redundant exploration of irrelevant areas.}

\label{fig:comparison}
\vspace{-2.5mm}
\end{figure}


\begin{figure}
\centering
  \includegraphics[scale=0.303]{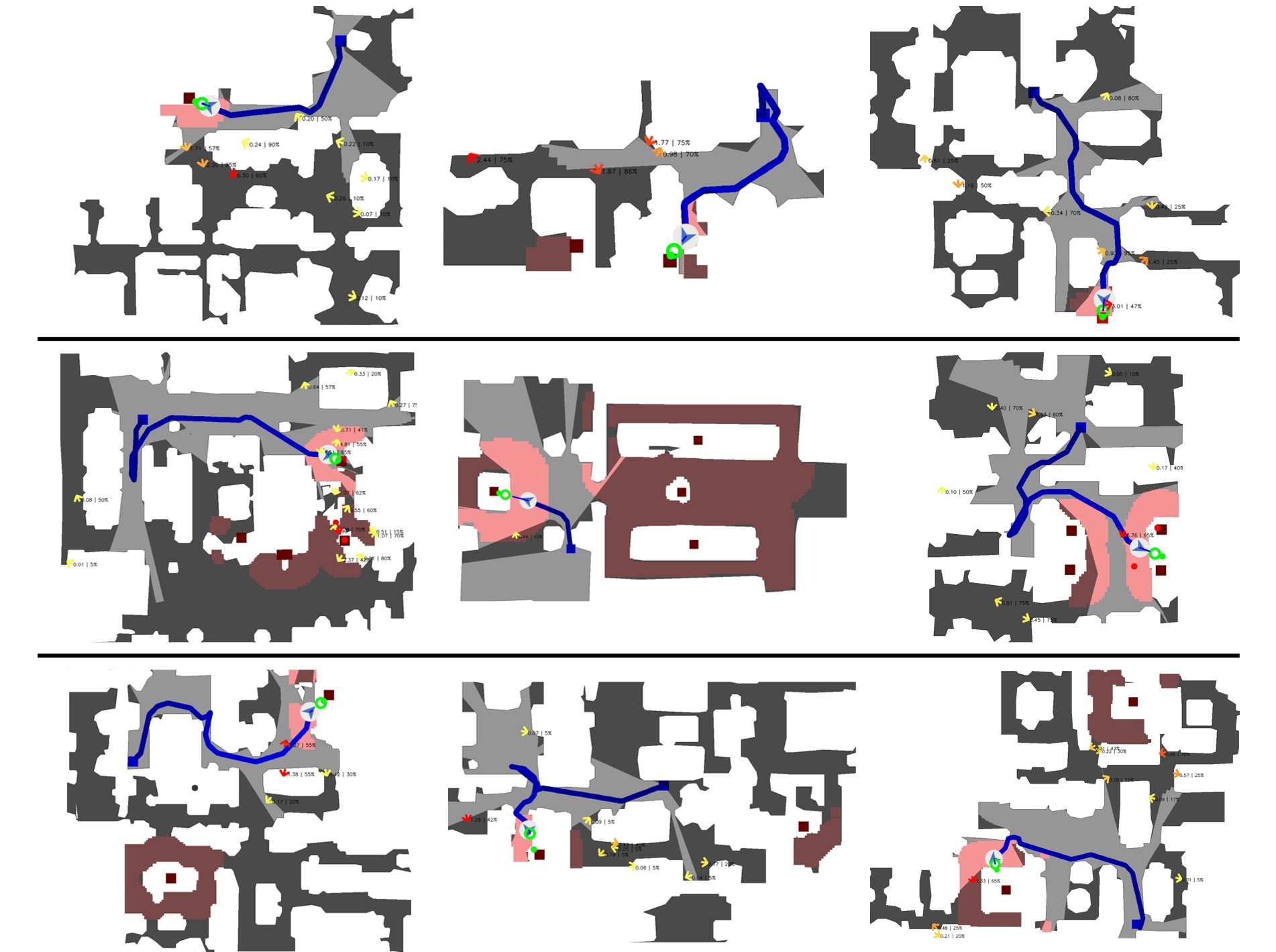}
\caption{\textbf{Additional Navigation Examples}. 
\textbf{Top:} OVON scenes with goals (left to right) \emph{refrigerator}, \emph{picture}, and \emph{dishwasher}.
\textbf{Middle:} MP3D scenes with goals \emph{stool}, \emph{table}, and \emph{cushion}.
\textbf{Bottom:} HM3D scenes with goals \emph{sofa}, \emph{toilet}, and \emph{bed}.
All experiments are conducted using the same system configuration and parameters across datasets.}

\label{fig:nav_qual_result}
\vspace{-1mm}
\end{figure}

\begin{figure*}[t]
\centering
\includegraphics[scale=0.32]{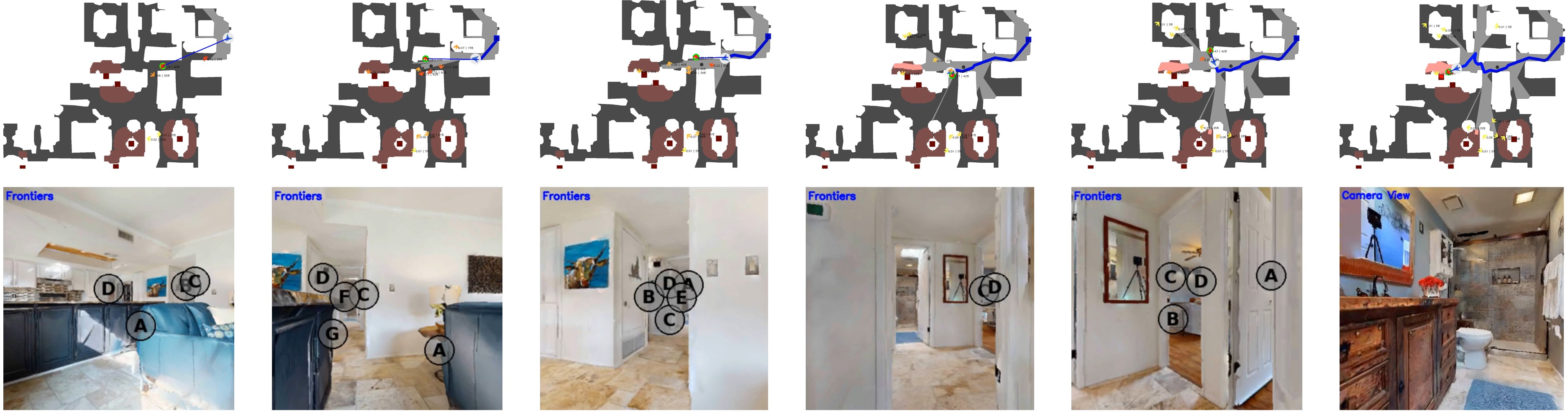}
\hspace{0.5cm}
\includegraphics[scale=0.315]{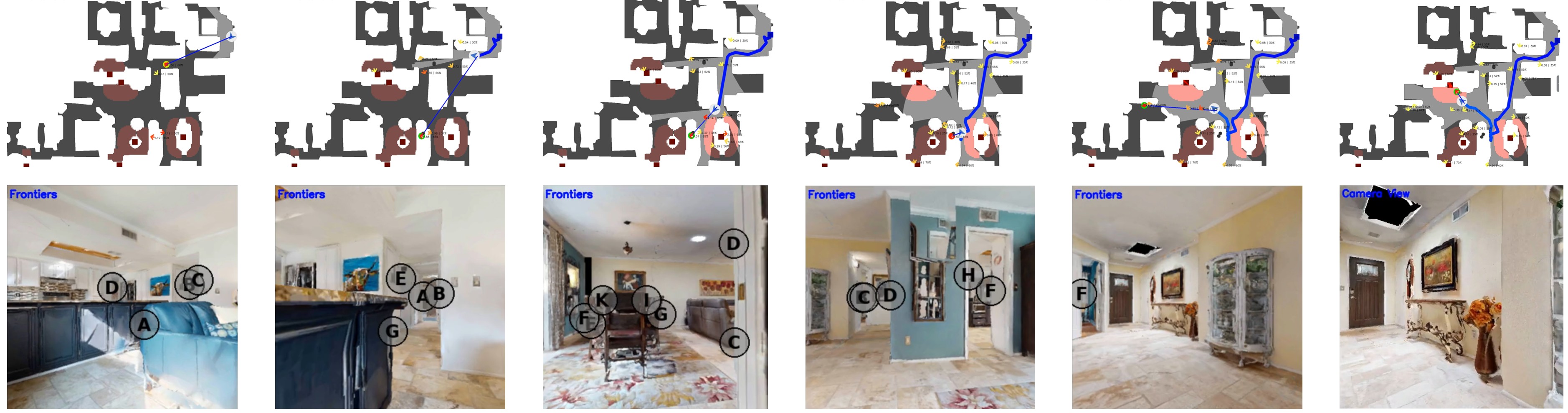}
\caption{\textbf{OpenFrontier Navigation Example with Different Goal Contexts} (HM3D: \texttt{5cdEh9F2hJL}).
Top: target is \emph{``plant in the bathroom.''}
Bottom: target is \emph{``plant.''}
The robot is initialized at the same starting location in both runs.
From left to right, we show selected frames along the navigation trajectory together with the corresponding image observations overlaid with detected frontiers. The rightmost image is the final observation that triggers termination once the target enters the field of view.
Despite observing similar frontier locations, OpenFrontier assigns different relevance probabilities depending on the goal context.
As a result, the top trajectory prioritizes regions likely associated with the bathroom, while the bottom trajectory moves toward the living room, which also commonly contains plants.}

\label{fig:complex_context}
\vspace{-1mm}
\end{figure*}

\begin{figure}
\centering
  \includegraphics[scale=0.039,trim=0 3.7cm 0 0,clip]{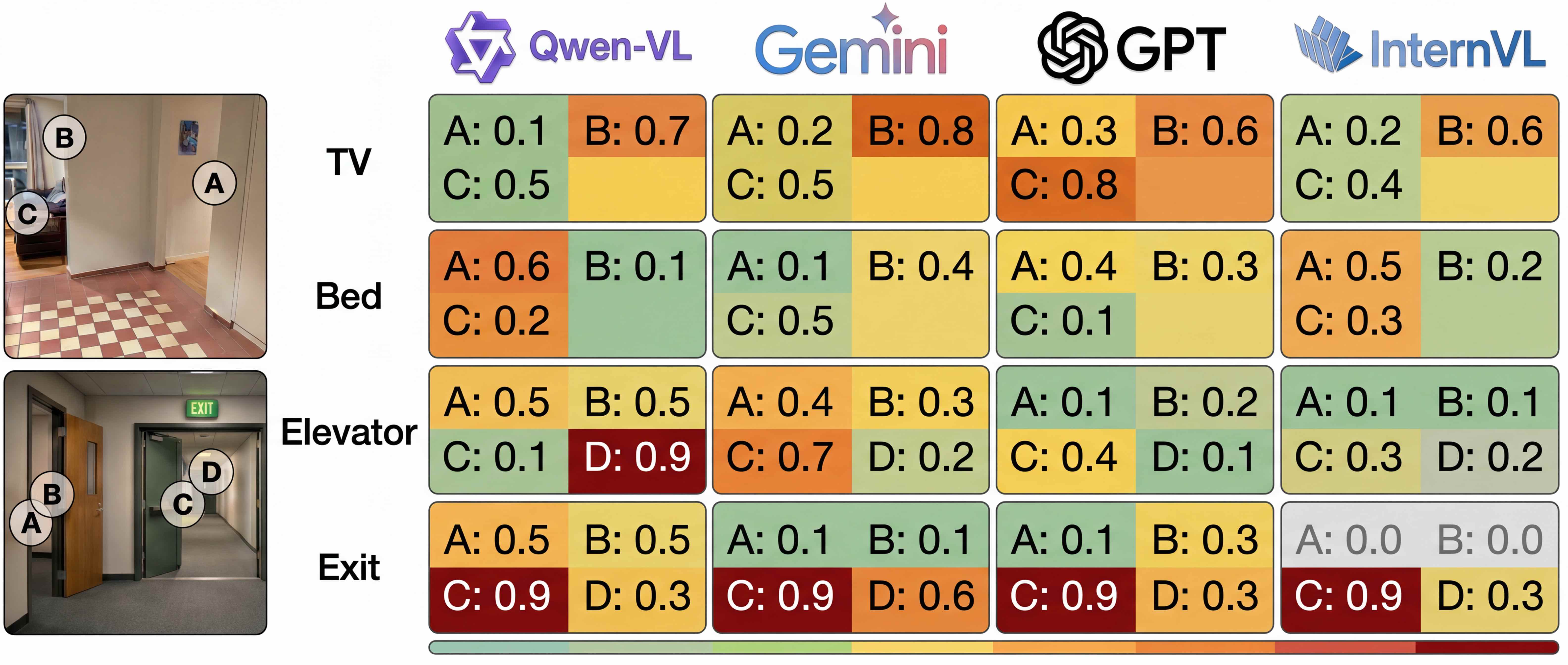}
\caption{\textbf{Visual Frontier Reasoning Examples.}
Four different VLMs are evaluated for frontier probability estimation (without normalization) using the set-of-marks querying strategy (left to right: Qwen3-VL, Gemini-2.5, GPT-4o, InternVL-3.5).
All models operate on the same real-world image with the same prompt after frontier detection.}

\label{fig:vlm_comparison}
\vspace{-1mm}
\end{figure}

\begin{figure*}[t]
\centering
\includegraphics[width=.95\linewidth]{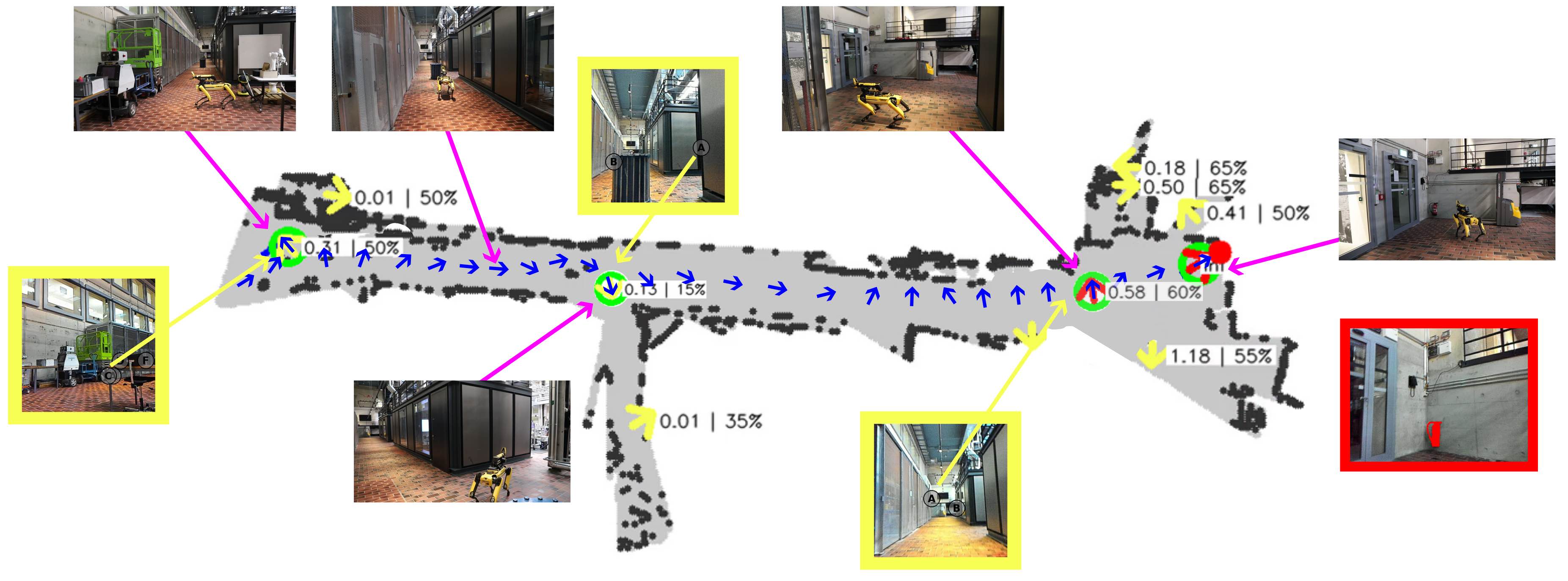}
\caption{\textbf{Real-world Deployment}. Example of a deployment of OpenFrontier queried to find a fire extinguisher. The blue arrows illustrate the path of the robot through the environment. The yellow boxes mark the VLM prompt images for different keyframes, connected to some key 3D frontiers marked during navigation on the map. The red box shows the detection of the target object that signals successful task completion.}
\label{fig:robot}
\vspace{-1mm}
\end{figure*}

\subsection{ObjectNav Benchmark Evaluation}

OpenFrontier achieves competitive performance across all three benchmarks compared to state-of-the-art baselines. Quantitative results are summarized in Table~\ref{table:objnav}. As the system only processes keyframes and operates on a sparse set of frontiers, it avoids high-frequency inference such as step-by-step mapping updates used by History-Augmented VLM, or direct action inference as in Uni-NaVid. Under a single, uniform configuration across all scenes, OpenFrontier outperforms most baselines on all three datasets. Compared to the strongest competing methods, OpenFrontier is 1.5\% lower in SPL on HM3D validation and 0.5\% lower in SR on OVON than Uni-NaVid, and 0.3\% lower in SR than UniGoal. Notably, Uni-NaVid is a VLN model fine-tuned on these benchmarks and their action spaces, while UniGoal relies on dense mapping, scene graph construction, and the combined use of both a LLM and a VLM model. Despite these additional assumptions, UniGoal performs more than 20\% lower in SR than OpenFrontier on HM3D. Overall, OpenFrontier adopts a minimal design while maintaining stable and competitive performance across benchmarks. Additionally, we provide a failure case decomposition in Fig.~\ref{fig:failure_sankey}, with more detailed analysis in the Appendix.

To further inspect qualitative behavior beyond aggregate numbers, we re-ran two representative baselines, InstructNav and VLFM, under the same evaluation setup on the entire HM3D validation set. InstructNav uses both advanced VLM and LLM models, while VLFM adopts a frontier-based strategy combined with dense mapping and feature integration. As shown in Fig.~\ref{fig:comparison}, OpenFrontier demonstrates higher navigation efficiency in complex scenes with multiple branching corridors, consistent with the quantitative results. Fig.~\ref{fig:nav_qual_result} presents additional qualitative examples across all three evaluation datasets.
In these scenarios, OpenFrontier demonstrates effective spatial and semantic reasoning, allowing the agent to bias exploration toward goal-relevant regions and reach the target efficiently.
Additional qualitative results are provided in the Appendix.

\begin{table}[t]
\centering
\renewcommand{\arraystretch}{1.1}
\begin{tabular*}{\columnwidth}{@{\extracolsep{\fill}}lcc}
\toprule
\textbf{Method (VLM)} & \textbf{SR (\%)$\uparrow$} & \textbf{SPL (\%)$\uparrow$} \\
\midrule
UniGoal (LLaVA 1.6) & 54.5 & 25.1 \\
History-Aug. VLM (LLaVA 1.6) & 46.0 & 24.8 \\
InstructNav (Gemini-2.5-flash) & 52.0 & 22.0 \\
\midrule
OpenFrontier (LLaVA 1.6) & 74.2 & 35.3 \\
OpenFrontier (InternVL3\_5-8B) & 74.5 & 34.5 \\
OpenFrontier (Gemma-3-4b-it) & 76.9 & 33.7 \\
OpenFrontier (Gemini-2.5-flash) & \textbf{77.3} & \textbf{35.6} \\
\bottomrule
\end{tabular*}
\caption{\textbf{VLM Sensitivity and Cross-Method Comparison} on HM3D.
The swapped VLM here refers to the frontier-reweighting reasoning model; the InstructNav (Gemini-2.5-flash) entry is a re-run with the substituted VLM and therefore differs from the original number in Table~\ref{table:objnav}.
Within OpenFrontier, VLM-induced variance is small ($\le 3$\% SR), showing that our framework offers a stable and universal interface for leveraging VLMs. At a matched VLM, OpenFrontier substantially outperforms baselines, indicating that the gain is driven by the visual-frontier design rather than by VLM strength.}
\label{tab:hm3d_vlm}
\end{table}

Vision-language models are used in OpenFrontier for two purposes: frontier utility reweighting and goal verification. Goal verification is triggered only when a candidate target is detected by SAM3 \cite{carion2025sam3}. Neither component requires task-specific training or fine-tuning, allowing the VLM to be easily replaced with alternative models. To evaluate this flexibility, we run the full HM3D evaluation using several other VLMs. As shown in Table~\ref{tab:hm3d_vlm}, replacing Gemini-2.5-flash \cite{comanici2025gemini} with other publicly available or open-source models, including the older LLaVA 1.6 \cite{liu2023visual}, results in only a marginal performance decrease ($\le 3$\% SR) while maintaining strong compatibility with our framework.
To further isolate the contribution of the system design from that of the VLM, the same table also reports a cross-method comparison in which baselines and OpenFrontier are evaluated using the same VLM backbone.
At a matched VLM, OpenFrontier substantially outperforms UniGoal, History-Aug.\ VLM, and InstructNav, indicating that our gains are not attributable to VLM strength alone but to the visual-frontier representation and the system design around it. We further evaluate a broader set of VLMs on real-world images to compare their image-space reasoning ability with frontier selection capability. The qualitative comparison in Fig. \ref{fig:vlm_comparison} illustrates the differences in reasoning behavior across models. Despite differences in the exact probability values generated by different VLMs, all tested models provide reasonable probability estimates. These results suggest that stronger VLMs can improve OpenFrontier through a simple model switch, while the framework remains robust to the choice of VLM.

Beyond category-level object navigation, we also evaluate OpenFrontier under more complex language conditions that include spatial or appearance context, which the quantitative protocol does not easily capture. Fig.~\ref{fig:complex_context} shows an example comparing the goals \textit{``Plant''} and \textit{``Plant in Bathroom''}. The resulting trajectories and frontier probability assignments differ accordingly, with OpenFrontier prioritizing bathroom-related frontiers under the spatially conditioned query. Additional examples are provided in the Appendix. These results demonstrate that OpenFrontier is able to exploit contextual reasoning from VLMs and successfully ground it onto frontiers to support goal specifications with more complex context.

\subsection{Real-World Deployments}

Finally, we evaluate OpenFrontier in real-world settings.
The results in Fig.~\ref{fig:vlm_comparison} already demonstrate the visual frontier identification module on real indoor images with human-specified language targets, verifying its generalization beyond simulation. We further deploy OpenFrontier on a legged robot (Boston Dynamics Spot) and conduct navigation tasks in a large-scale indoor environment. The system is integrated within ROS, with RGB observations provided by the camera mounted on the robot arm end-effector. The camera pose is obtained from the robot’s onboard vision--inertial odometry. The robot is started at different initial locations and given different language inputs for the target, without any prior knowledge of the scene layout.
Fig.~\ref{fig:robot} shows an example in which the robot autonomously navigates to a fire extinguisher in a large indoor environment without any human intervention.
{We further conduct a small-scale quantitative evaluation in the indoor environment, covering 5 different target objects with 10 navigation trials in total, and obtain a 70\% success rate.}

\section{Insights}
OpenFrontier occupies an intermediate design point between recent advanced navigation approaches that either rely on rich 3D map representations or require training VLN/VLA models for navigation settings.
Despite its zero-shot setup, OpenFrontier demonstrates stable and competitive performance across benchmarks and transfers effectively from simulation to real-world deployment.
In this section, we discuss several insights that help explain these observations.

\subsection{Why the Design Is Effective}
A key observation from OpenFrontier is that \emph{a lightweight navigation pipeline can be highly effective when appropriate representations and system interfaces are used}. {We attribute the effectiveness of OpenFrontier to three aspects of the visual-frontier abstraction.}

\begin{figure}[!t]
\centering
\textcolor{red}{%
\begin{minipage}[c]{0.44\linewidth}
\centering
\includegraphics[width=\linewidth]{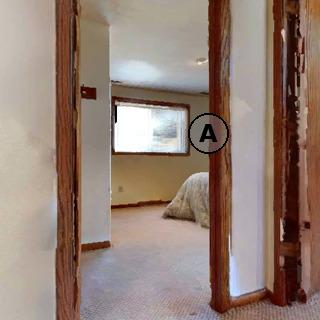}
\end{minipage}\hspace{0.01\linewidth}%
\begin{minipage}[c]{0.183\linewidth}
\centering
\includegraphics[width=\linewidth]{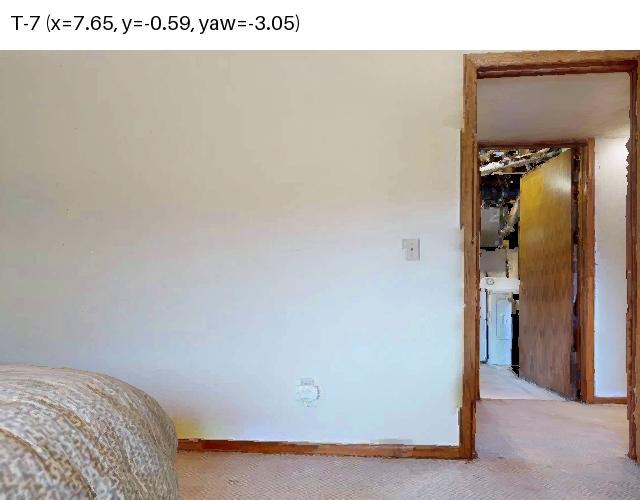}\\[2pt]
\includegraphics[width=\linewidth]{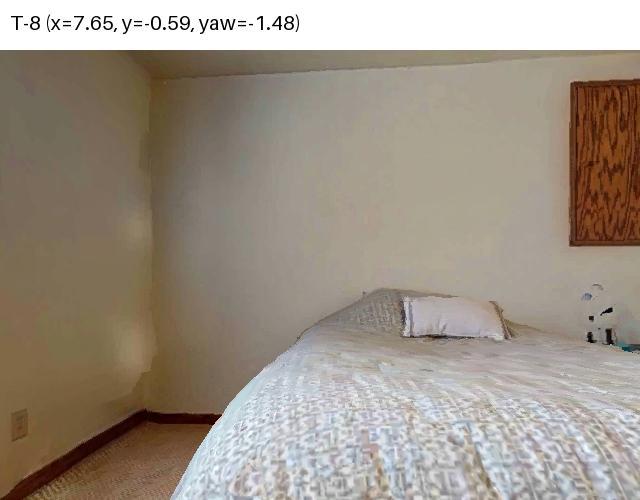}\\[2pt]
\includegraphics[width=\linewidth]{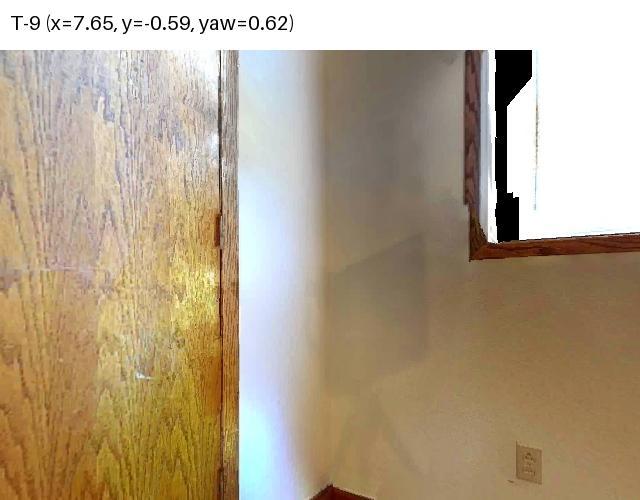}
\end{minipage}%
}
\caption{{\textbf{History-Augmented Frontier Reasoning}.
Left: current observation with set-of-marks frontier annotations.
Right: three prior keyframes from the same room, passed as additional context.
Given television as the navigation target, without history, the VLM assigns $p{=}0.65$ to Frontier~A based on category-level priors about bedrooms.
With history, the VLM recognizes that the bedroom has already been observed without a television and lowers the probability to $p{=}0.10$.}}
\label{fig:history_aug}
\end{figure}

{\textbf{(1) Detection and reasoning operate in a shared image space.}
Because visual frontiers are detected and evaluated on the same RGB observation, the VLM sees the full visual context behind each candidate without any lossy intermediate representation, and is asked to perform a task it is much closer to its training setting: 2D visual reasoning rather than 3D spatial reasoning over multi-view inputs or abstract scene-level representations.
This yields three direct consequences.
\emph{(a)~Direct Adaptation.} No fine-tuning is needed for the VLMs to give high-quality reasoning results.
\emph{(b)~Robustness to the VLM backbone.}
Swapping the reweighting VLM across four backbones, including the older LLaVA~1.6, induces only $\le 3$\% SR variation (Table~\ref{tab:hm3d_vlm}). Combined with the absence of any fine-tuning or in-context learning, this gives OpenFrontier a truly zero-shot operating mode in which most of the inductive load is carried by the representation rather than by the VLM. The same image-space framing also explains our advantage at matched VLM backbones (Table~\ref{tab:hm3d_vlm}).
\emph{(c)~Natural extensibility.} The query interface is straightforward to extend with additional context. As a proof-of-concept, a history-augmented variant simply appends a small set of recent keyframes to the VLM query (Fig.~\ref{fig:history_aug}), so the model can refine a frontier's relevance in light of what has already been observed, while frontier detection, global management, and the rest of the pipeline remain unchanged.}

{\textbf{(2) The utility merges complementary exploration and exploitation priors.}
The frontier utility $g_i = p_i\,\hat{g}_i$ combines a geometric exploration prior $\hat{g}_i$ from FrontierNet (the expected unobserved volume revealed by visiting the frontier) with a semantic exploitation prior $p_i$ from the VLM (high-level scene and object knowledge conditioned on the goal).
These priors are genuinely complementary: FrontierNet has no notion of task semantics, while FrontierNet outperforms current VLMs on estimating unobserved 3D volume. Combining them in a single utility lets each compensate for the other's weakness and naturally balances coverage of unexplored regions with goal-directed bias.}

{\textbf{(3) The system has a sparse, transferable, and hierarchically aligned interface.}
Detected in 2D, visual frontiers are then grounded in 3D, where they directly serve as actionable navigation targets. Their sparsity makes them robust to noisy sensor input, and inexpensive to maintain and update over long horizons, providing a form of global memory without dense semantic state. The same abstraction also aligns naturally with the standard high-level / low-level decomposition of robot navigation: frontiers carry the high-level semantic decisions while a low-level controller handles execution, letting each component operate at the level of abstraction where it is most effective. Future work may explore tighter but principled integration between these two levels, enabling mutual feedback while preserving the benefits of modularity.}

\begin{figure*}[!t]
\centering
\includegraphics[width=0.32\textwidth,trim=0 3cm 0 0,clip]{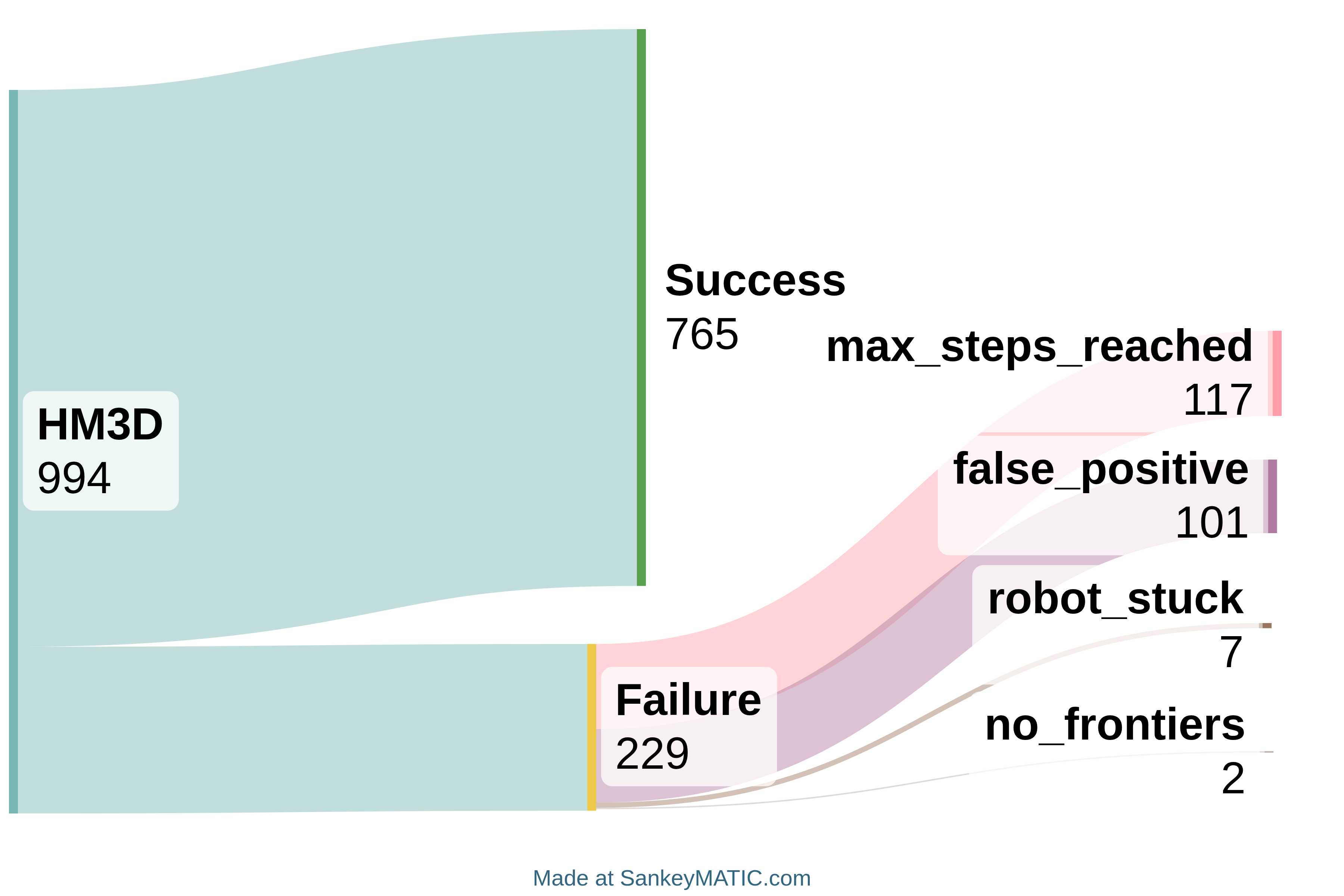}\hfill
\includegraphics[width=0.32\textwidth,trim=0 3cm 0 0,clip]{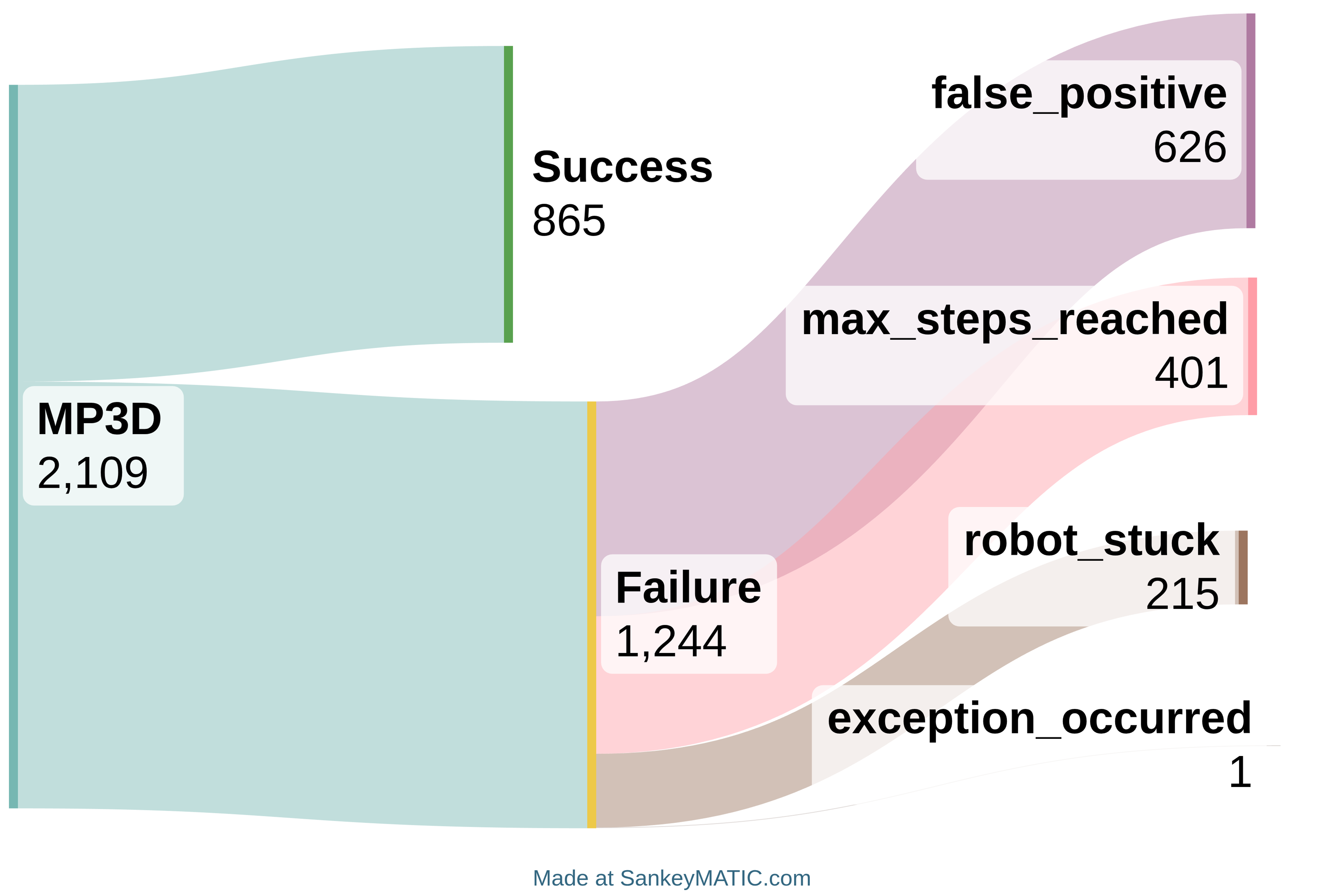}\hfill
\includegraphics[width=0.32\textwidth,trim=0 3cm 0 0,clip]{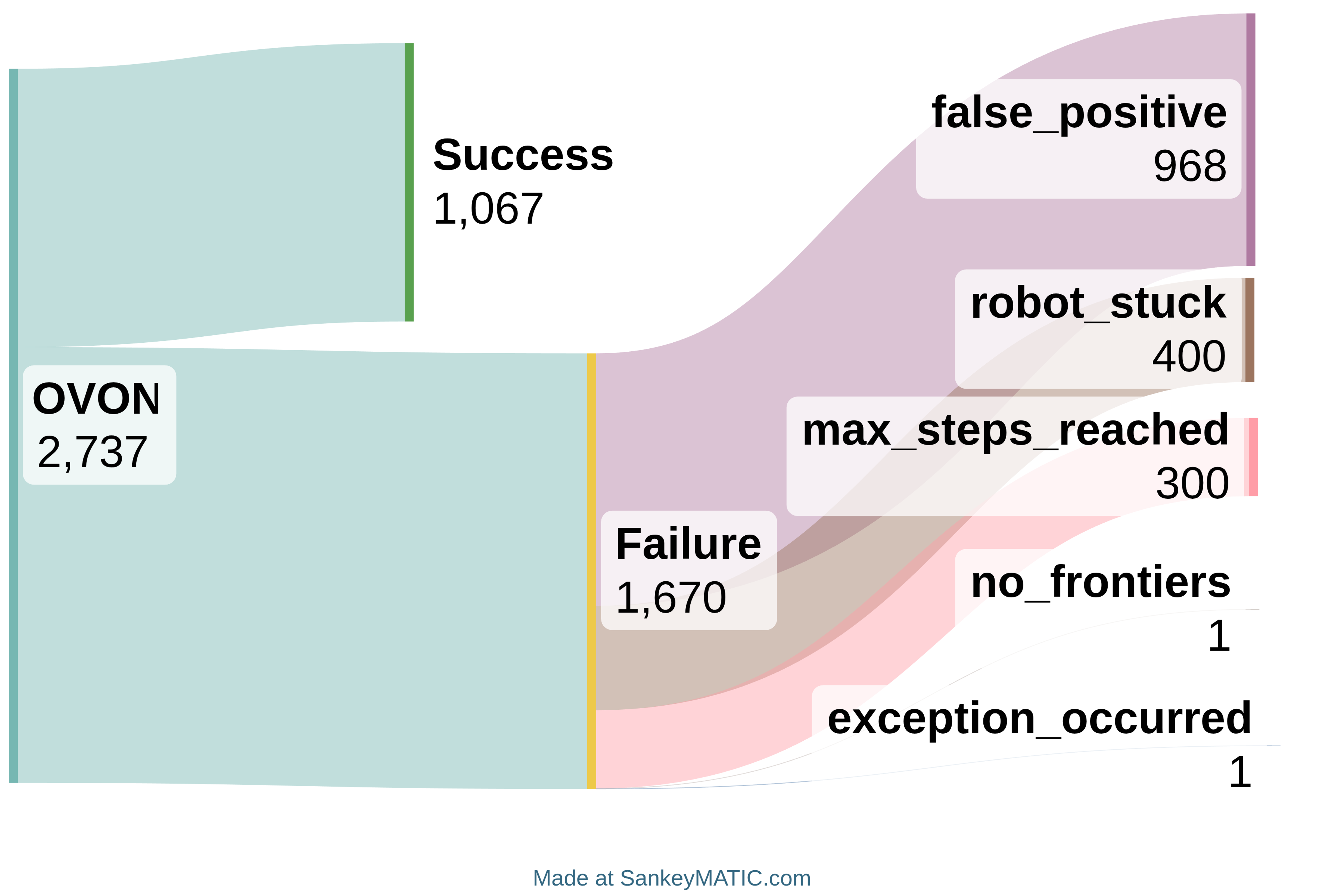}
\caption{\textbf{Failure Case Analysis.}
(Left to right: HM3D, MP3D, and OVON)
Across all three benchmarks, the two most common failure modes are false-positive target detections and termination due to exhausting the step budget.}
\label{fig:failure_sankey}
\vspace{-1mm}
\end{figure*}

\subsection{When and Why It Fails}
Despite strong benchmark performance, OpenFrontier remains susceptible to several failure modes.
As Fig.~\ref{fig:failure_sankey} shows, failure types differ across benchmarks but follow a consistent pattern.
One major challenge lies in termination criteria, which is non-trivial in open-world navigation settings.
Some failures are influenced by limitations of the Habitat benchmark itself.
However, our analysis indicates that these cases also expose a broader limitation of OpenFrontier, namely its limited ability to recover from failure once progress stalls. Even after a target is detected and navigation reduces to point-goal execution, the robot may move in incorrect directions or become trapped in local minima.
In such cases, the system currently lacks a mechanism to quickly recognize failure and trigger re-reasoning or replanning at the semantic level.
Enabling reliable failure detection and recovery is expected to significantly improve robustness and overall performance, and remains an important direction for future work.

%% file: conclusion.tex
\section{Conclusion}

We presented OpenFrontier, a zero-shot framework that enables general language-conditioned navigation by grounding vision-language priors onto visual navigation frontiers. By treating navigation as a discrete subgoal selection problem and performing perception and semantic reasoning primarily in the 2D image domain, OpenFrontier avoids dense semantic mapping, task-specific policy training, and fine-tuning. Frontiers serve as sparse and physically grounded anchors that bridge high-level semantic reasoning with metric navigation, enabling efficient exploration and goal-directed behavior in both closed-set and open-set environments. We demonstrated that this design achieves competitive performance across multiple object-goal navigation benchmarks and transfers robustly to real-world deployment on a mobile legged robot. Our results suggest that effective grounding and system-level abstraction, rather than increasingly complex models or training pipelines, play a central role in scalable open-world navigation. We believe OpenFrontier provides a practical and flexible foundation for integrating future vision-language models into robotic navigation systems without requiring costly retraining or dense environment representations.

\vspace{-0.4\baselineskip}

%% file: sup_tech.tex
\section{Methodology and Implementation Details}

\subsection{Configuration Settings and Key Parameters}

For simulation evaluation and benchmarking, we use Habitat~0.2.4 as the simulation environment.
All experiments follow the standard ObjectNav task configuration, using the default action-space agent with a step size of 0.25\,m and a turn angle of 30$^\circ$.
The agent is equipped with a 640$\times$480 RGB-D camera, and each episode is capped at a maximum of 500 steps, consistent with the Habitat ObjectNav Challenge settings~\cite{habitatchallenge2022,habitatchallenge2023}.
Camera intrinsic parameters and scene bounds are obtained directly from the simulator.

All benchmarking experiments are conducted on a headless server equipped with an NVIDIA RTX~4090 GPU with 24\,GB of VRAM.
FrontierNet~\cite{boysun2025frontiernet}, SAM3~\cite{carion2025sam3}, and InternVL~\cite{zhu2025internvl3} (when applicable) are executed locally on the GPU.
Gemini and Gemma~\cite{gemmateam2025gemma3technicalreport} models are accessed via API calls.
The key parameters used in the simulation benchmarks are summarized in Table~\ref{tab:params}.
Unless otherwise specified, the same configuration is used across all experiments.

\subsection{Emergency rotation mechanism}

Occasionally, a benchmarking episode begins with the agent facing a wall or a room corner where no valid frontiers are visible, or the agent may drive itself into configurations in which no frontier can be identified to continue exploration. To address this, an emergency rotation mechanism is introduced by inserting three artificial ``rotation frontiers'' with \( g = 1 \) at the agent's position, oriented to the left, right, and opposite the current viewing direction, thereby forcing an in-place rotation. If, after clearing all three rotation frontiers, no new exploration frontiers are detected, the filtering threshold is relaxed by updating \( g^{\text{min}} \gets 0.5 \cdot g^{\text{min}} \). This procedure is repeated up to a maximum of \( n_{\text{emergency}} \) iterations, after which the failure condition \textit{no\_frontiers} is triggered.


\subsection{VLM Prompt}

\subsubsection{Frontier Probability Estimation (Set-of-Marks)}

\begin{quote}
\ttfamily
Assume the labeled frontiers \{labels\} represent possible places to go.
Each frontier is a detected boundary between explored and unexplored space.

Estimate the probability that each frontier leads to (or is already around) a
\textit{\{target\_object\}} when moving towards it and continuing exploration.
Unseen labels should have a probability of 0.5.

Consider long-term navigation potential, not only immediate visibility.
Some frontiers may lead to larger unexplored regions.
Use the local neighborhood context around each frontier.

Return only a JSON list with one dictionary.
Each key is a frontier label; each value is a list:
(1) a probability in $[0,1]$, and
(2) a short explanation.

Format:
\{\{"A": [0.3, "reason"], "B": [0.2, "reason"], ...\}\}
\end{quote}

\subsubsection{Target Presence Verification}

\begin{quote}
\ttfamily
Based on this image, estimate the probability that a
\textit{\{target\_object\}} is in the camera field of view,
within five meters, and reachable.

If the object is reflected in a mirror, behind glass,
barely visible, heavily occluded, or unreachable,
it should be considered absent.

Probabilities should be close to 0 (absent) or 1 (present).
Add one sentence of reasoning.

Return a JSON list with one dictionary.

Format:
\{\{"probability": 0.9, "reason": "reason"\}\}
\end{quote}

{
\subsubsection{Optional add-on for history enhancement}
\begin{quote}
\ttfamily
[...] Consider the history of images of already visited places, and adjust the probabilities accordingly, reducing probabilities for frontiers that lead into already visited areas or look similar to already visited places. [...] 
\end{quote}
}

\begin{table}[]
\centering
\begin{tabular}{l|l}
\toprule
\multicolumn{2}{c}{\textbf{Frontier}} \\
\midrule
$r_{near}$ & 0.2m \\
$r_{goal}$ & 1.0m \\
$r_{clear}$ & 0.5 \\
$\epsilon_{stall}$ & 0.1m \\ 
$T_{stall}$ & 15 \\
$p_{presence}$ & 0.7 \\
$n_{emergency}$ & 3 \\
$\tau_{merge}$& 1.8m \\
$g^{min}$ & 0.5 \\
$K$ & 11 \\
Prediction Interval & 6 steps \\
$\alpha$ & 10 \\
\midrule
\multicolumn{2}{c}{\textbf{PointNav}} \\
\midrule
Backbone & Resnet50 \\
RNN Type & LSTM \\
Hidden Layers & 2 \\
\midrule
\multicolumn{2}{c}{\textbf{Models}} \\
\midrule 
FrontierNet Weights & rgbd\_11cls \\
PointNav Weights  & vlfm/pointnav\_weights \\
SAM3 Tokenizer & bpe\_simple\_vocab\_16e6 \\
Gemini Model & gemini-2.5-flash \\
Gemma 3 Model & gemma-3-4b-it \\
InternVL Model & InternVL3\_5-8B \\
\midrule
\end{tabular}

\caption{\textbf{Configuration Settings} for simulated benchmarking.}

    \label{tab:params}
\end{table}

\begin{algorithm}[t]
\caption{InsertViewpointIfAny}
\label{alg:viewpoint}
\begin{algorithmic}[1]
\Require Frontier set $\mathcal{F}$, object set $\mathcal{X}$
\Require Target object $\mathbf{L}$, separation distance $r_{sep}$
\Require RGB image $\mathbf{I}$, depth image $D_{cam}$
\Require Robot position $p_r$, robot pose $\mathcal{R}_{world}$
\State $\mathcal{M} \gets \textsc{GetSegmentationMasks}(\mathbf{I}, \mathbf{L})$
\If{$|\mathcal{M}| > 0$}
    \State $\mathcal{P}_{cam} \gets \textsc{MaskDepth}(D_{cam}, \mathcal{M})$
    \State $\mathcal{P}_{world} \gets \textsc{ProjectToWorld}(\mathcal{P}_{cam}, \mathcal{R}_{world})$
    \State $\mathcal{X} \gets \textsc{AddCentroid}(\mathcal{P}_{world}, \mathcal{X})$
    \State $\mathcal{X} \gets \textsc{DBSCAN}(\mathcal{X})$
    \For{$\forall \mathbf{x_i} \in \mathcal{X}$}
        \State $\mathbf{f_z} \gets \dfrac{\mathbf{x_i} - p_r}{\|\mathbf{x_i} - p_r\|}$
        \State $\mathbf{f_x} \gets \dfrac{\hat{\mathbf{y}} \times \mathbf{f_z}}{\|\hat{\mathbf{y}} \times \mathbf{f_z}\|}$
        \State $\mathbf{f_y} \gets \dfrac{\mathbf{f_z} \times \mathbf{f_x}}{\|\mathbf{f_z} \times \mathbf{f_x}\|}$
        \State $p_f \gets x_i - r_{sep}\mathbf{f_z}$
        \State $
        f \gets
        \begin{bmatrix}
        \mathbf{f_x} & \mathbf{f_y} & \mathbf{f_z} & p_f \\
        0 & 0 & 0 & 1
        \end{bmatrix}$
        \State $\mathbf{F} \gets \textsc{CreateFrontierWithObject}(f, x_i)$
        \State $\mathcal{F} \gets \mathcal{F}\cup\{\textbf{F}\}$
    \EndFor
\EndIf
\end{algorithmic}
\end{algorithm}

\begin{algorithm}[t]
\caption{HandleStall}
\label{alg:stall}
\begin{algorithmic}[1]
\Require Current distance $d$, previous distance $d_{\text{prev}}$
\Require Stall counter \texttt{stall}
\Require Minimum progress $\epsilon_{\text{stall}}$, stall threshold $T_{\text{stall}}$
\If{$d_{\text{prev}} - d \ge \epsilon_{\text{stall}}$}
    \State $\texttt{stall} \gets 0$
\Else
    \State $\texttt{stall} \gets \texttt{stall} + 1$
\EndIf
\If{$\texttt{stall} \ge T_{\text{stall}}$}
    \State $\texttt{drop} \gets \textbf{true}$
\Else
    \State $\texttt{drop} \gets \textbf{false}$
\EndIf
\State $d_{\text{prev}} \gets d$
\State \Return $(\texttt{drop}, \texttt{stall}, d_{\text{prev}})$
\end{algorithmic}
\end{algorithm}

\begin{algorithm}[t]
\caption{VerifyAndSetTarget}
\label{alg:target_presence}
\begin{algorithmic}[1]
\Require Distance $d$, threshold $r_{near}$
\Require Image $\mathbf{I}$, target object $\mathbf{L}$
\Require Presence threshold $p_{\text{presence}}$
\Require Frontier $\textbf{F}^\star$
\State $\texttt{hasTarget} \gets 0$
\If{$d < r_{near}$}
    \State $p \gets \textsc{VLMDetect}(\mathbf{I}, \mathbf{L})$
    \If{$p > p_{\text{presence}}$}
        \State $\texttt{hasTarget} \gets 1$
        \State $\mathbf{x}^* \gets \textsc{GetLinkedObject}(\textbf{F}^\star)$
    \EndIf
\EndIf
\State \Return $(\texttt{hasTarget}, \mathbf{x}*)$
\end{algorithmic}
\end{algorithm}

\subsection{Real-World Experiment}

We implement OpenFrontier as a ROS package and deploy it on a Boston Dynamics Spot robot with an arm.
A calibrated RGB camera mounted on the arm's end-effector provides $640 \times 480$ images at approximately 3\,Hz.
Depth information is estimated from RGB images using Metric3Dv2~\cite{hu2024metric3d}.
FrontierNet, SAM3, and Metric3D are executed on a desktop workstation equipped with an NVIDIA RTX~4090 GPU with 24\,GB of VRAM, while the Gemini-2.5-Flash model is queried via API calls. For real-world deployment, we do not use the DD-PPO policy.
Instead, we construct an occupancy grid using the metric depth estimates and perform conventional map-based path planning. Mapping is carried out using Wavemap~\cite{reijgwart2023wavemap}.
Planning is implemented as a ROS service that is triggered either by changes in the selected goal or at fixed time intervals, using the RRT-star planner from OMPL~\cite{sucan2012the-open-motion-planning-library}, following the setup in~\cite{boysun2025frontiernet}.
Path execution is handled by a simple pure-pursuit controller which sends velocity command to the robot.

Frontier prediction, object segmentation, and mapping are performed asynchronously at around 3\,Hz with latest incoming camera images.
Frontier management and goal selection run concurrently at approximately 16\,Hz, triggered by odometry updates from the robot. 

Key parameters specific to the real-world deployment are summarized in Table~\ref{tab:realworld}. Compared to simulation, we increase the strength of frontier filtering to better handle more challenging environments, such as cluttered background, glass surfaces, and suppress frontiers detected at high elevations.

\begin{table}[]
    \centering
\begin{tabular}{l|l}
\toprule
\multicolumn{2}{c}{\textbf{Frontier}} \\
\midrule
$r_{near}$ & 0.5m \\
$g^{min}$ & 0.01 \\
Prediction Interval & 1 step (3HZ loop) \\
Planning Interval  & 100 steps (16HZ loop) \\
Planning Algorithm & $RRT^\star$ \\
Max planning time & 1.5 sec \\
\midrule
\multicolumn{2}{c}{\textbf{Models}} \\
\midrule 
Metric3D & metric\_depth\_vit\_large\_800k \\
\midrule
\multicolumn{2}{c}{\textbf{Mapper and Planner}} \\
\midrule
Vox size & 0.1m \\
Max depth range & 3.5m \\
Max planning time & 1.5s \\
Min dist to occupied & 0.5m \\
Max dist to free & 1.5m \\ 
\midrule
\end{tabular}
\caption{\textbf{Configuration Settings} for real-world experiment.}
    \label{tab:realworld}
\end{table}

\subsection{Metric Calculation}

Overall success rate (SR) was obtained as the total amount of successful episodes over the total amount of available and feasible episodes. Success is determined by Habitat as "if on outputting a STOP command, the agent is within 1.0m Euclidean distance from any instance of the target object category AND the object can be viewed by an oracle from that stopping position by turning the agent or looking up/down." Overall SPL was calculated as the average of each individual episode's SPL reported by Habitat's internal metrics. For HM3D (V2), MP3D and OVON benchmarks, all available episodes specified in the Val split (Val Unseen for OVON) were used. Episodes whose shortest path length was reported as "infinite" by Habitat, i.e. those without any feasible path from starting location to a target object, were not included in the set of feasible episodes. 

%% file: sup_qualitative.tex
\section{Additional Results}

\subsection{Additional Exploration Examples on HM3D, MP3D, and OVON}
We present additional navigation examples on the three benchmark datasets used in the main paper: HM3D, MP3D, and OVON.
All experiments are conducted using the same system configuration and parameters as those reported in the main paper and described in the previous section.

Figures~\ref{fig:qualitative_results_hm3d}, \ref{fig:qualitative_results_mp3d}, and \ref{fig:qualitative_results_ovon} show representative navigation trajectories from HM3D, MP3D, and OVON, respectively.
Each example visualizes a top-down view of the robot’s exploration trajectory from the start location (blue square) to the target success region (red area), along with a snapshot of the frontier distribution at the final navigation step.

\begin{figure}
\centering
  \includegraphics[scale=0.15]{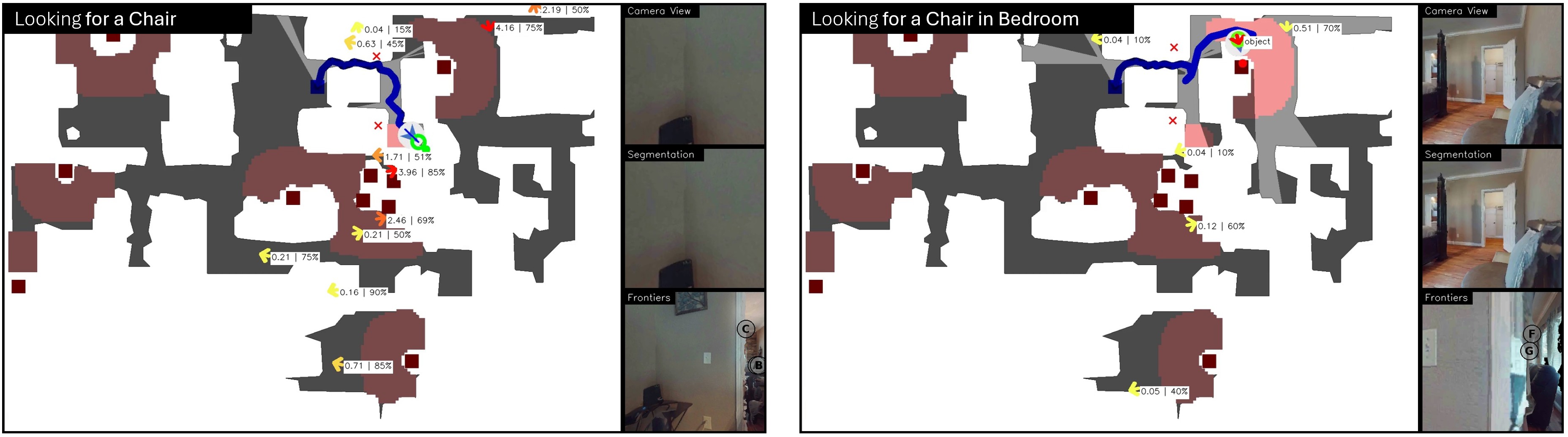}
  \includegraphics[scale=0.148]{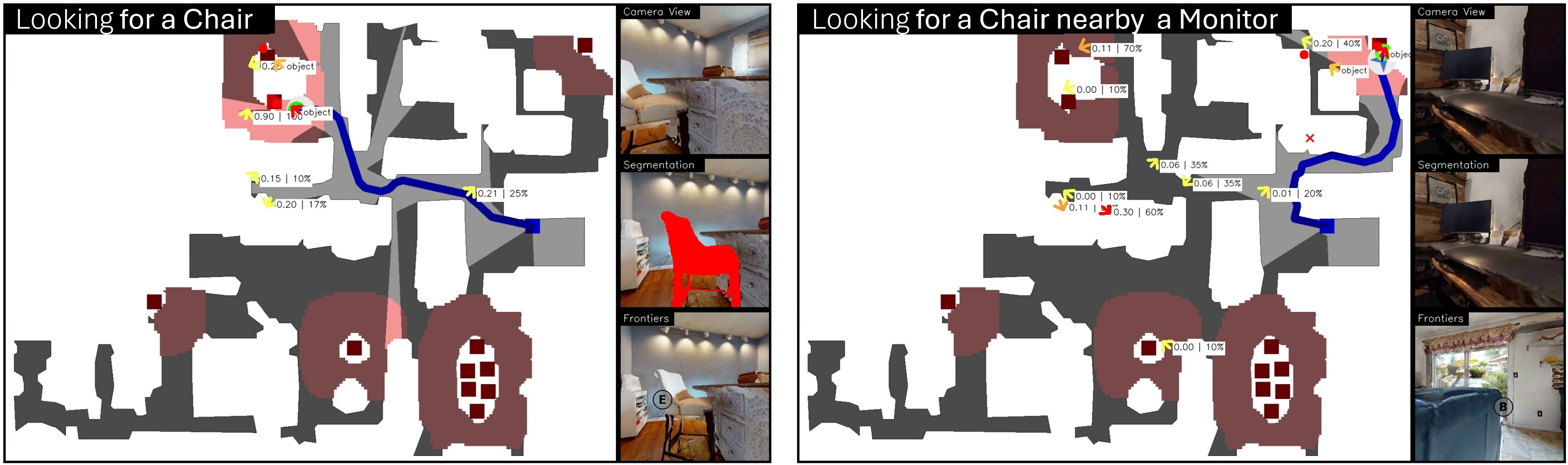}
\caption{\textbf{Context-aware Navigation to the Same Object.}
Top: \emph{chair} vs. \emph{chair in the bedroom}. Bottom: \emph{chair} vs. \emph{chair near a monitor}.}

\label{fig:context_nav}
\vspace{-1mm}
\end{figure}

\begin{figure*}[t]
    \centering

    \begin{subfigure}[t]{0.32\textwidth}
        \centering
        \includegraphics[width=\linewidth]{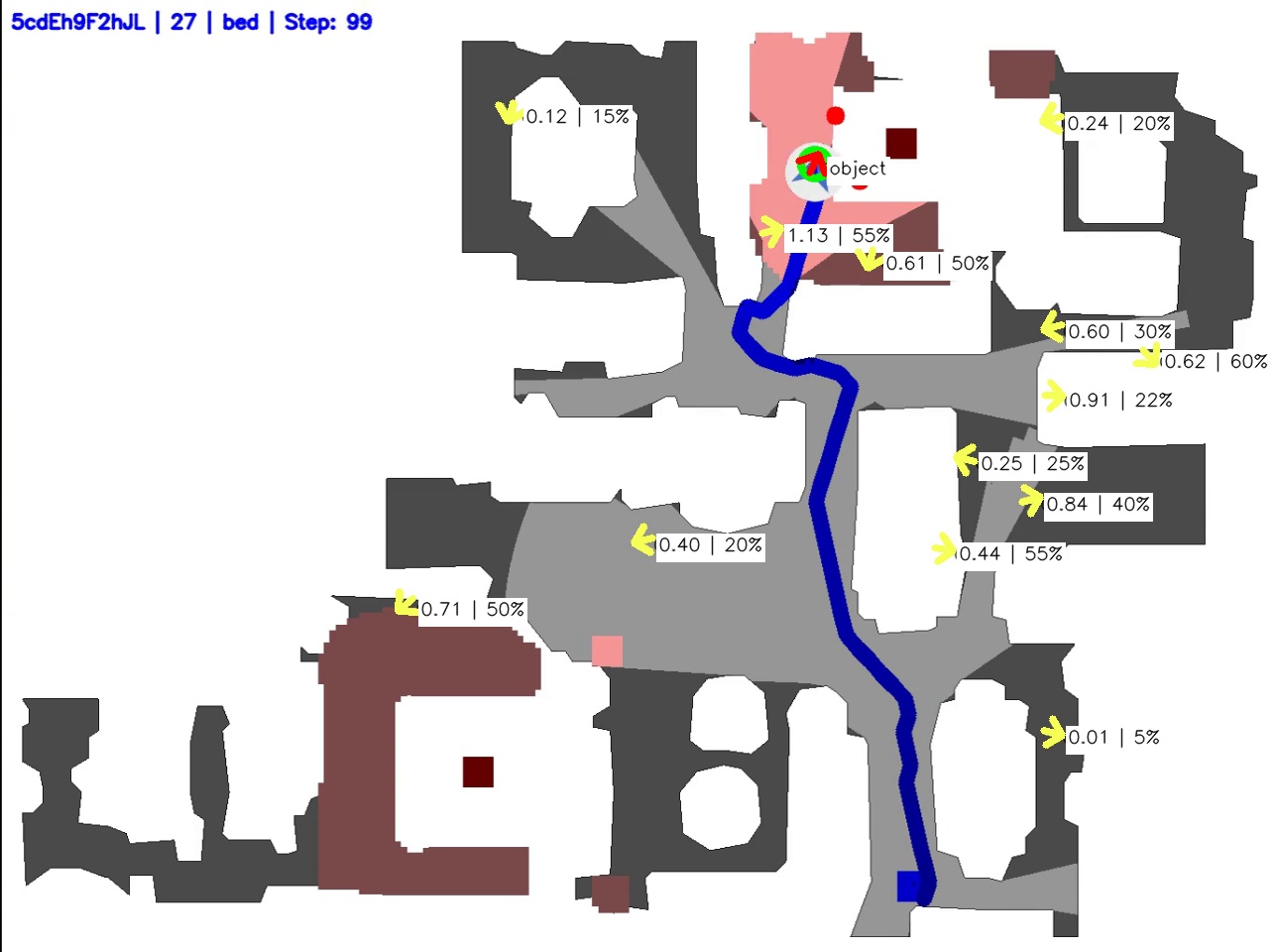}
        \caption{Scene: 5cdEh9F2hJL, Target: Bed}
    \end{subfigure}
    \hfill
    \begin{subfigure}[t]{0.32\textwidth}
        \centering
        \includegraphics[width=\linewidth]{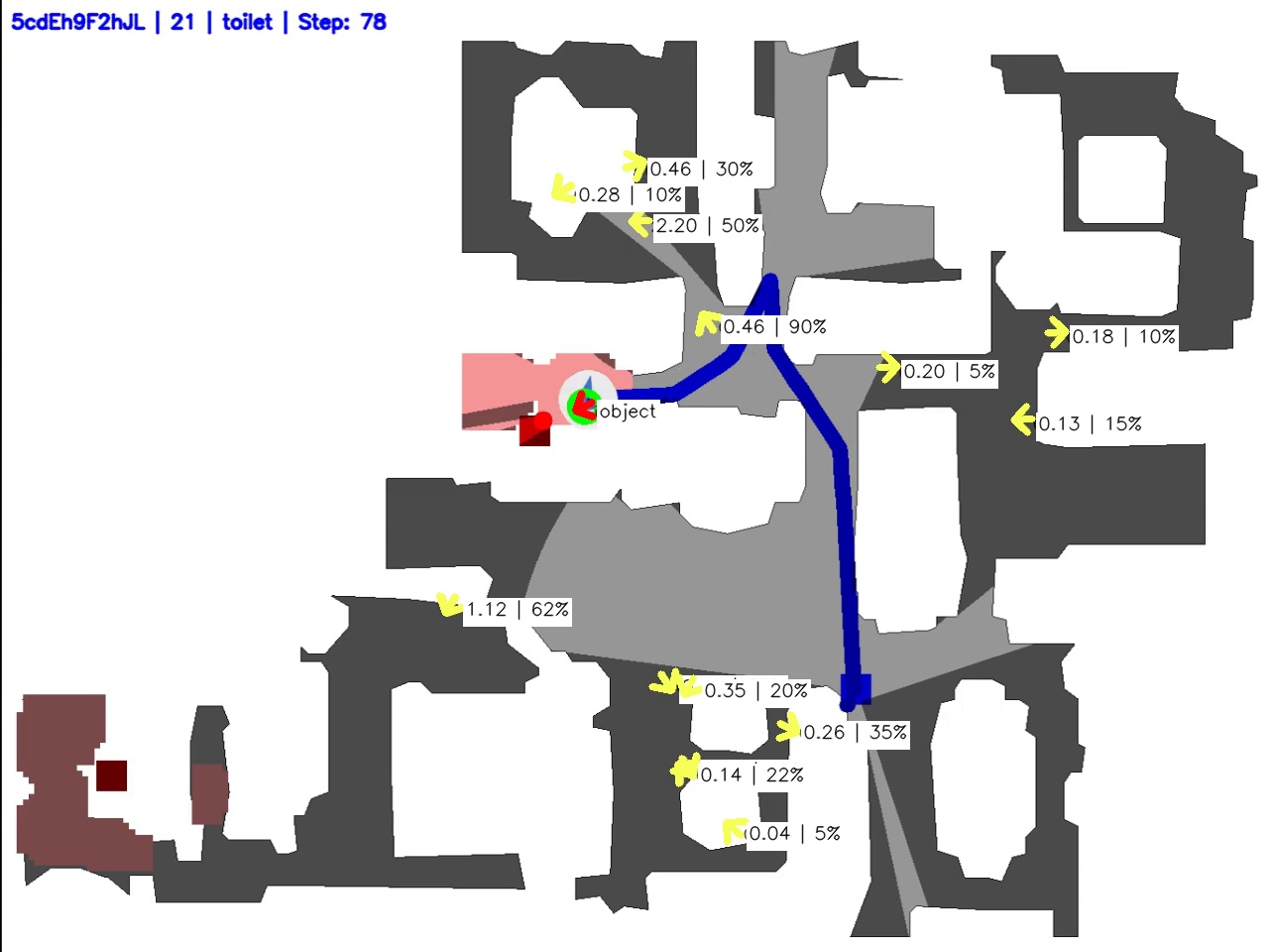}
        \caption{Scene: 5cdEh9F2hJL, Target: Toilet}
    \end{subfigure}
    \hfill
    \begin{subfigure}[t]{0.32\textwidth}
        \centering
        \includegraphics[width=\linewidth]{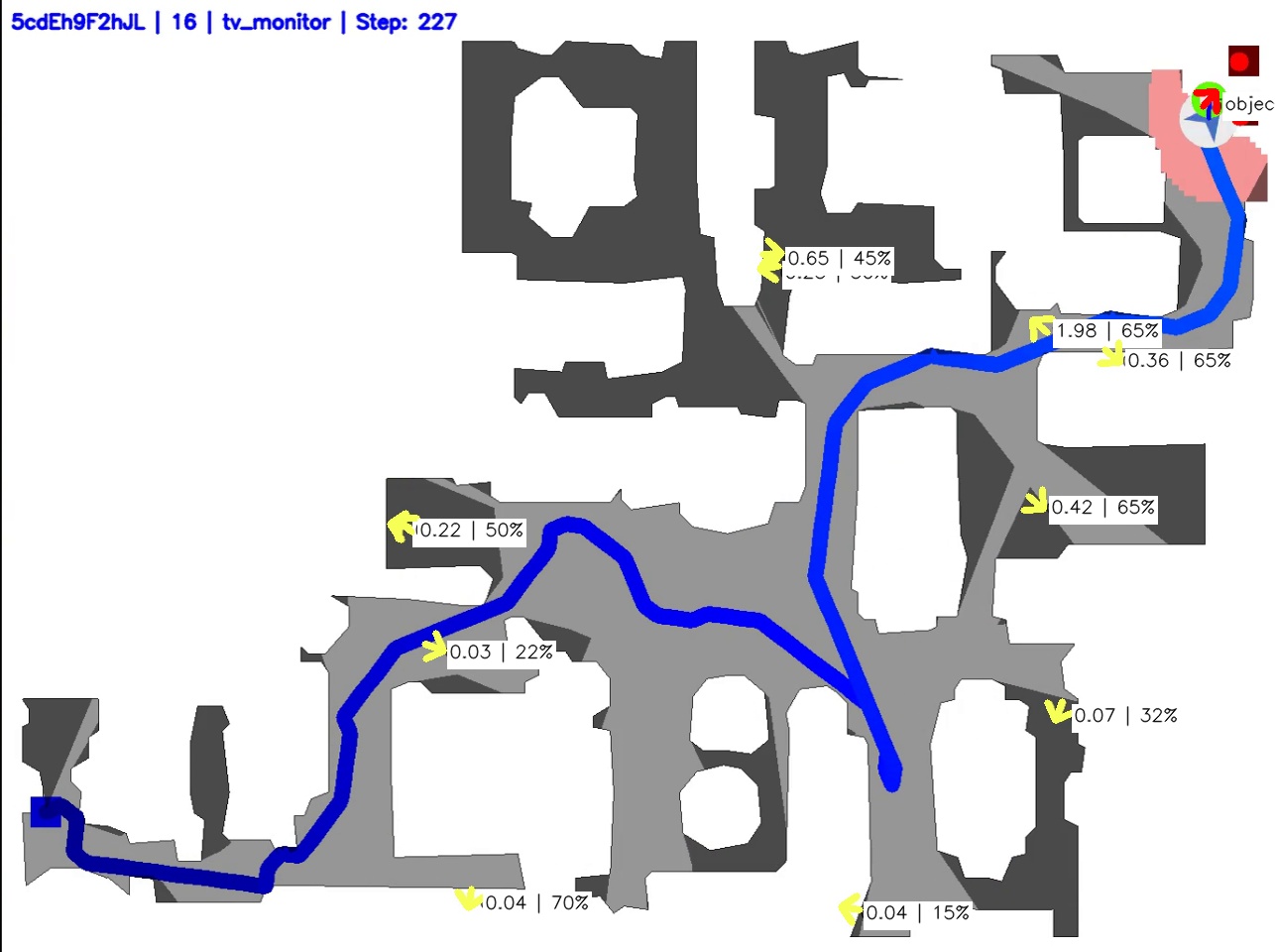}
        \caption{Scene: 5cdEh9F2hJL, Target: TV monitor}
    \end{subfigure}

    \vspace{0.6em}

    \begin{subfigure}[t]{0.32\textwidth}
        \centering
        \includegraphics[width=\linewidth]{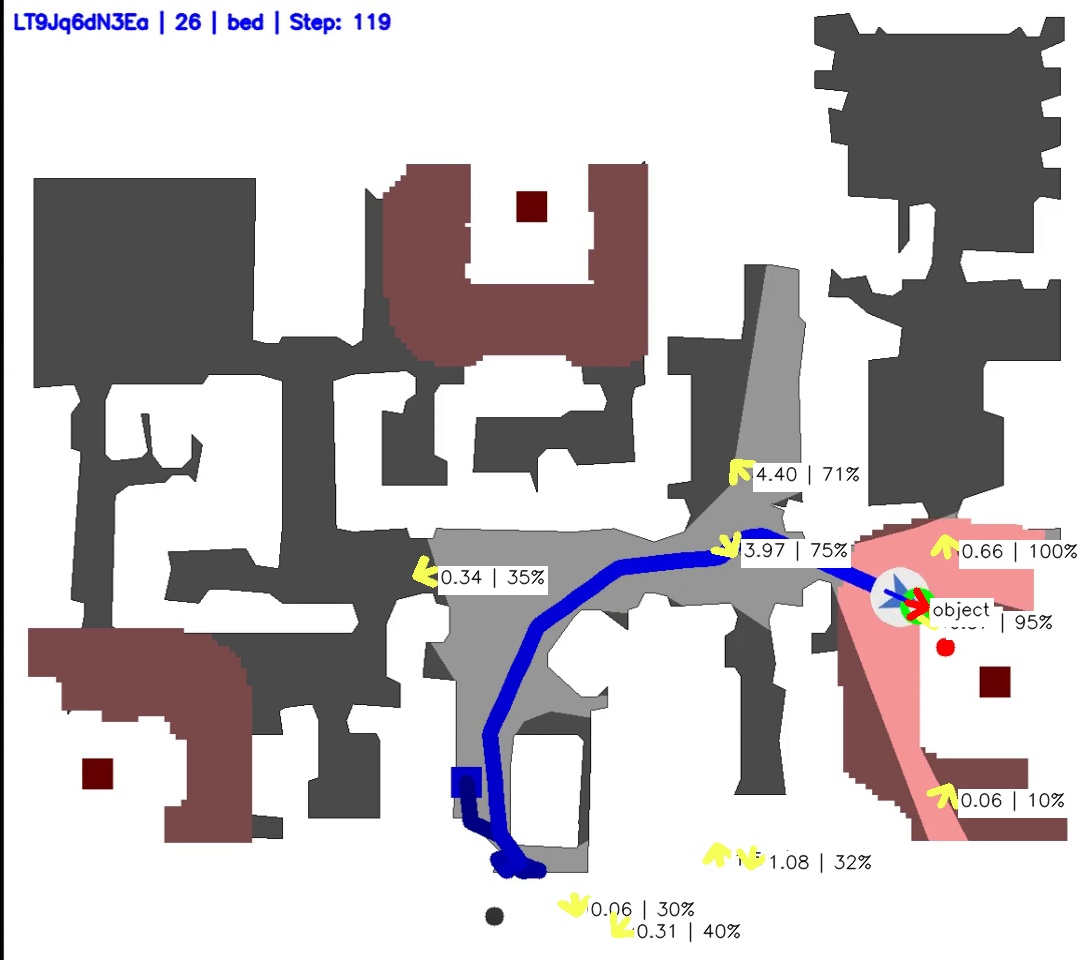}
        \caption{Scene: LT9Jq6dN3Ea, Target: Bed}
    \end{subfigure}
    \hfill
    \begin{subfigure}[t]{0.32\textwidth}
        \centering
        \includegraphics[width=\linewidth]{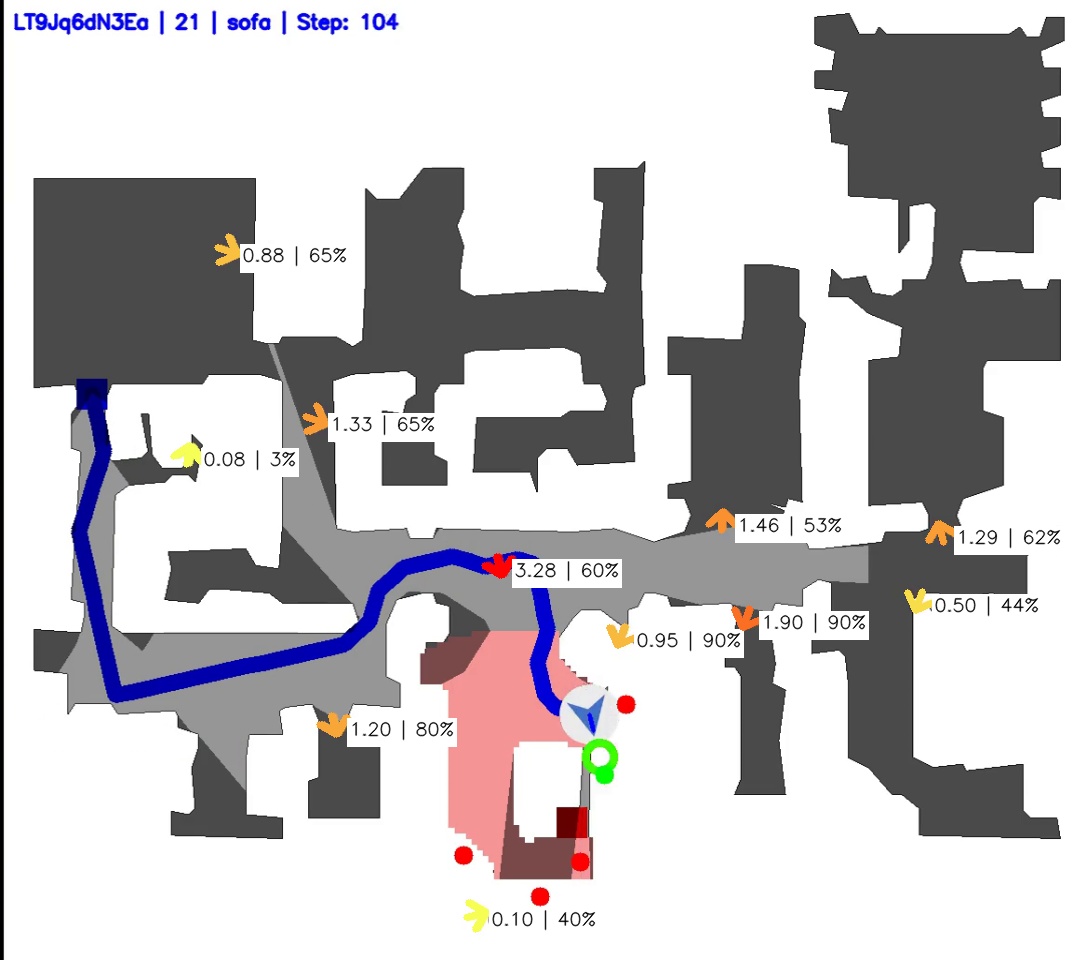}
        \caption{Scene: LT9Jq6dN3Ea, Target: Sofa}
    \end{subfigure}
    \hfill
    \begin{subfigure}[t]{0.32\textwidth}
        \centering
        \includegraphics[width=\linewidth]{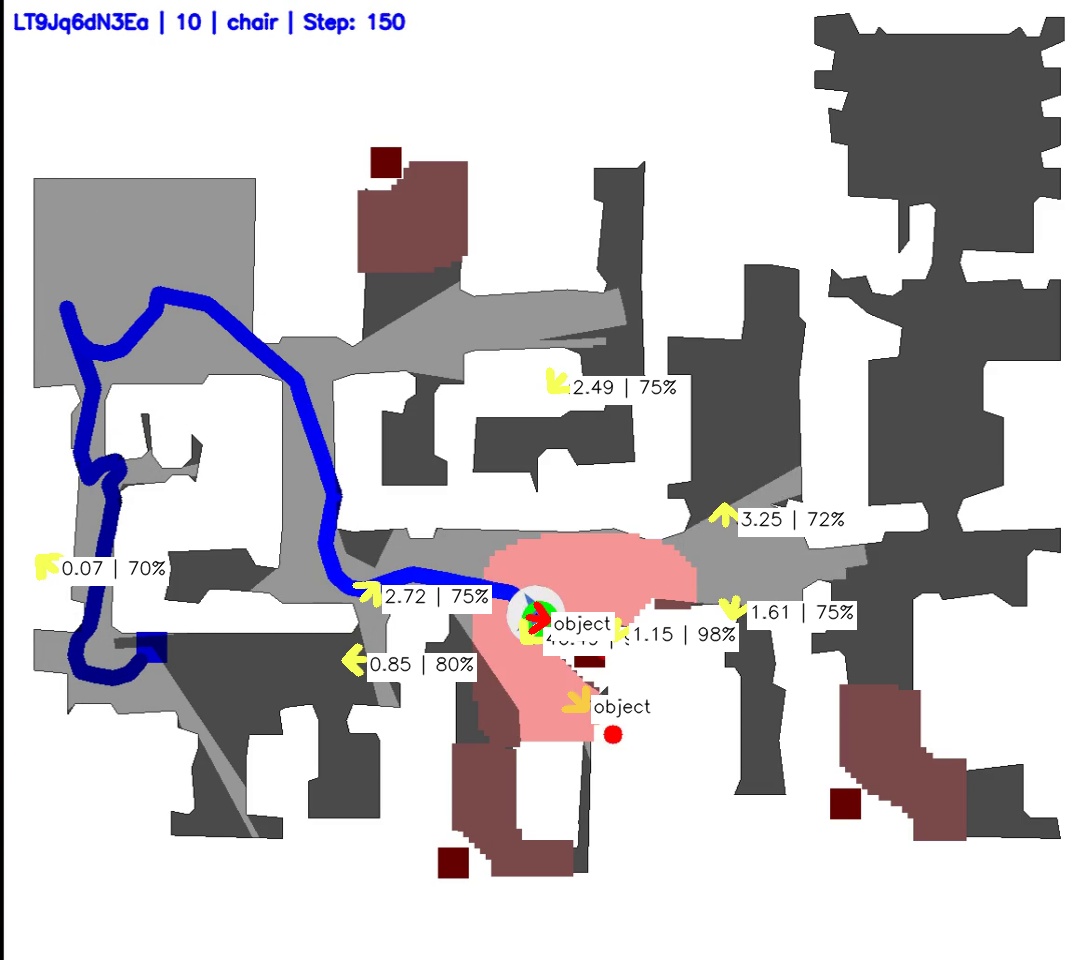}
        \caption{Scene: LT9Jq6dN3Ea, Target: Chair}
    \end{subfigure}

    \vspace{0.6em}

    \begin{subfigure}[t]{0.32\textwidth}
        \centering
        \includegraphics[width=\linewidth]{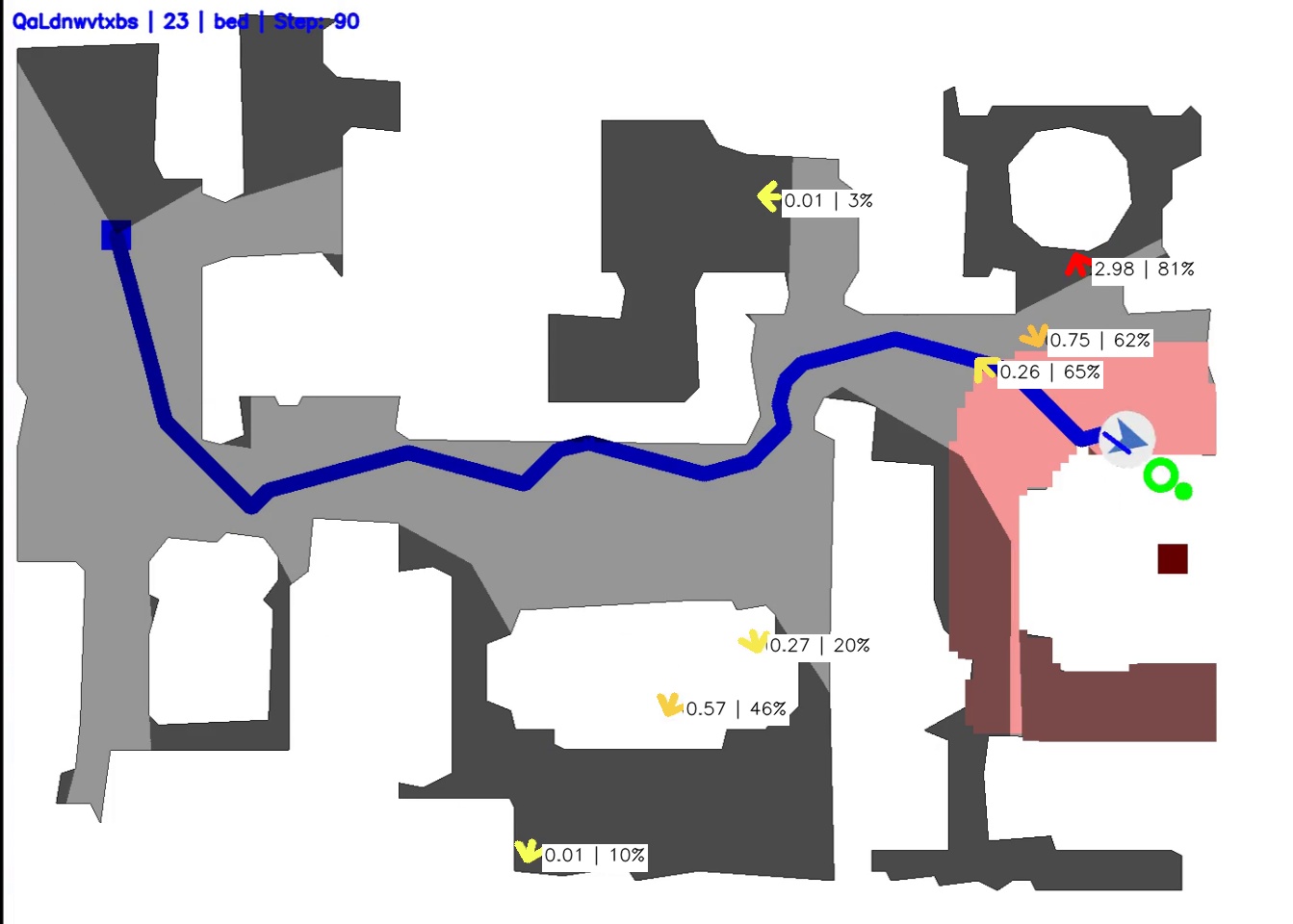}
        \caption{Scene: QaLdnwvtxbs, Targe: Bed}
    \end{subfigure}
    \hfill
    \begin{subfigure}[t]{0.32\textwidth}
        \centering
        \includegraphics[width=\linewidth]{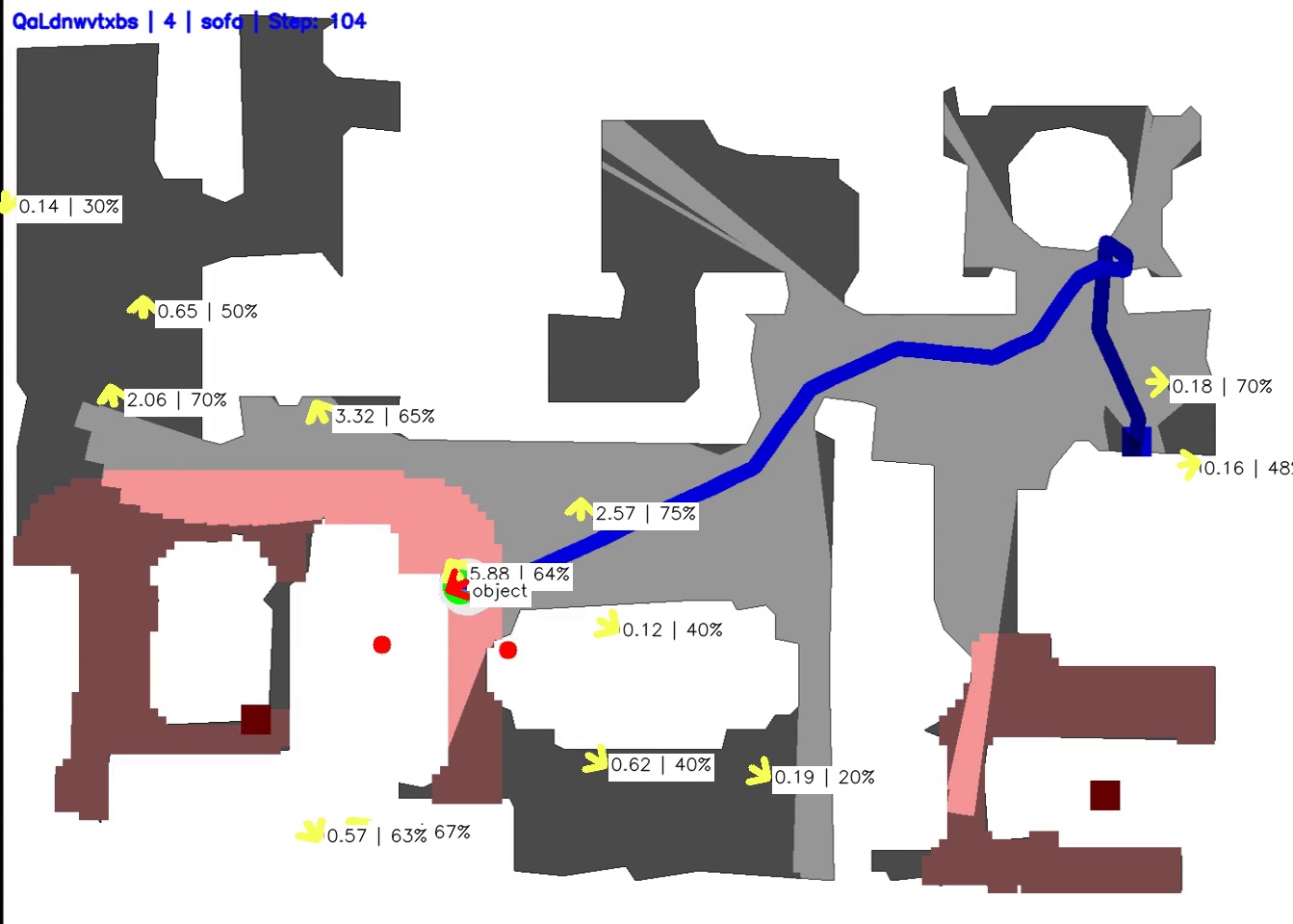}
        \caption{Scene: QaLdnwvtxbs, Targe: Sofa}
    \end{subfigure}
    \hfill
    \begin{subfigure}[t]{0.32\textwidth}
        \centering
        \includegraphics[width=\linewidth]{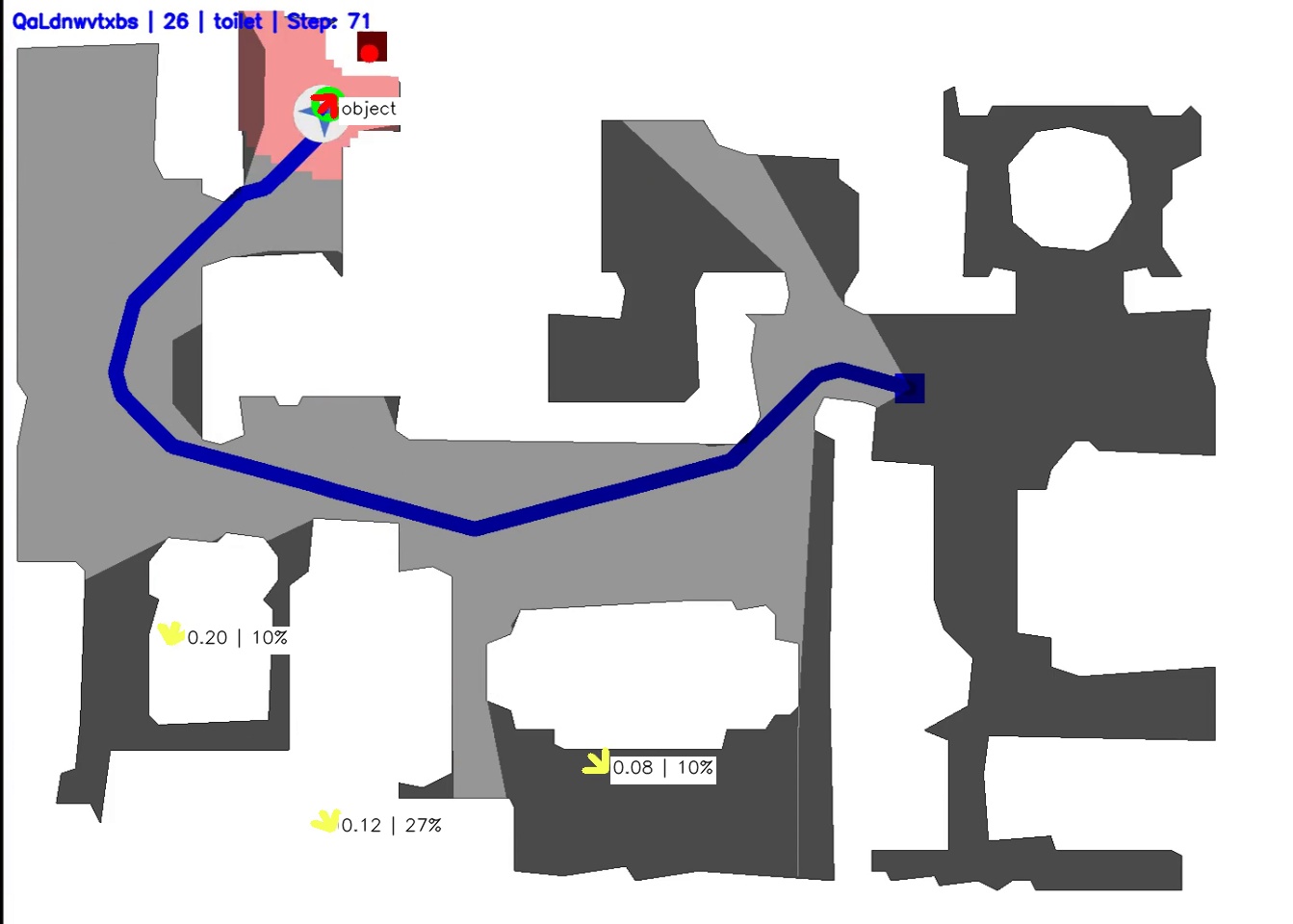}
        \caption{Scene: QaLdnwvtxbs, Targe: Toilet}
    \end{subfigure}

    \vspace{0.6em}

    \begin{subfigure}[t]{0.32\textwidth}
        \centering
        \includegraphics[width=\linewidth]{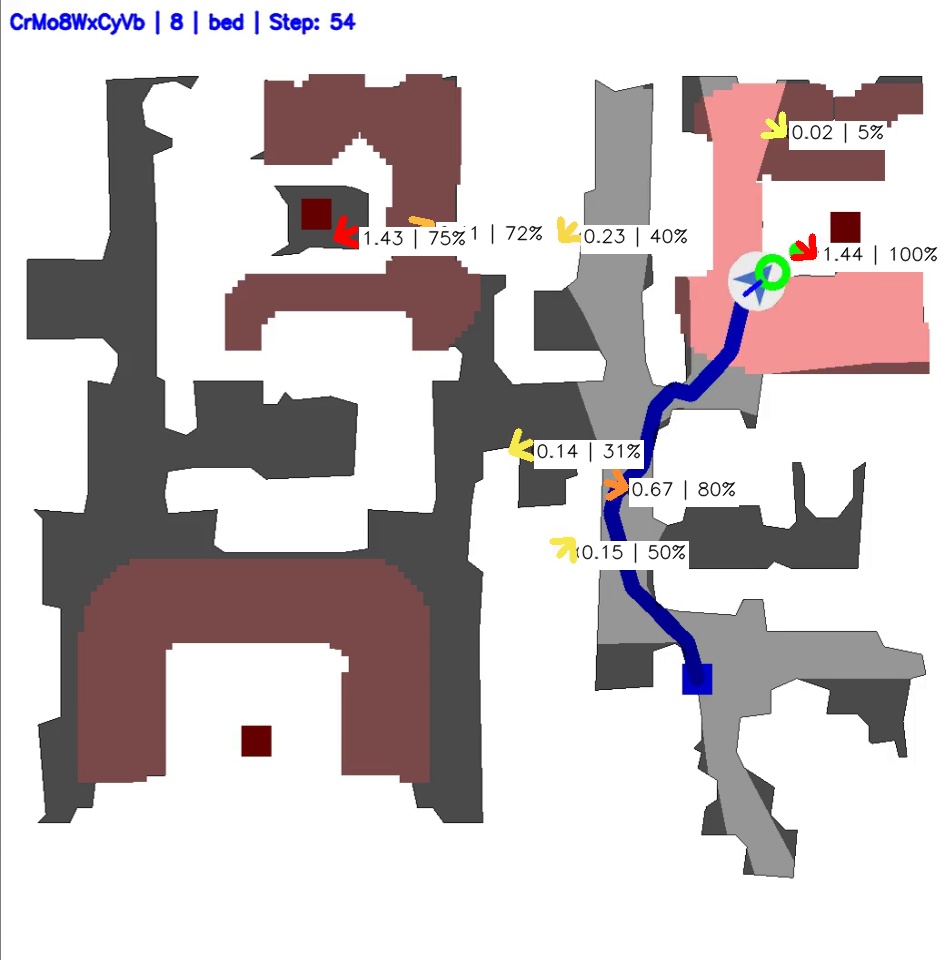}
        \caption{Scene: CrMo8WxCyVb, Target: Bed}
    \end{subfigure}
    \hfill
    \begin{subfigure}[t]{0.32\textwidth}
        \centering
        \includegraphics[width=\linewidth]{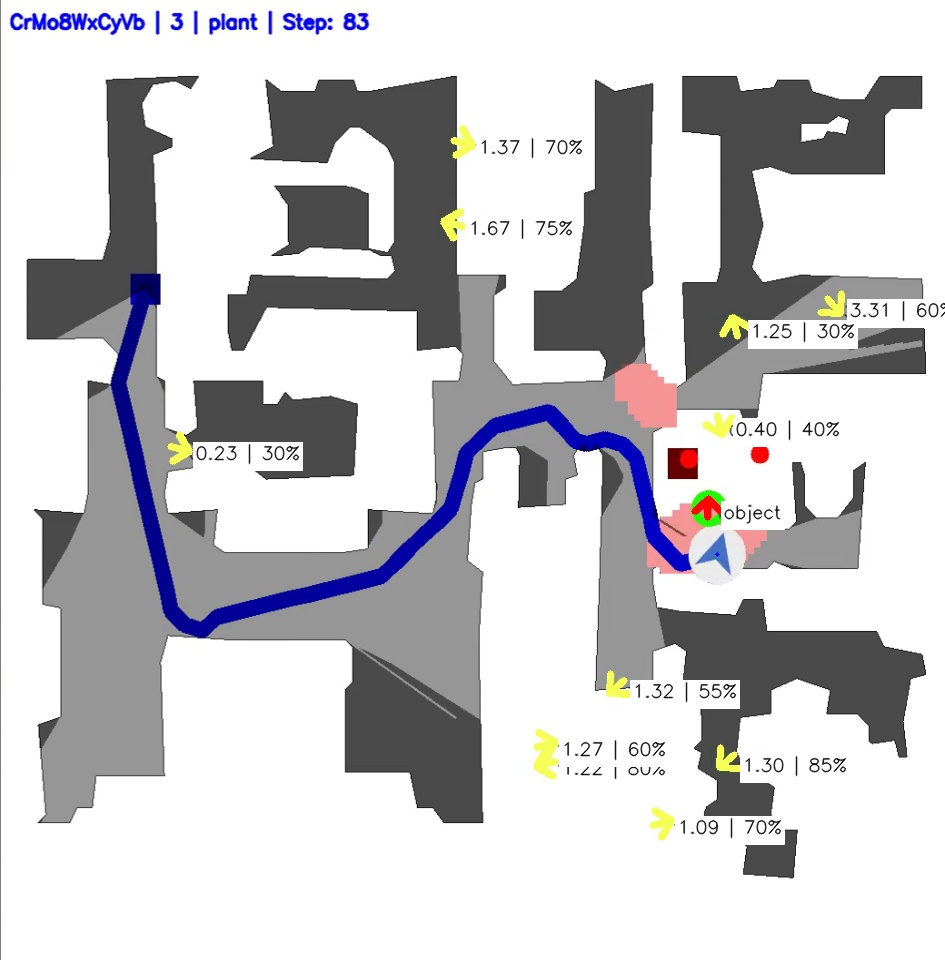}
        \caption{Scene: CrMo8WxCyVb, Target: Plant}
    \end{subfigure}
    \hfill
    \begin{subfigure}[t]{0.32\textwidth}
        \centering
        \includegraphics[width=\linewidth]{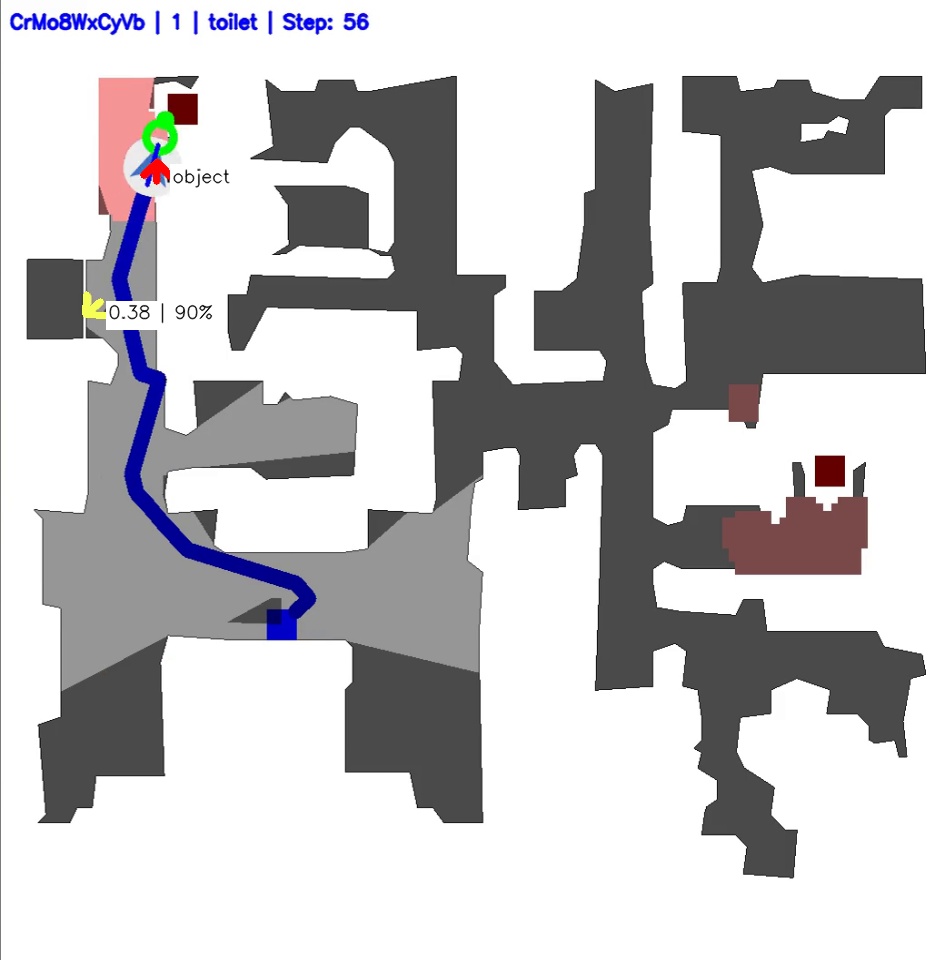}
        \caption{Scene: CrMo8WxCyVb, Target: Toilet}
    \end{subfigure}

\caption{\textbf{Additional Navigation Examples} in HM3D across four scenes and three target objects.}
    \label{fig:qualitative_results_hm3d}
\end{figure*}

\begin{figure*}[t]
    \centering

    \begin{subfigure}[t]{0.32\textwidth}
        \centering
        \includegraphics[width=\linewidth]{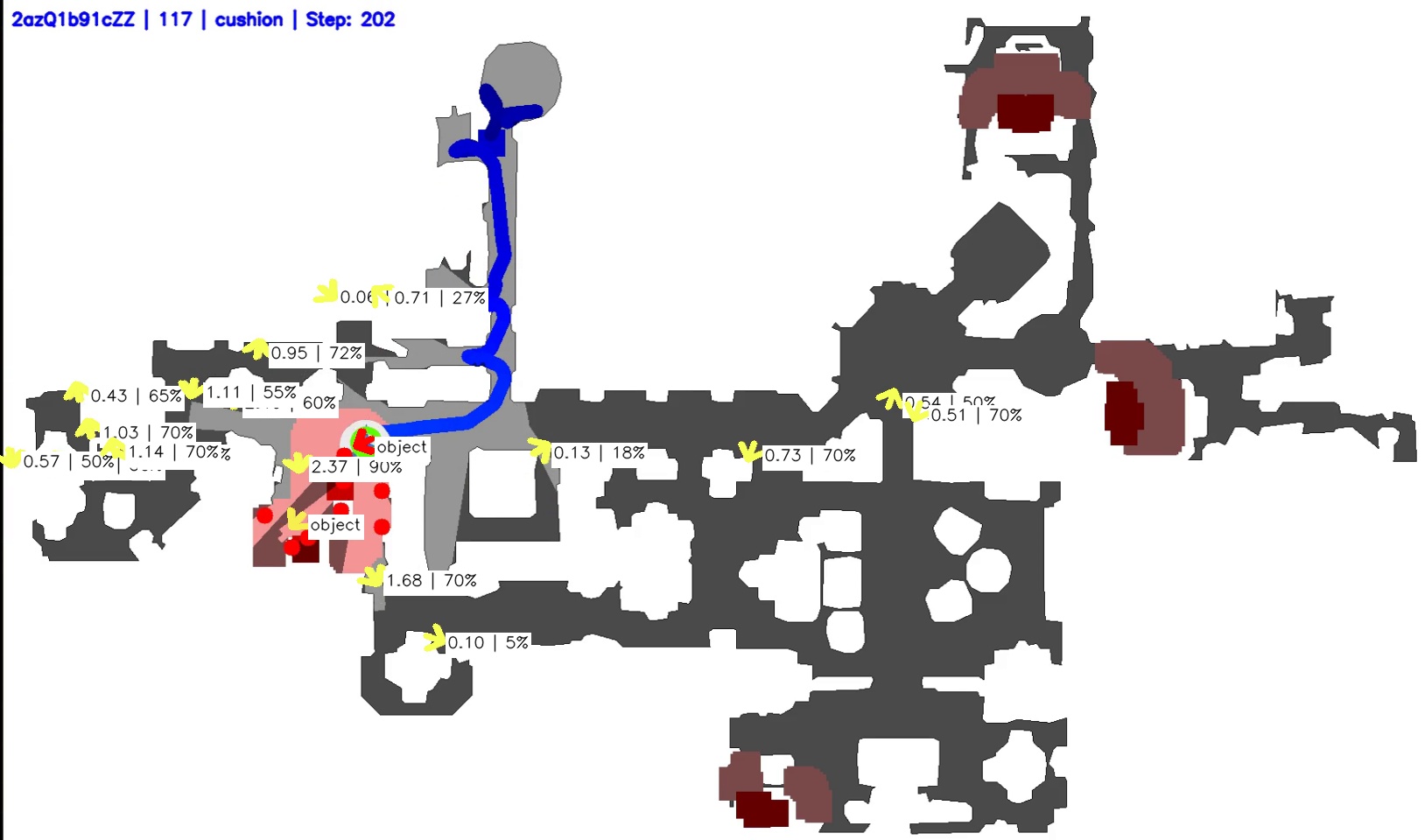}
        \caption{Scene: 2azQ1b91cZZ, Target: Cushion}
    \end{subfigure}
    \hfill
    \begin{subfigure}[t]{0.32\textwidth}
        \centering
        \includegraphics[width=\linewidth]{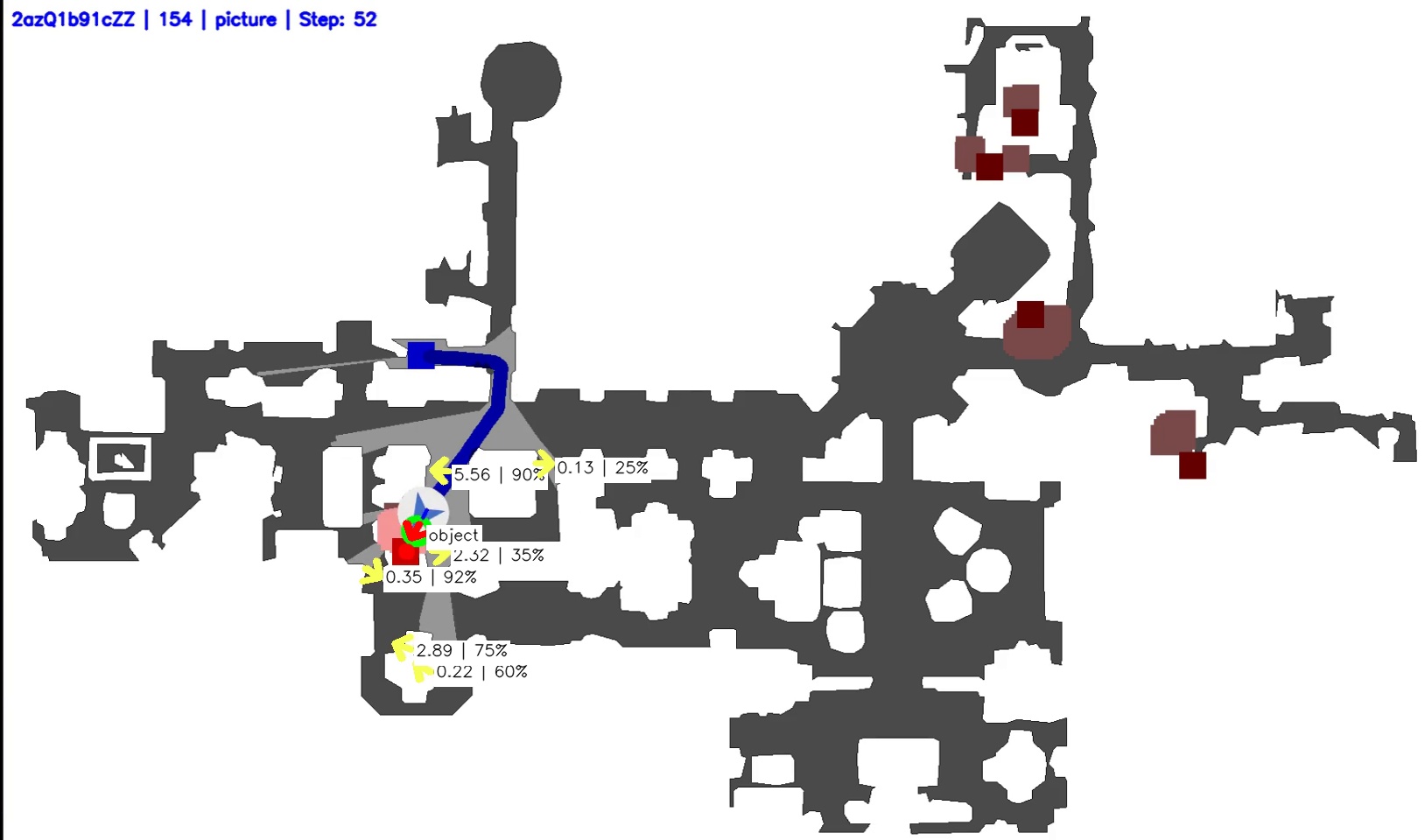}
        \caption{Scene: 2azQ1b91cZZ, Target: Picture}
    \end{subfigure}
    \hfill
    \begin{subfigure}[t]{0.32\textwidth}
        \centering
        \includegraphics[width=\linewidth]{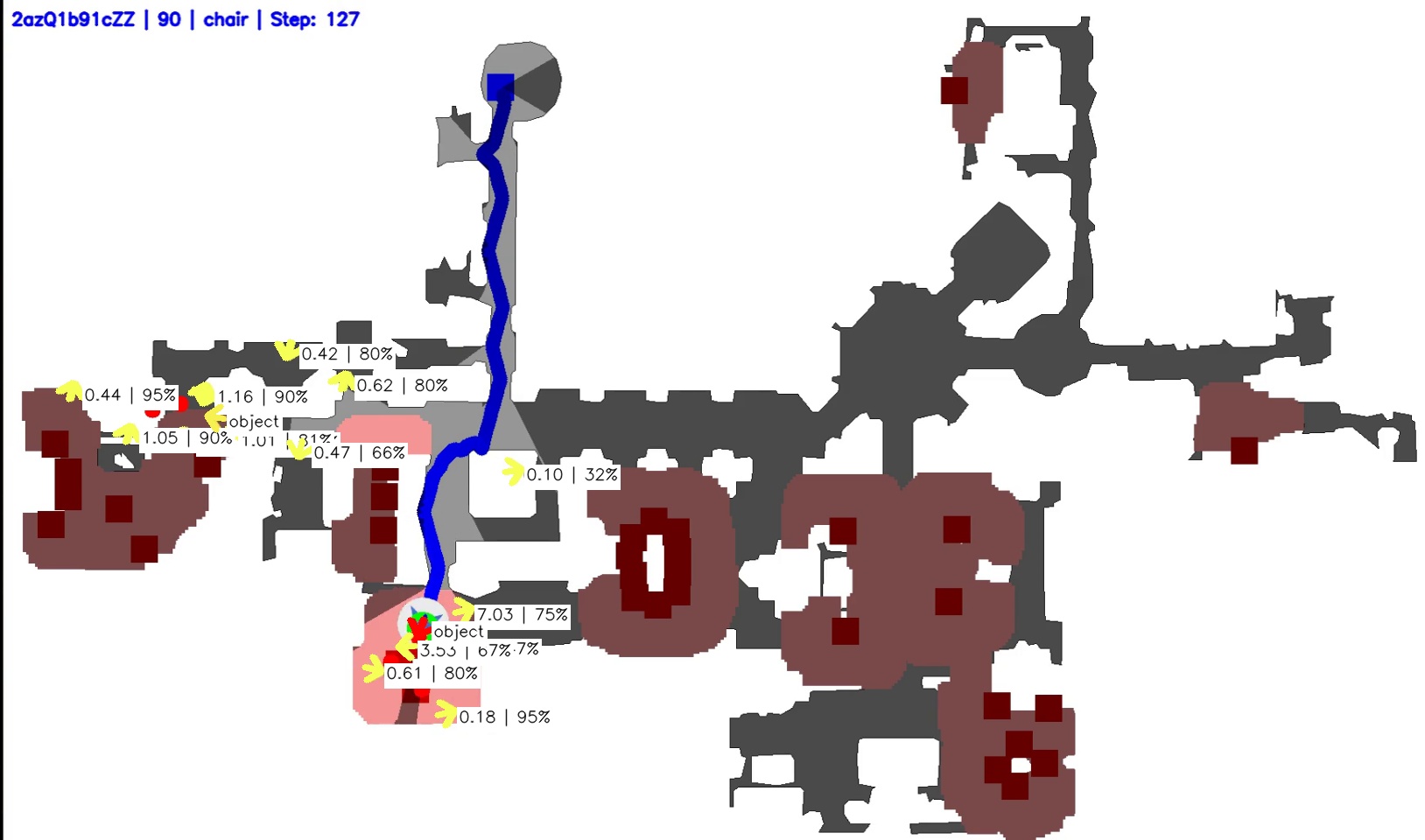}
        \caption{Scene: 2azQ1b91cZZ, Target: Chair}
    \end{subfigure}

    \vspace{0.6em}

    \begin{subfigure}[t]{0.32\textwidth}
        \centering
        \includegraphics[width=\linewidth]{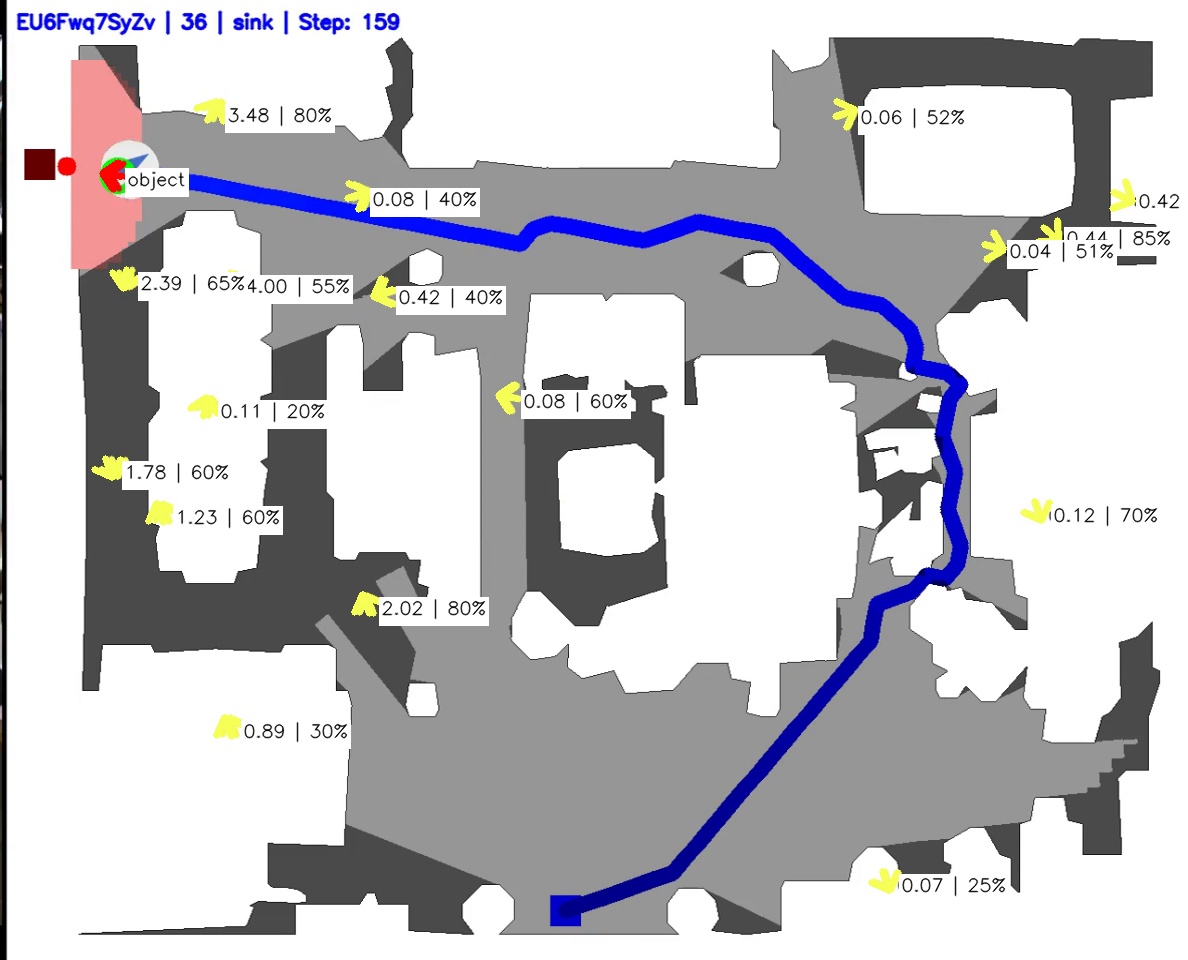}
        \caption{Scene: EU6Fwq7SyZv, Target: Sink}
    \end{subfigure}
    \hfill
    \begin{subfigure}[t]{0.32\textwidth}
        \centering
        \includegraphics[width=\linewidth]{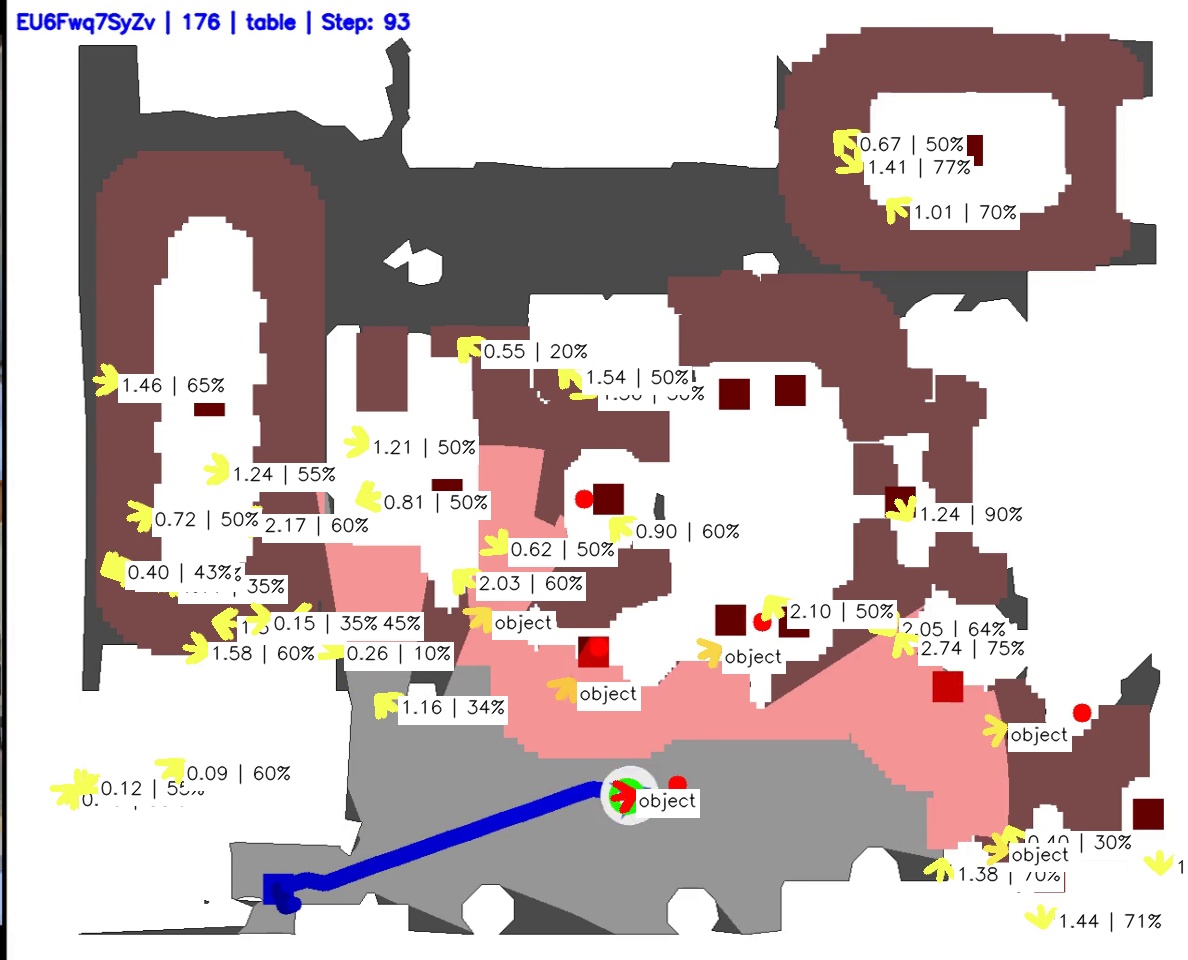}
        \caption{Scene: EU6Fwq7SyZv, Target: Table}
    \end{subfigure}
    \hfill
    \begin{subfigure}[t]{0.32\textwidth}
        \centering
        \includegraphics[width=\linewidth]{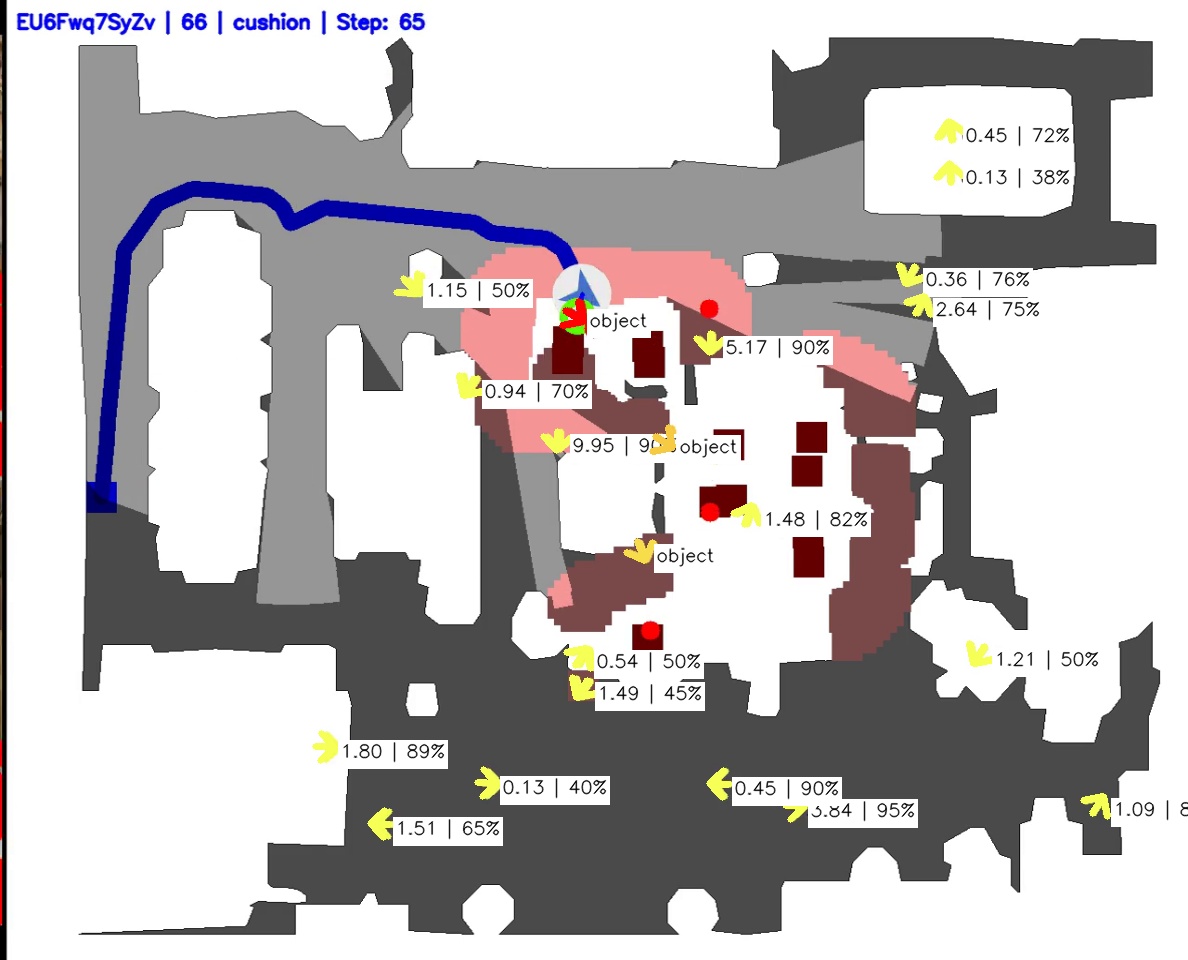}
        \caption{Scene: EU6Fwq7SyZv, Target: Cushion}
    \end{subfigure}

    \vspace{0.6em}

    \begin{subfigure}[t]{0.32\textwidth}
        \centering
        \includegraphics[width=\linewidth]{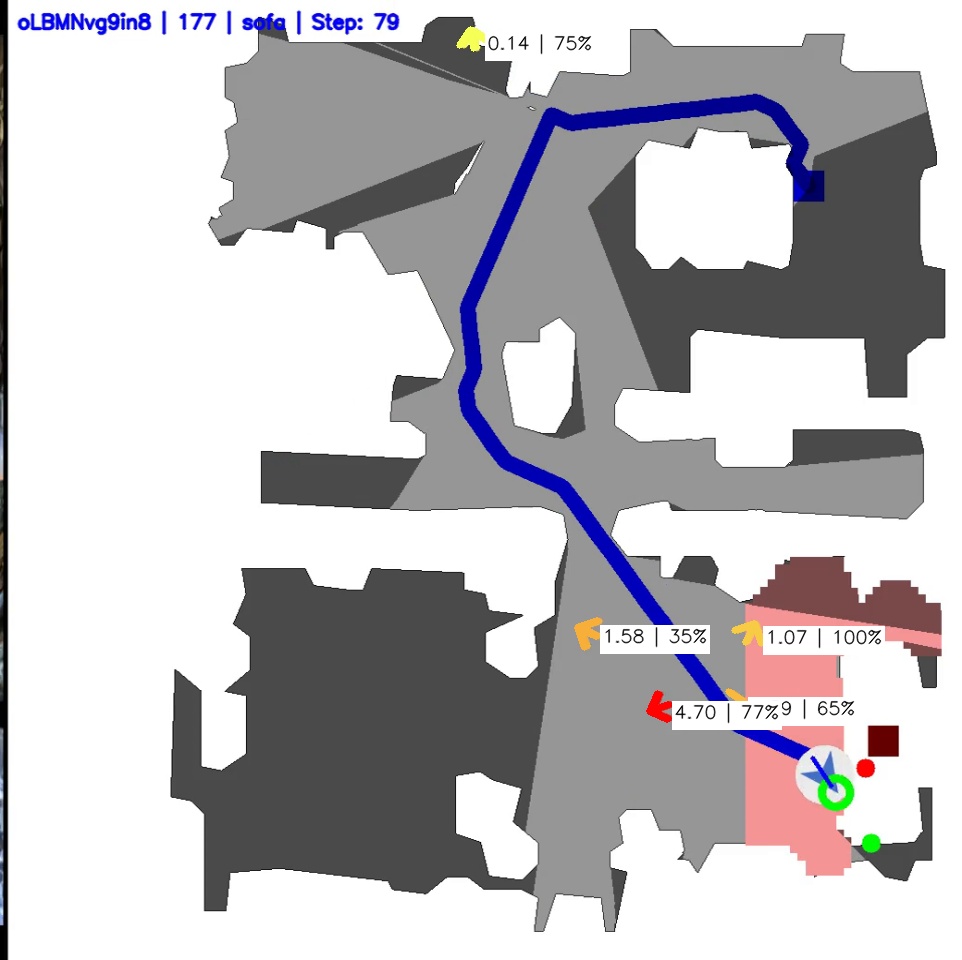}
        \caption{Scene: oLBMNvg9in8, Targe: Sofa}
    \end{subfigure}
    \hfill
    \begin{subfigure}[t]{0.32\textwidth}
        \centering
        \includegraphics[width=\linewidth]{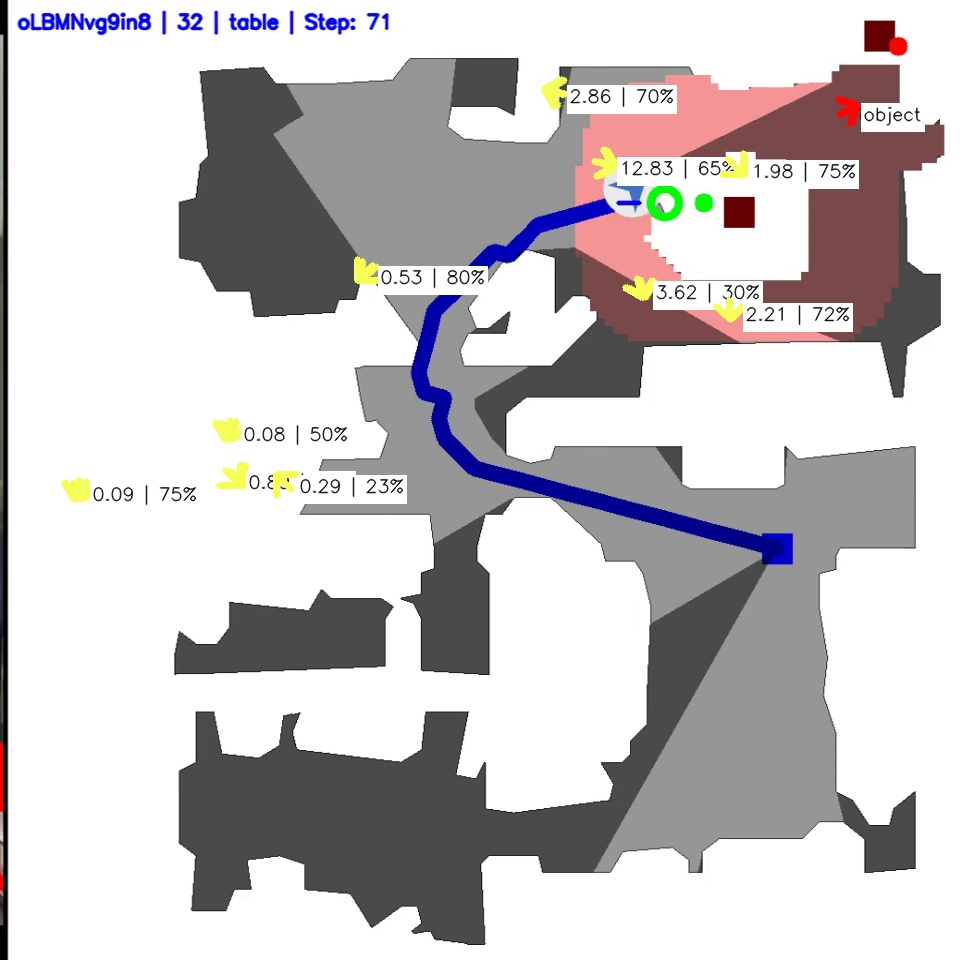}
        \caption{Scene: oLBMNvg9in8, Targe: Table}
    \end{subfigure}
    \hfill
    \begin{subfigure}[t]{0.32\textwidth}
        \centering
        \includegraphics[width=\linewidth]{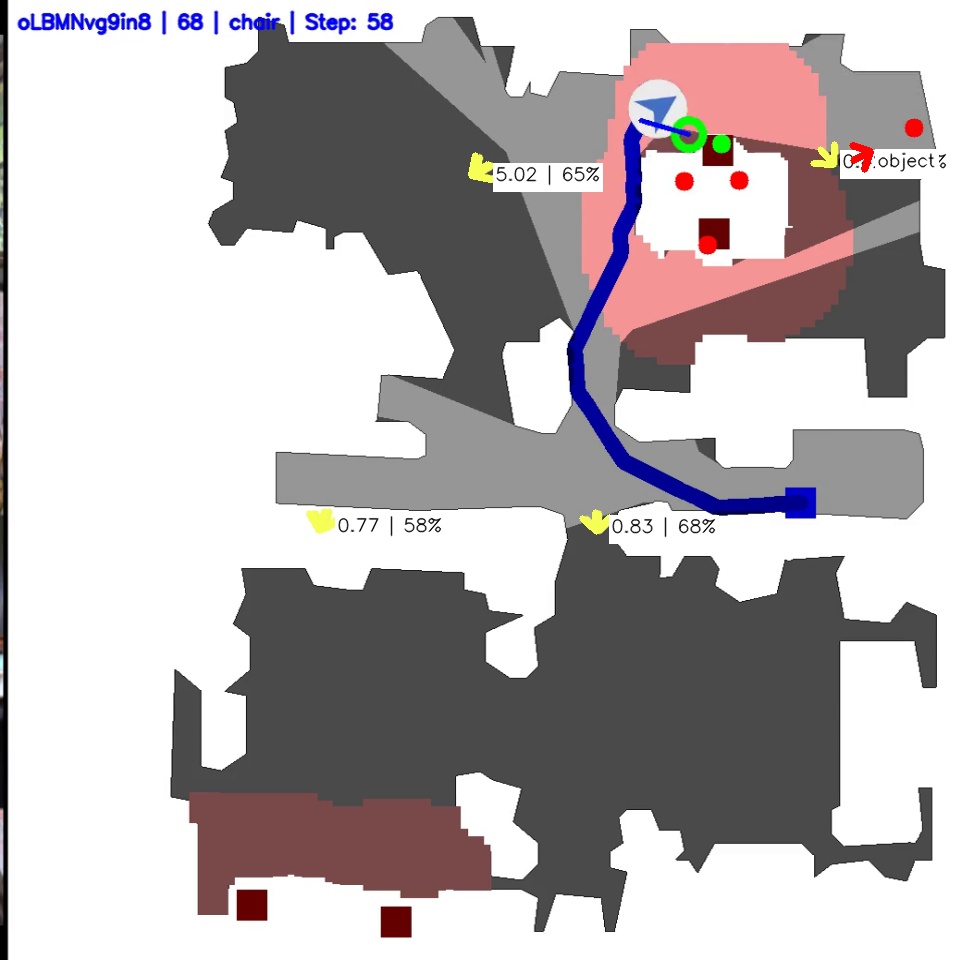}
        \caption{Scene: oLBMNvg9in8, Targe: Chair}
    \end{subfigure}

    \vspace{0.6em}

    \begin{subfigure}[t]{0.32\textwidth}
        \centering
        \includegraphics[width=\linewidth]{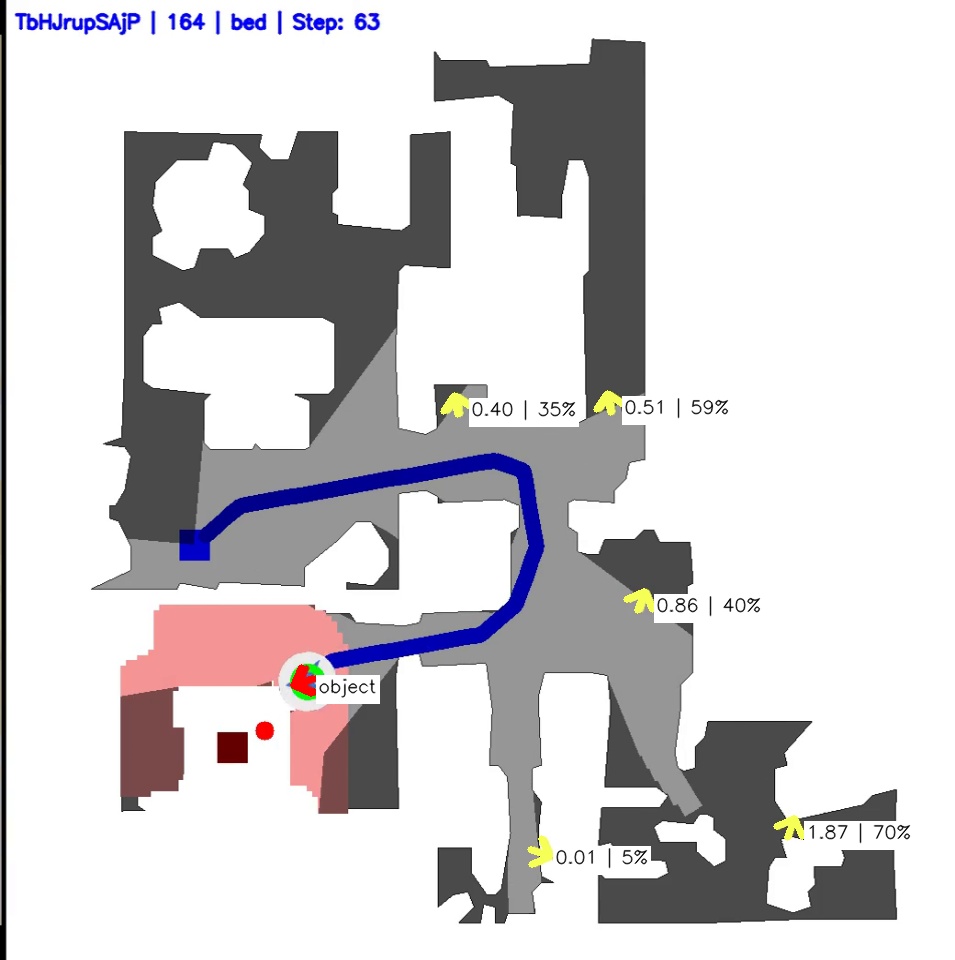}
        \caption{Scene: TbHJrupSAjP, Target: Bed}
    \end{subfigure}
    \hfill
    \begin{subfigure}[t]{0.32\textwidth}
        \centering
        \includegraphics[width=\linewidth]{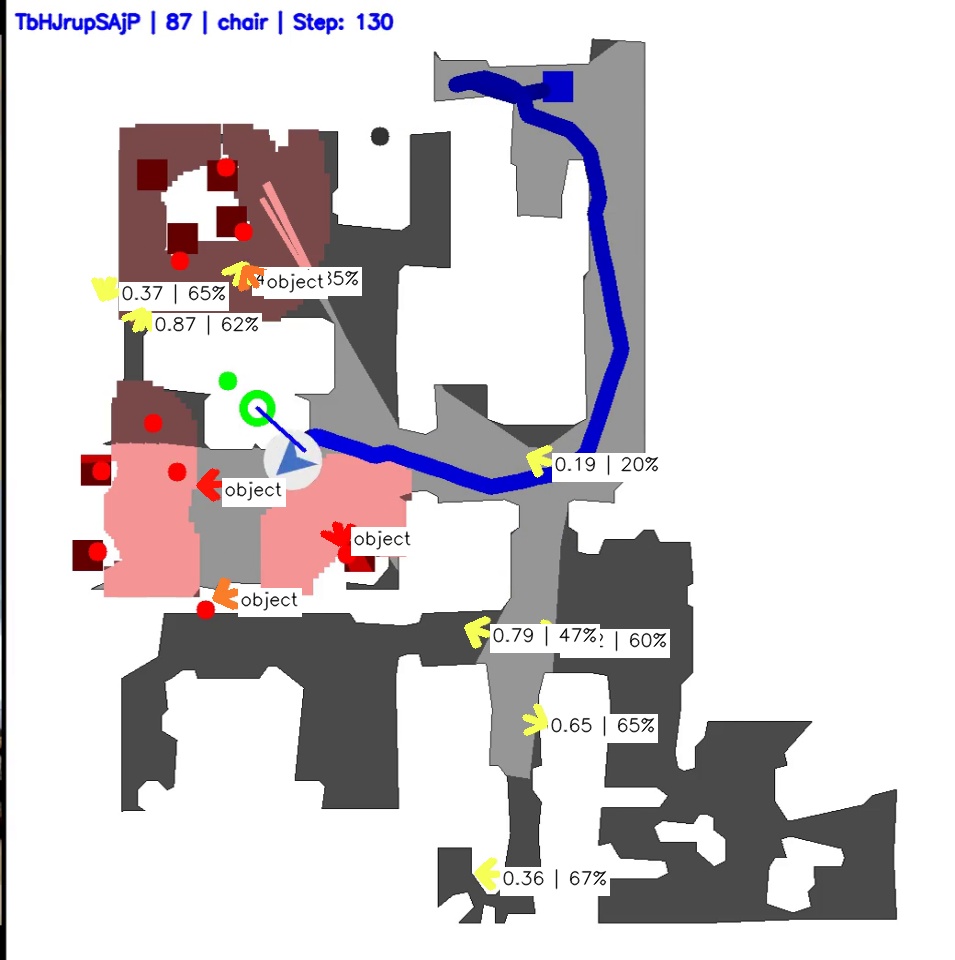}
        \caption{Scene: TbHJrupSAjP, Target: Chair}
    \end{subfigure}
    \hfill
    \begin{subfigure}[t]{0.32\textwidth}
        \centering
        \includegraphics[width=\linewidth]{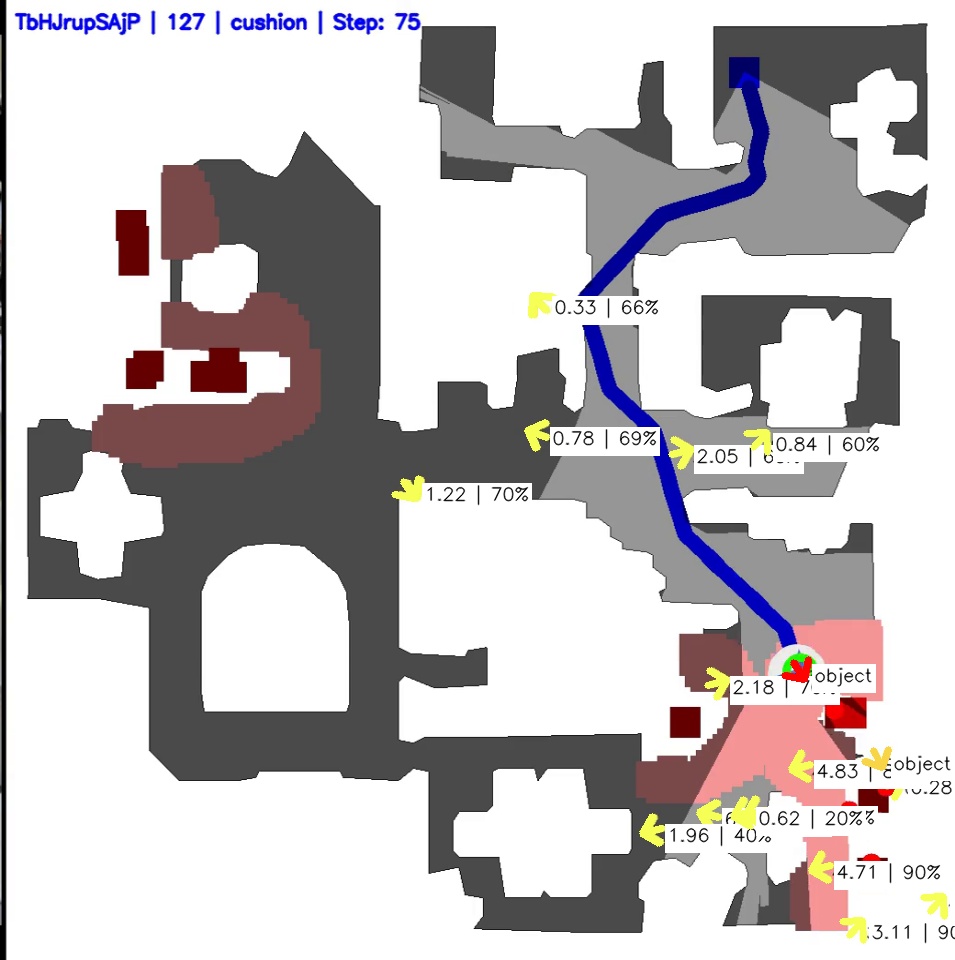}
        \caption{Scene: TbHJrupSAjP, Target: Cushion}
    \end{subfigure}

\caption{\textbf{Additional Navigation Examples} in MP3D across four scenes and three target objects.}
    \label{fig:qualitative_results_mp3d}
\end{figure*}

\begin{figure*}[t]
    \centering

    \begin{subfigure}[t]{0.32\textwidth}
        \centering
        \includegraphics[width=\linewidth]{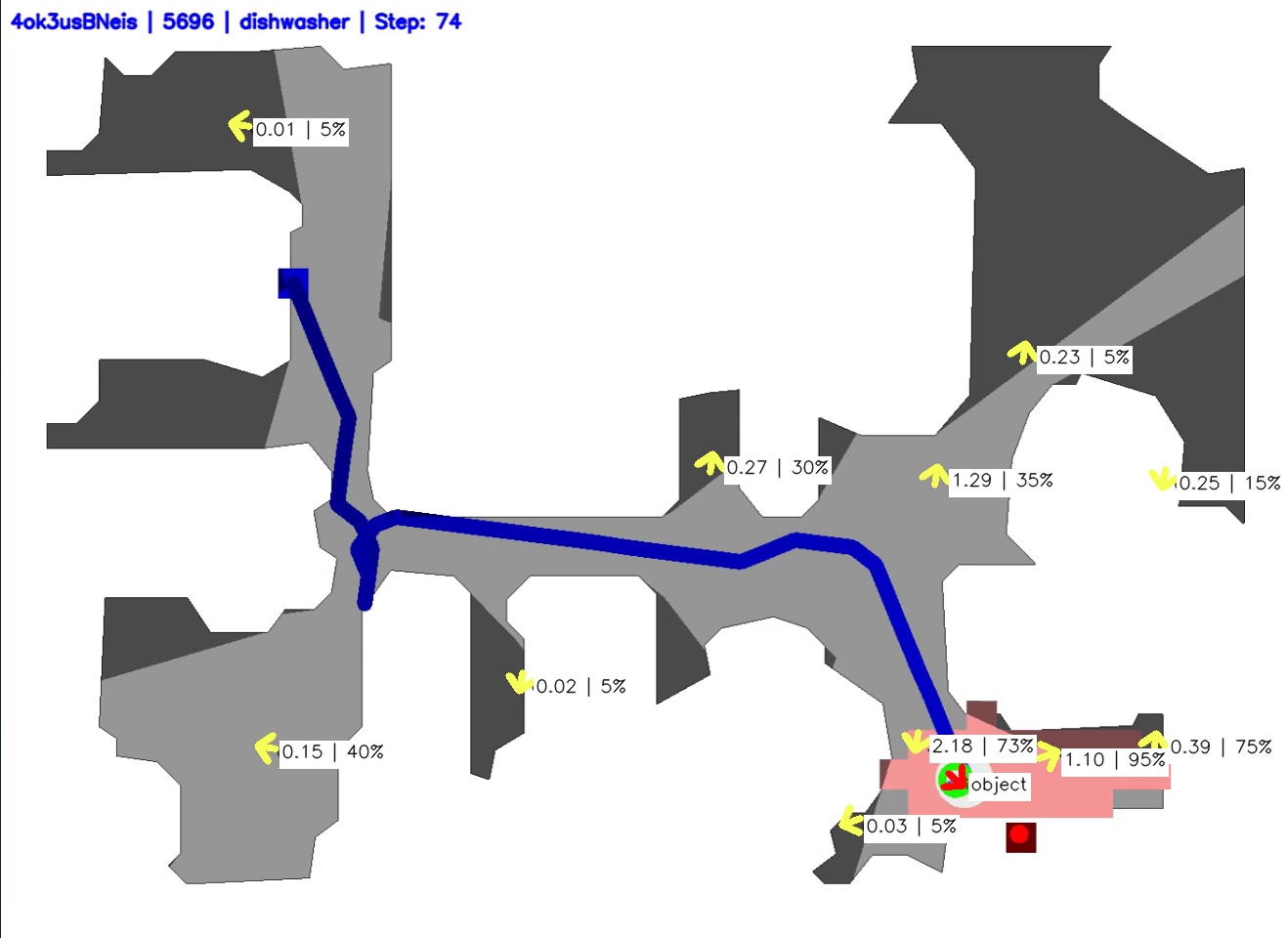}
        \caption{Scene: 4ok3usBNeis, Target: Dishwasher}
    \end{subfigure}
    \hfill
    \begin{subfigure}[t]{0.32\textwidth}
        \centering
        \includegraphics[width=\linewidth]{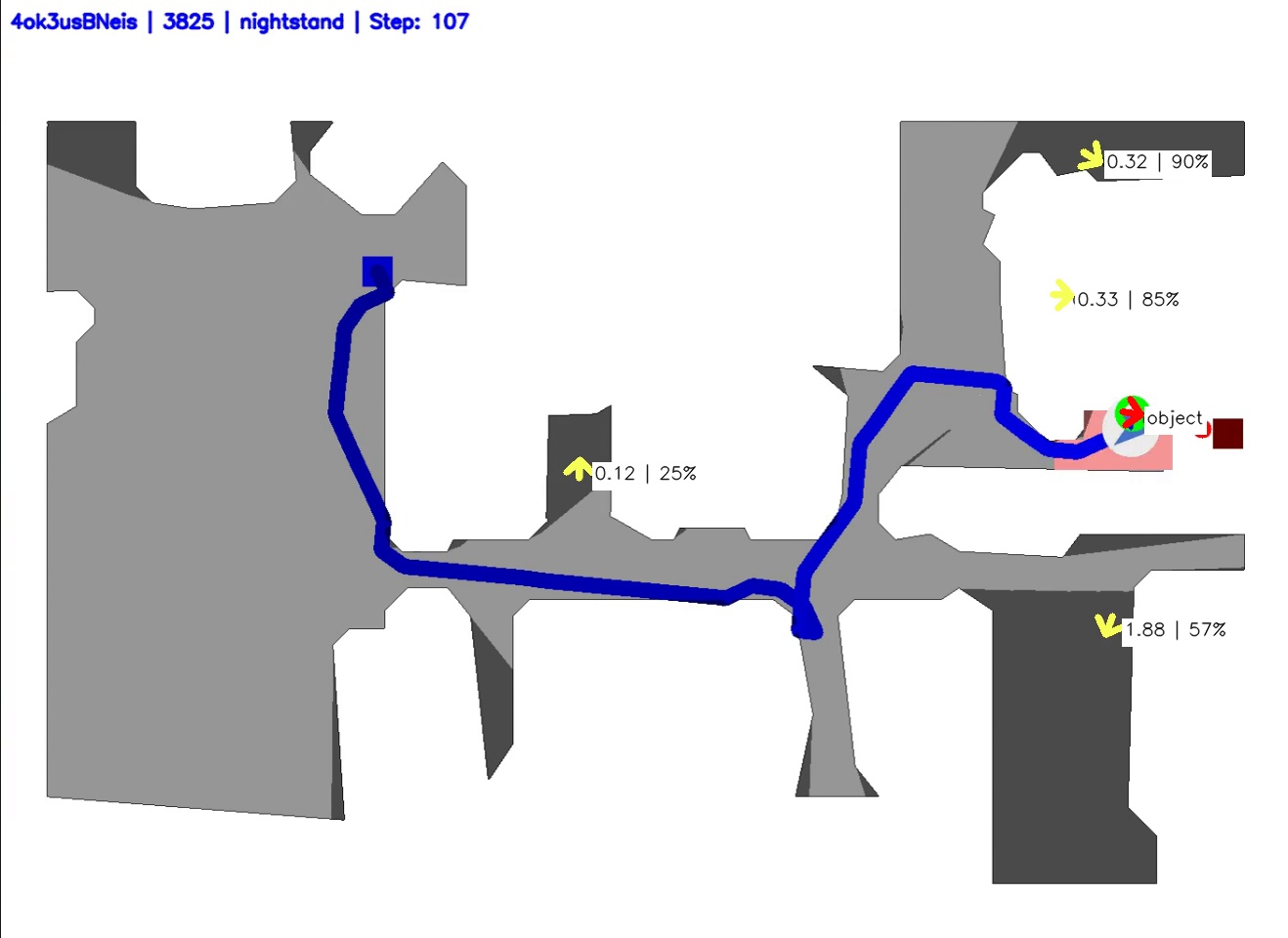}
        \caption{Scene: 4ok3usBNeis, Target: Nightstand}
    \end{subfigure}
    \hfill
    \begin{subfigure}[t]{0.32\textwidth}
        \centering
        \includegraphics[width=\linewidth]{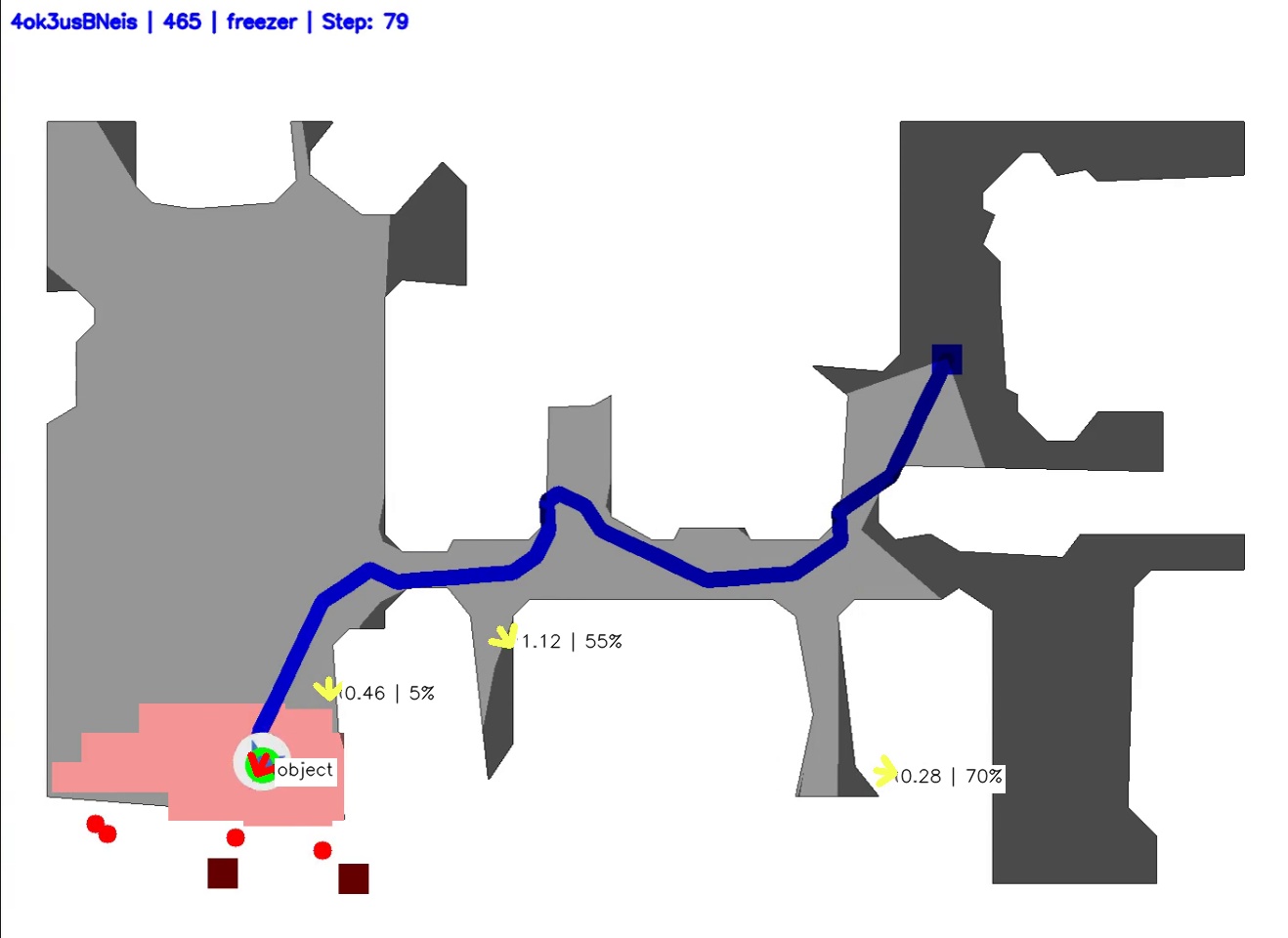}
        \caption{Scene: 4ok3usBNeis, Target: Freeze}
    \end{subfigure}

    \vspace{0.6em}

    \begin{subfigure}[t]{0.32\textwidth}
        \centering
        \includegraphics[width=\linewidth]{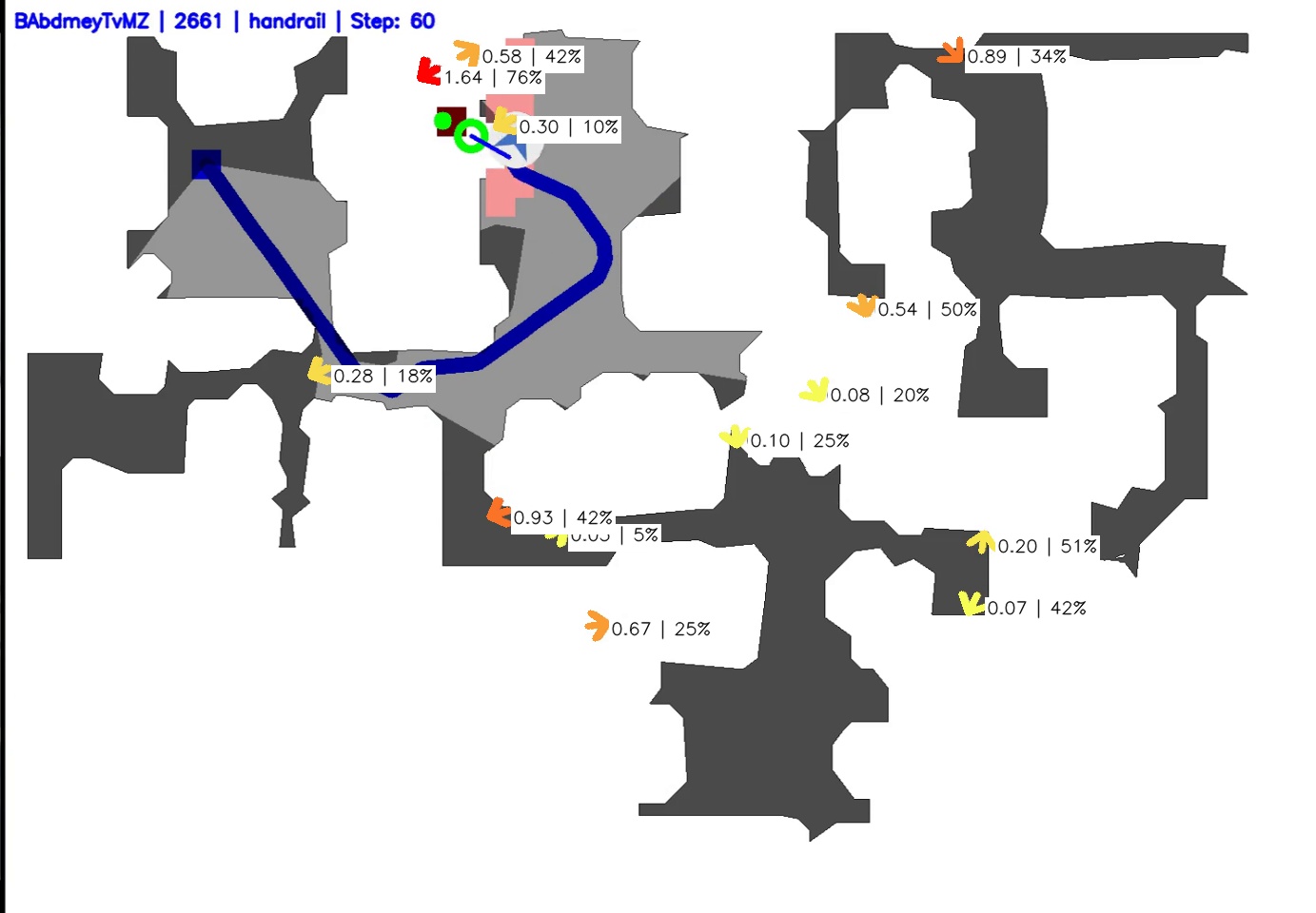}
        \caption{Scene: BAbdmeyTvMZ, Target: Handrail}
    \end{subfigure}
    \hfill
    \begin{subfigure}[t]{0.32\textwidth}
        \centering
        \includegraphics[width=\linewidth]{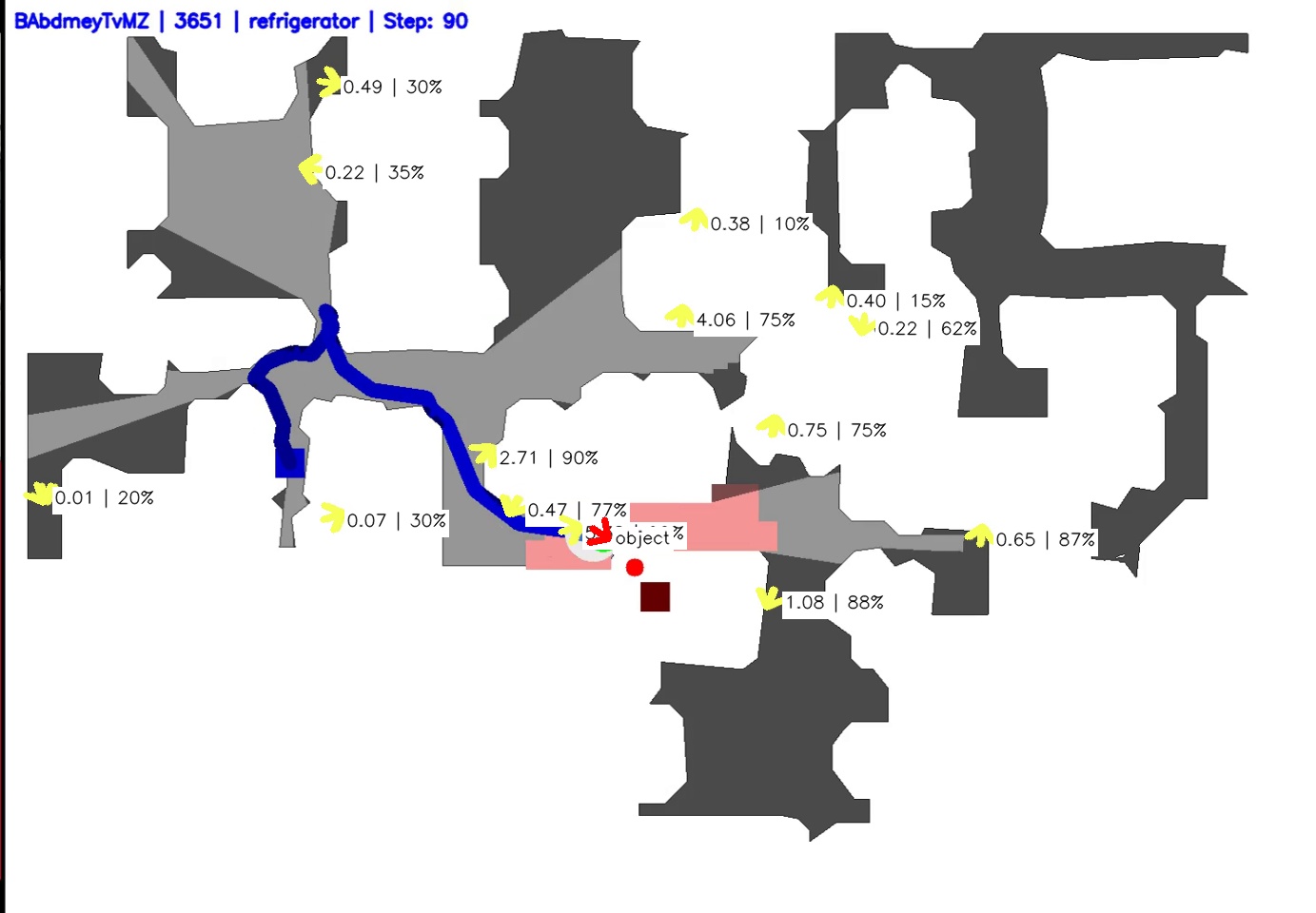}
        \caption{Scene: BAbdmeyTvMZ, Target: Refrigerator}
    \end{subfigure}
    \hfill
    \begin{subfigure}[t]{0.32\textwidth}
        \centering
        \includegraphics[width=\linewidth]{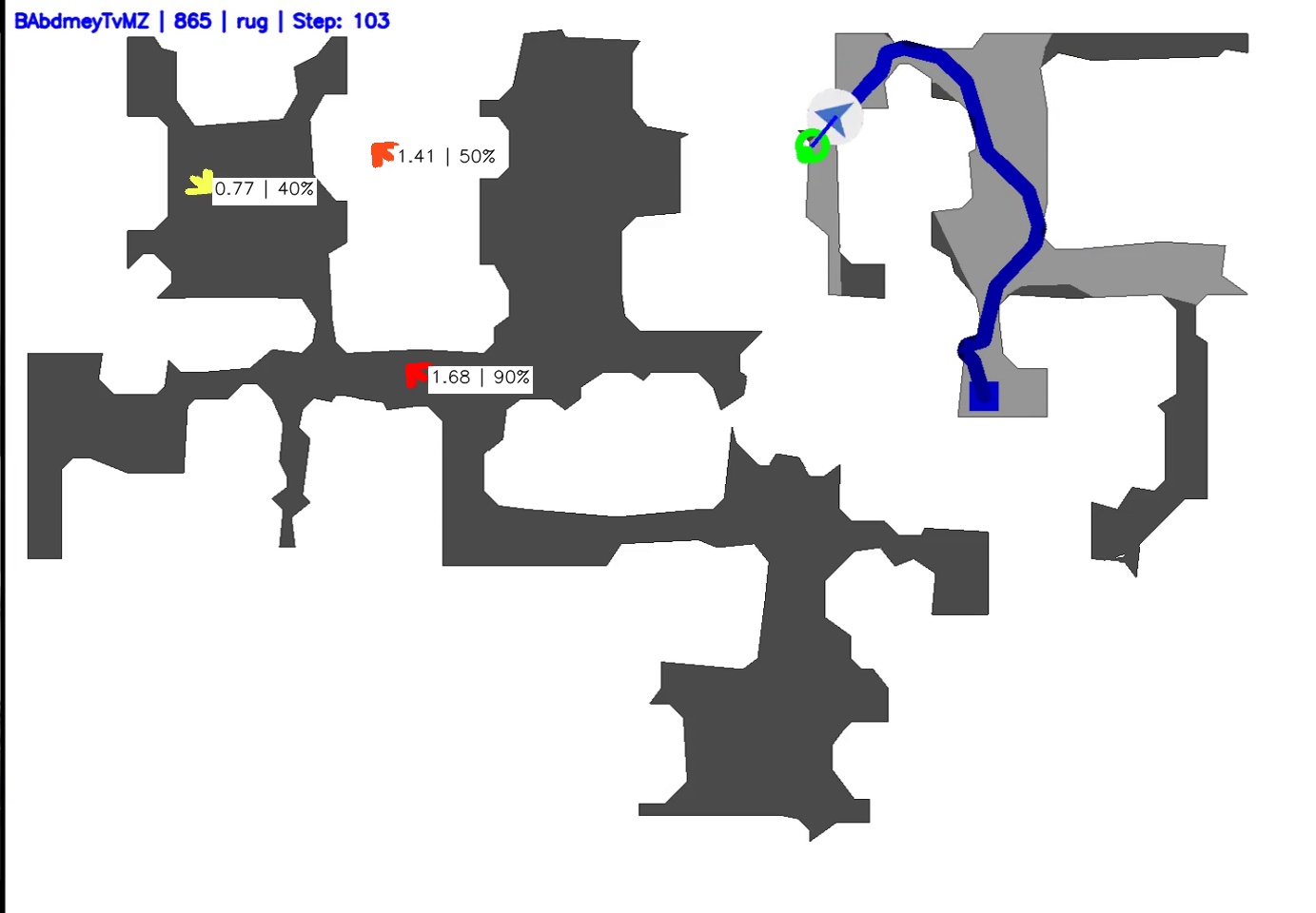}
        \caption{Scene: BAbdmeyTvMZ, Target: Rug}
    \end{subfigure}

    \vspace{0.6em}

    \begin{subfigure}[t]{0.32\textwidth}
        \centering
        \includegraphics[width=\linewidth]{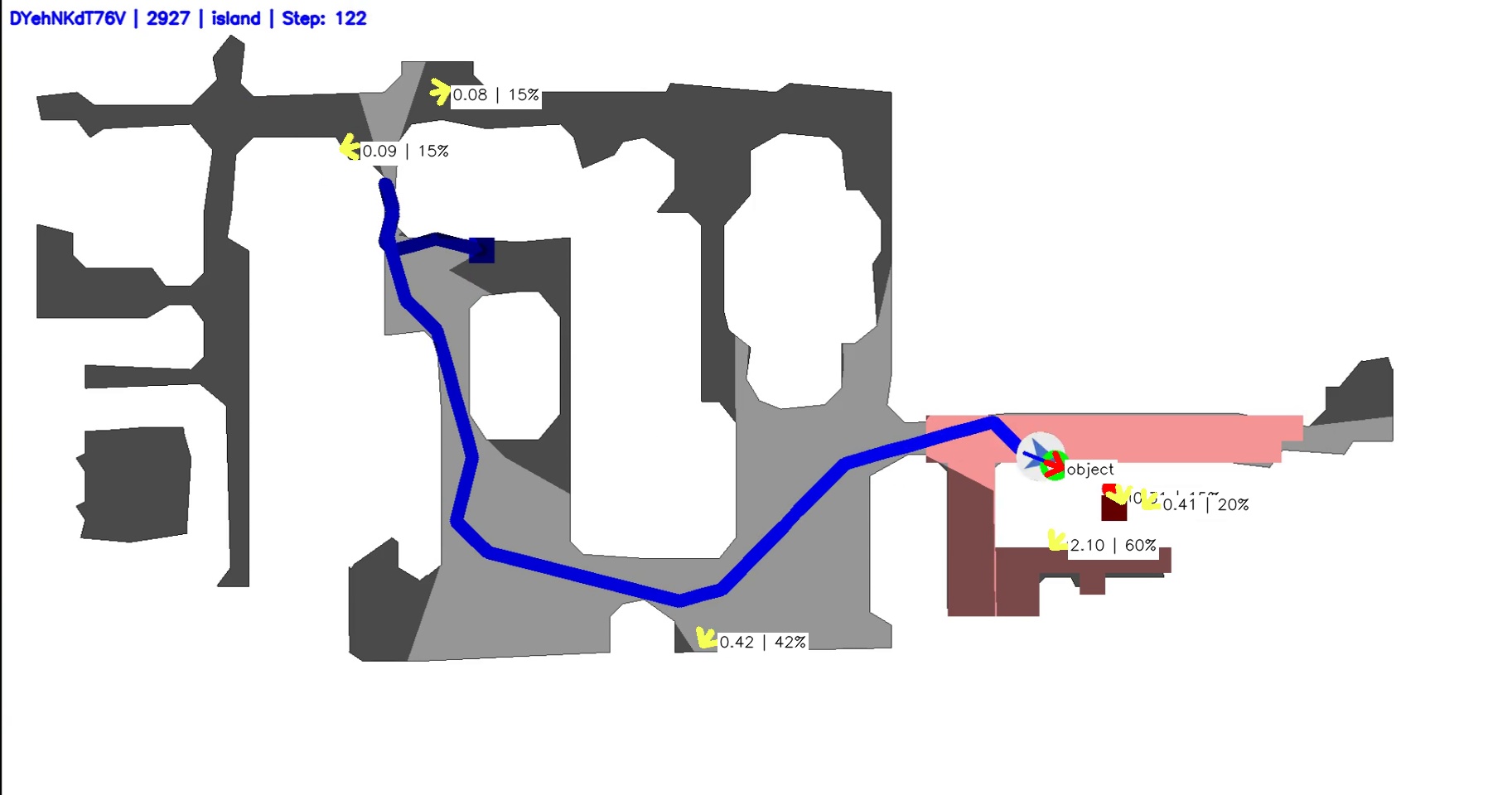}
        \caption{Scene: DYehNKdT76V, Targe: Island}
    \end{subfigure}
    \hfill
    \begin{subfigure}[t]{0.32\textwidth}
        \centering
        \includegraphics[width=\linewidth]{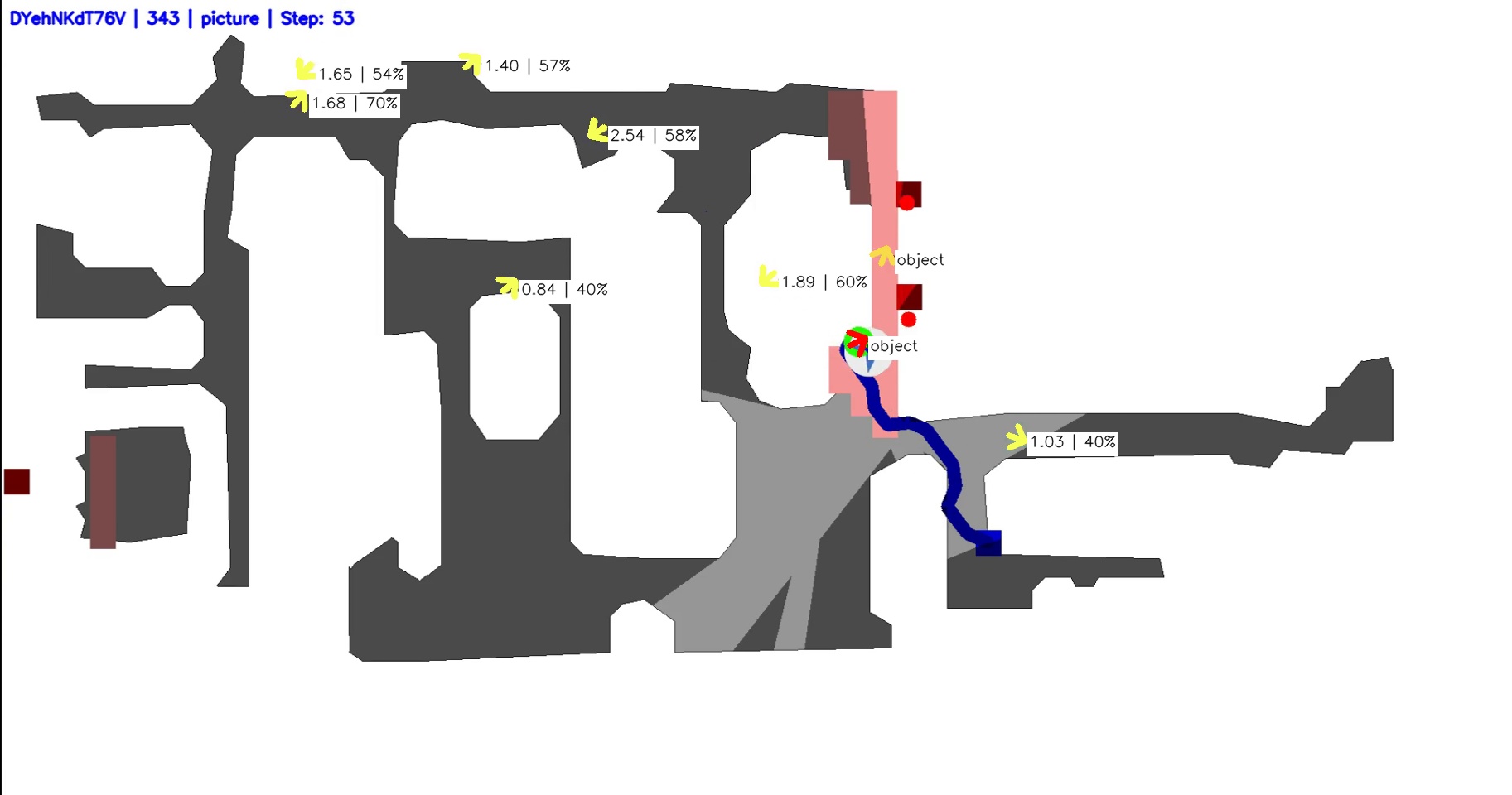}
        \caption{Scene: DYehNKdT76V, Targe: Picture}
    \end{subfigure}
    \hfill
    \begin{subfigure}[t]{0.32\textwidth}
        \centering
        \includegraphics[width=\linewidth]{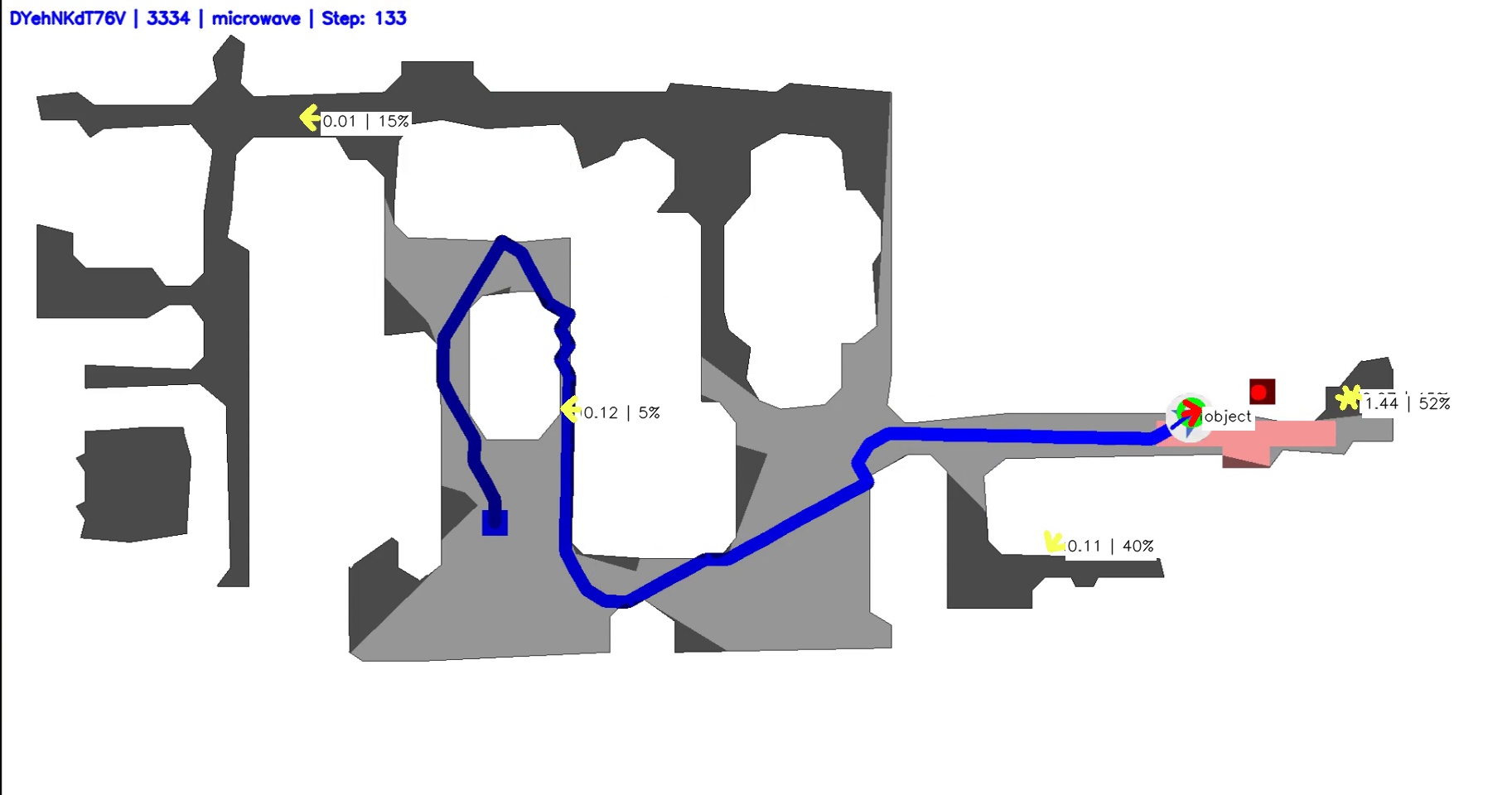}
        \caption{Scene: DYehNKdT76V, Targe: Microwave}
    \end{subfigure}

    \vspace{0.6em}

    \begin{subfigure}[t]{0.32\textwidth}
        \centering
        \includegraphics[width=\linewidth]{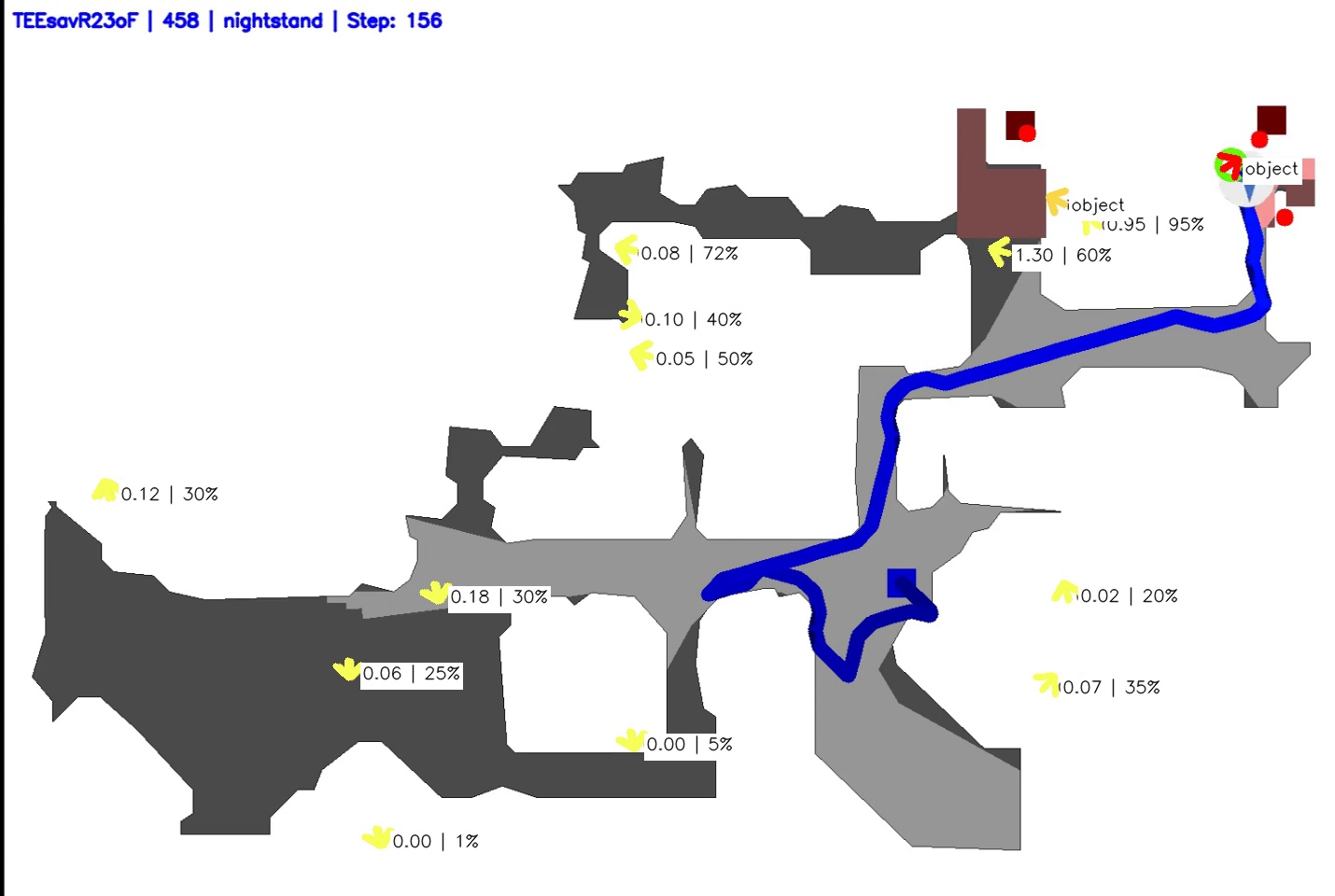}
        \caption{Scene: TEEsavR23oF, Target: Nightstand}
    \end{subfigure}
    \hfill
    \begin{subfigure}[t]{0.32\textwidth}
        \centering
        \includegraphics[width=\linewidth]{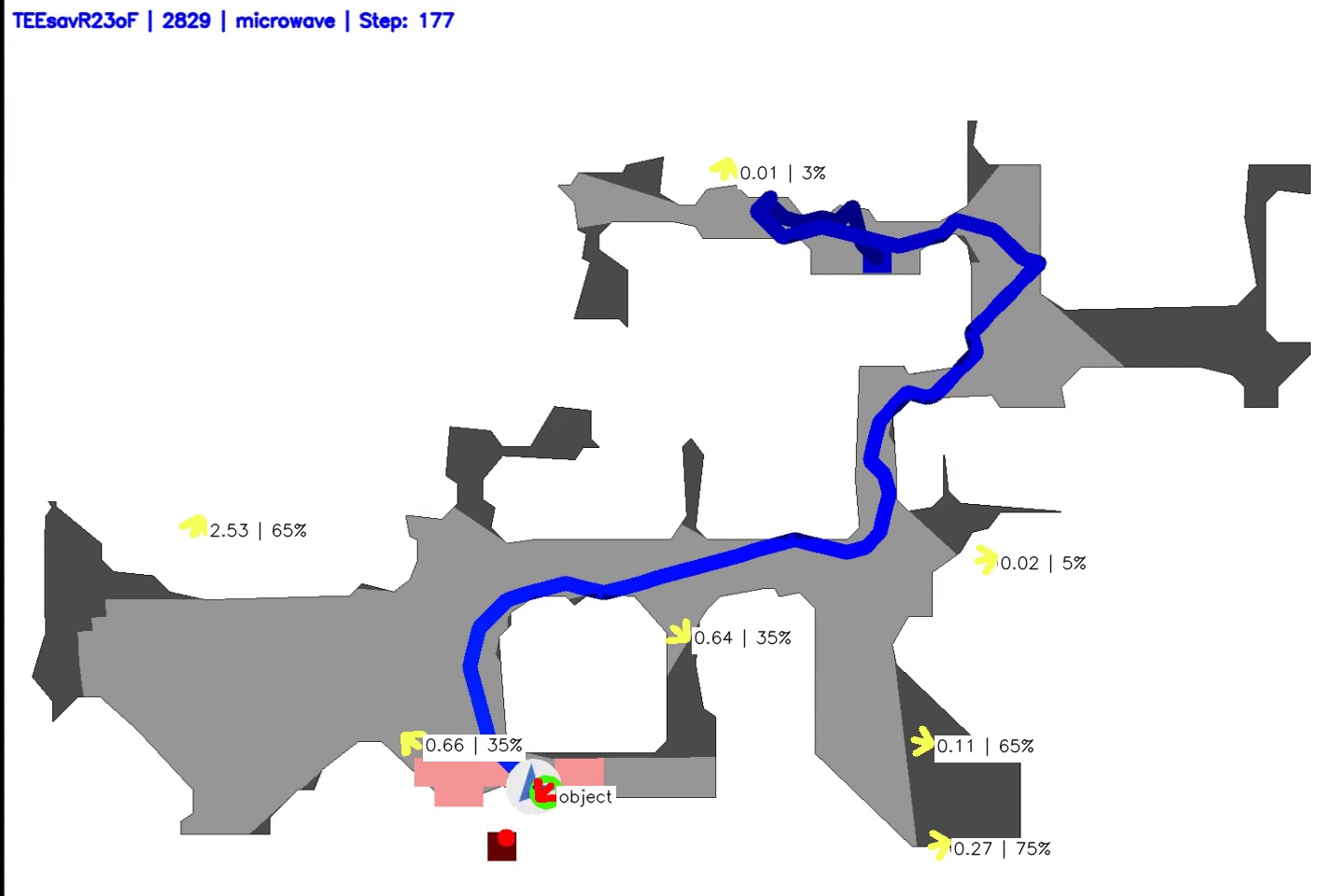}
        \caption{Scene: TEEsavR23oF, Target: Microwave}
    \end{subfigure}
    \hfill
    \begin{subfigure}[t]{0.32\textwidth}
        \centering
        \includegraphics[width=\linewidth]{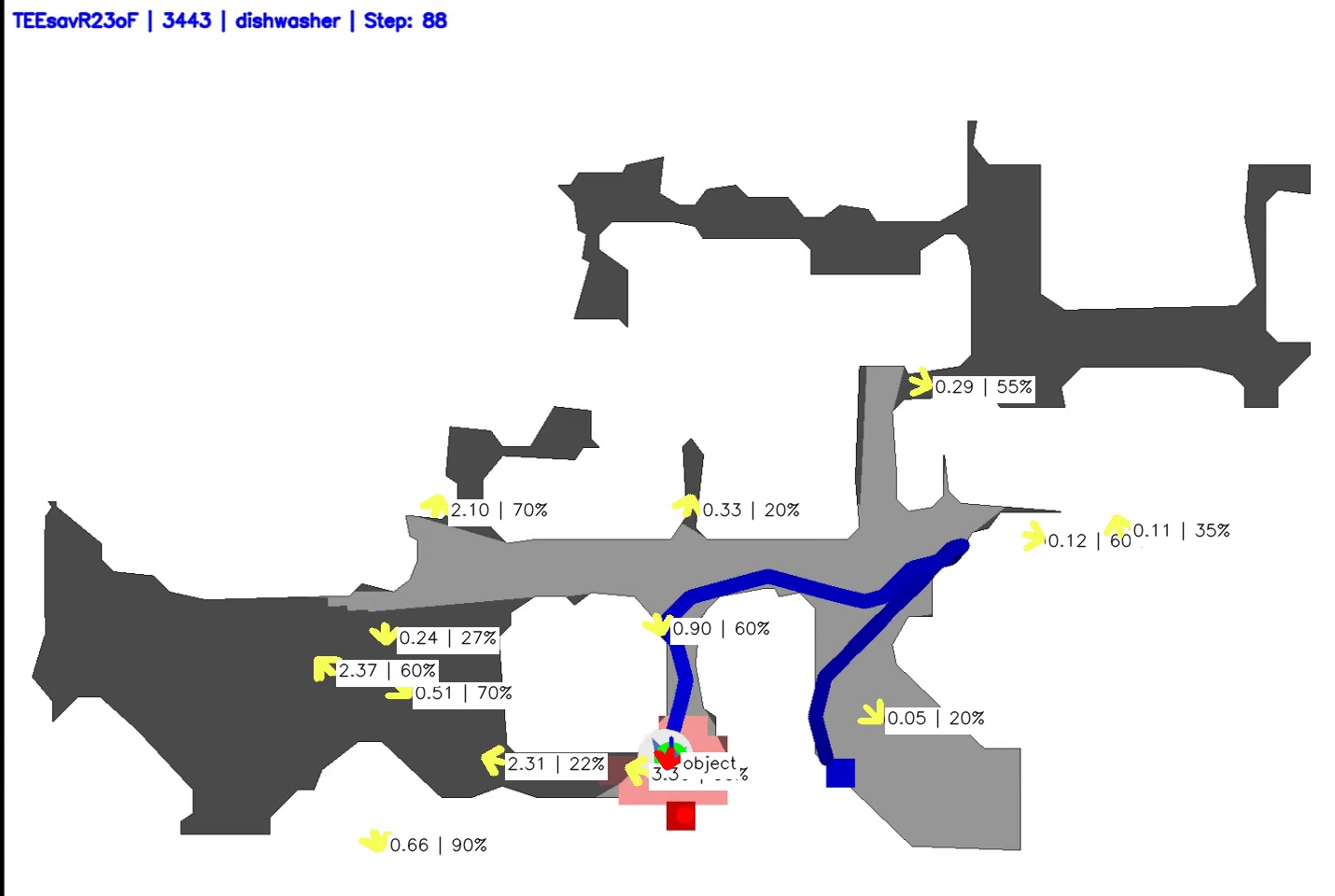}
        \caption{Scene: TEEsavR23oF, Target: Dishwasher}
    \end{subfigure}

\caption{\textbf{Additional Navigation Examples} in OVON across four scenes and three target objects.}
    \label{fig:qualitative_results_ovon}
\end{figure*}

We additionally include examples of navigation under more contextualized goal specifications in Fig. \ref{fig:context_nav}. In these cases, the robot is instructed to navigate to the same object within the same scene, but with different contextual descriptions of the goal, and successfully completes the corresponding tasks. 

These additional results complement the quantitative evaluations in the main paper and further demonstrate the robustness of OpenFrontier across both closed-set object-goal navigation and more open-set scenarios.
Videos that show the full navigation process are provided in the supplementary material.

\subsection{Additional results of ablations with Different Vision--Language Models}

As discussed in Section~IV-B of the main paper, we conduct additional experiments to analyze the sensitivity of OpenFrontier to the choice of vision-language model used for frontier relevance probability estimation. Beyond the qualitative comparisons shown in Fig.~7 of the main paper, we evaluate the full pipeline with different VLMs on the entire HM3D evaluation set.
The quantitative results are summarized in Table~\ref{tab:hm3d_vlm}.

\begin{figure*}[t]
    \centering

    \begin{subfigure}[t]{0.32\textwidth}
        \centering
        \includegraphics[width=\linewidth]{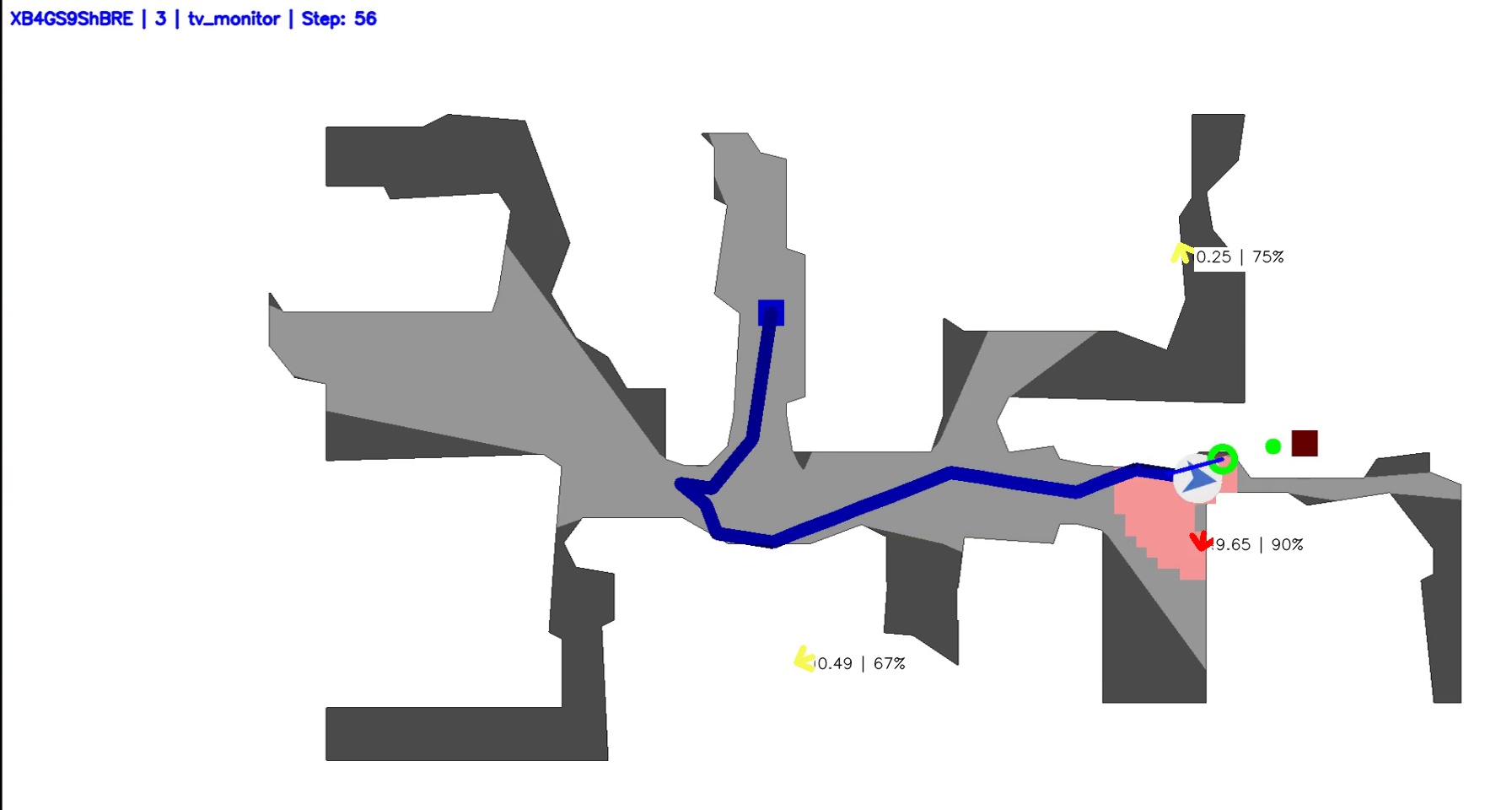}
        \caption{Scene: XB4GS9ShBRE, VLM: Gemini Target: TV Monitor}
    \end{subfigure}
    \hfill
    \begin{subfigure}[t]{0.32\textwidth}
        \centering
        \includegraphics[width=\linewidth]{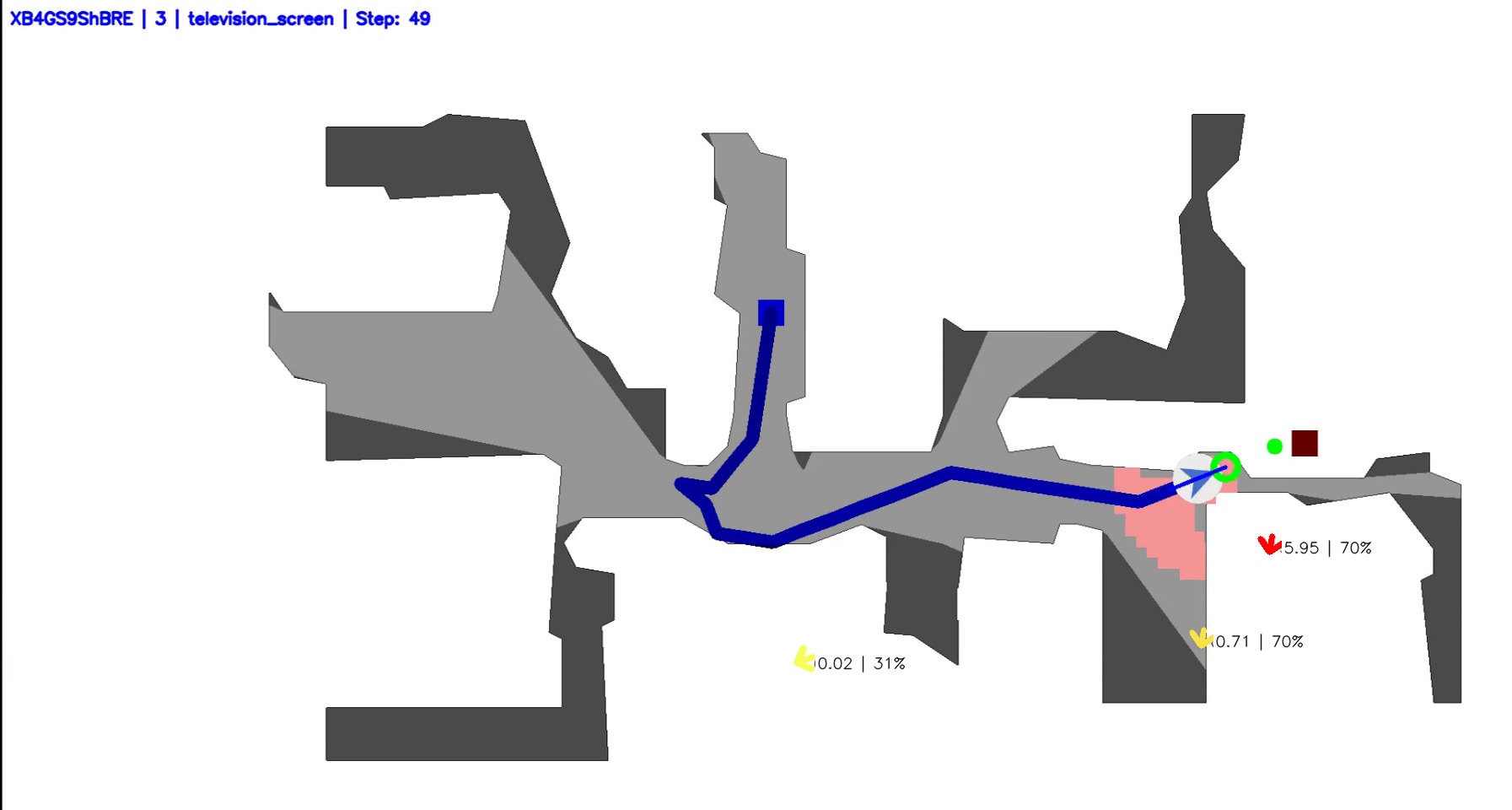}
        \caption{Scene: XB4GS9ShBRE, VLM: Gemma3 Target: TV Monitor}
    \end{subfigure}
    \hfill
    \begin{subfigure}[t]{0.32\textwidth}
        \centering
        \includegraphics[width=\linewidth]{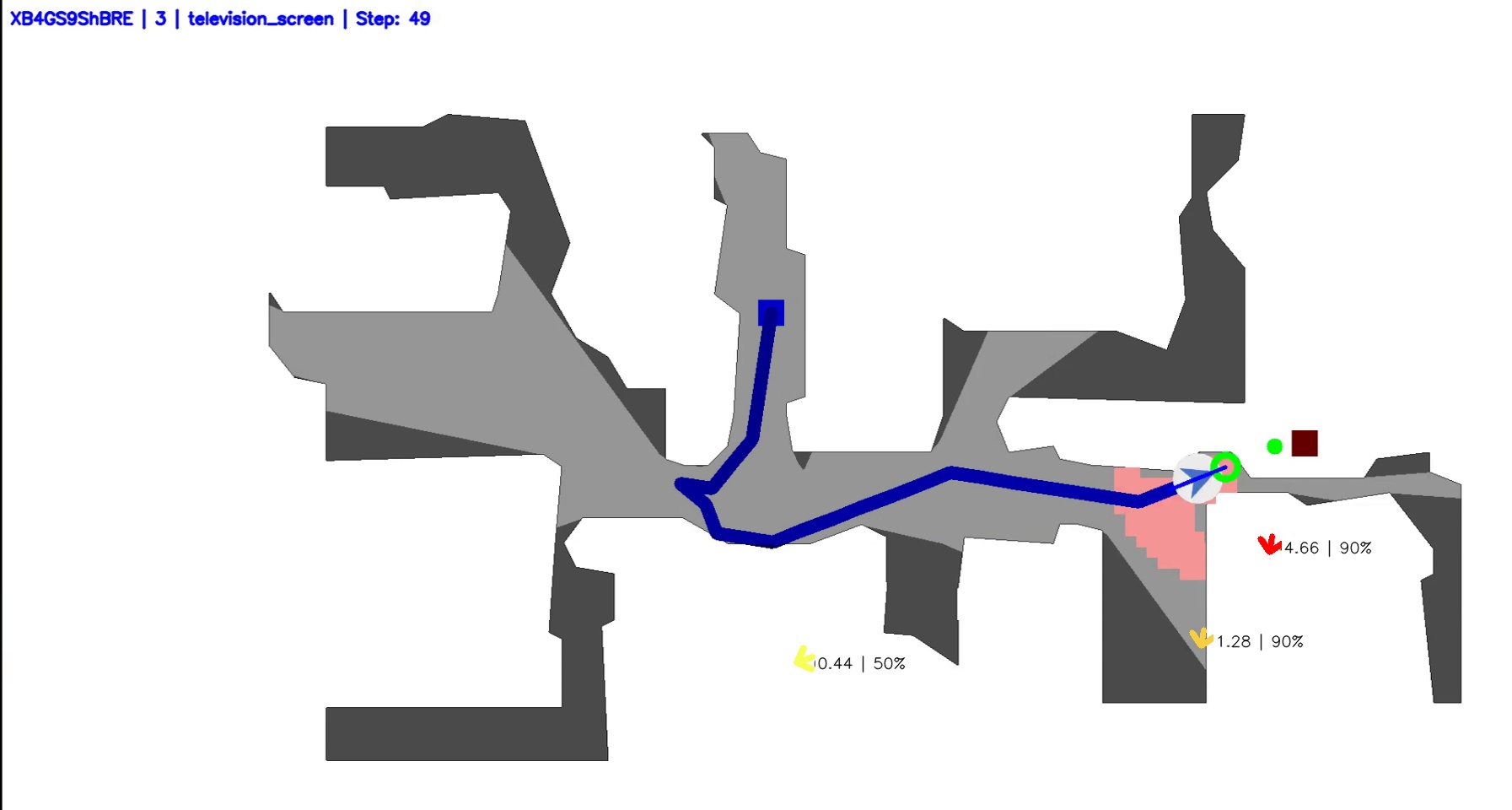}
        \caption{Scene: XB4GS9ShBRE, VLM: InternVL3 Target: TV Monitor}
    \end{subfigure}

    \vspace{0.6em}

    \begin{subfigure}[t]{0.32\textwidth}
        \centering
        \includegraphics[width=\linewidth]{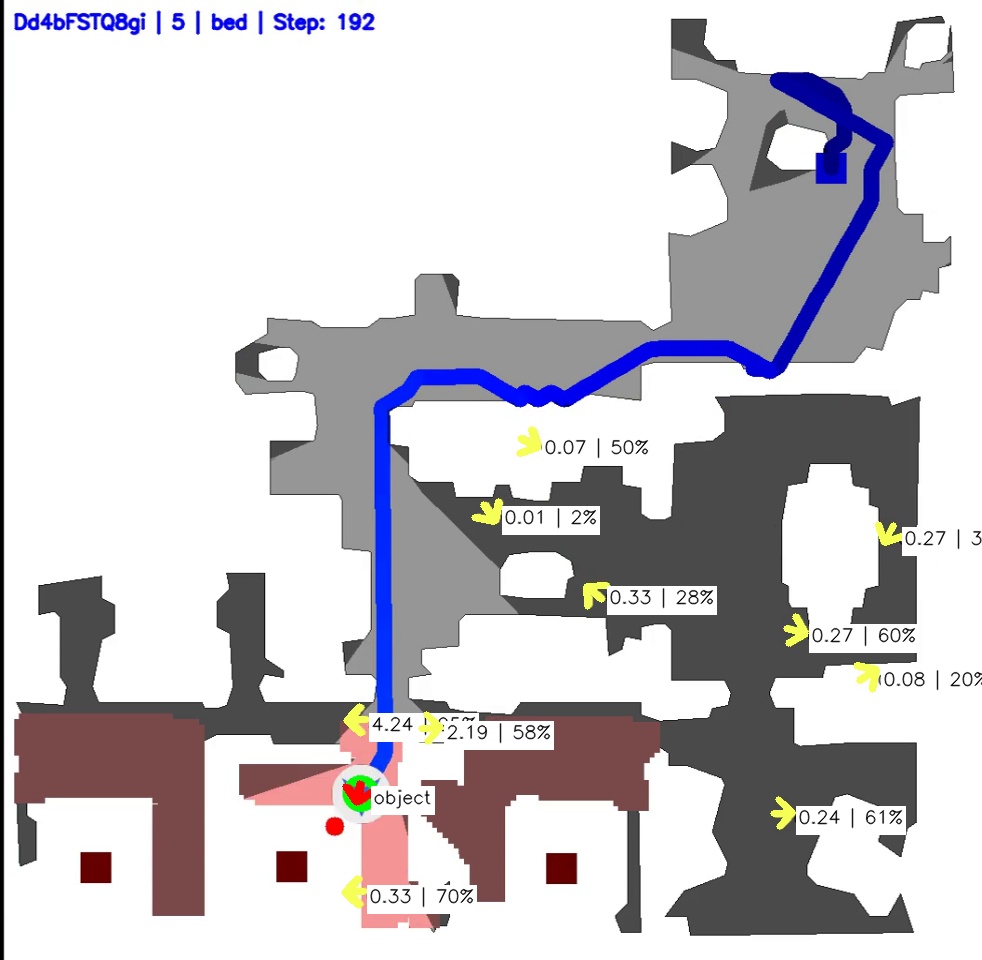}
        \caption{Scene: Dd4bFSTQ8gi, VLM: Gemini Target: Bed}
    \end{subfigure}
    \hfill
    \begin{subfigure}[t]{0.32\textwidth}
        \centering
        \includegraphics[width=\linewidth]{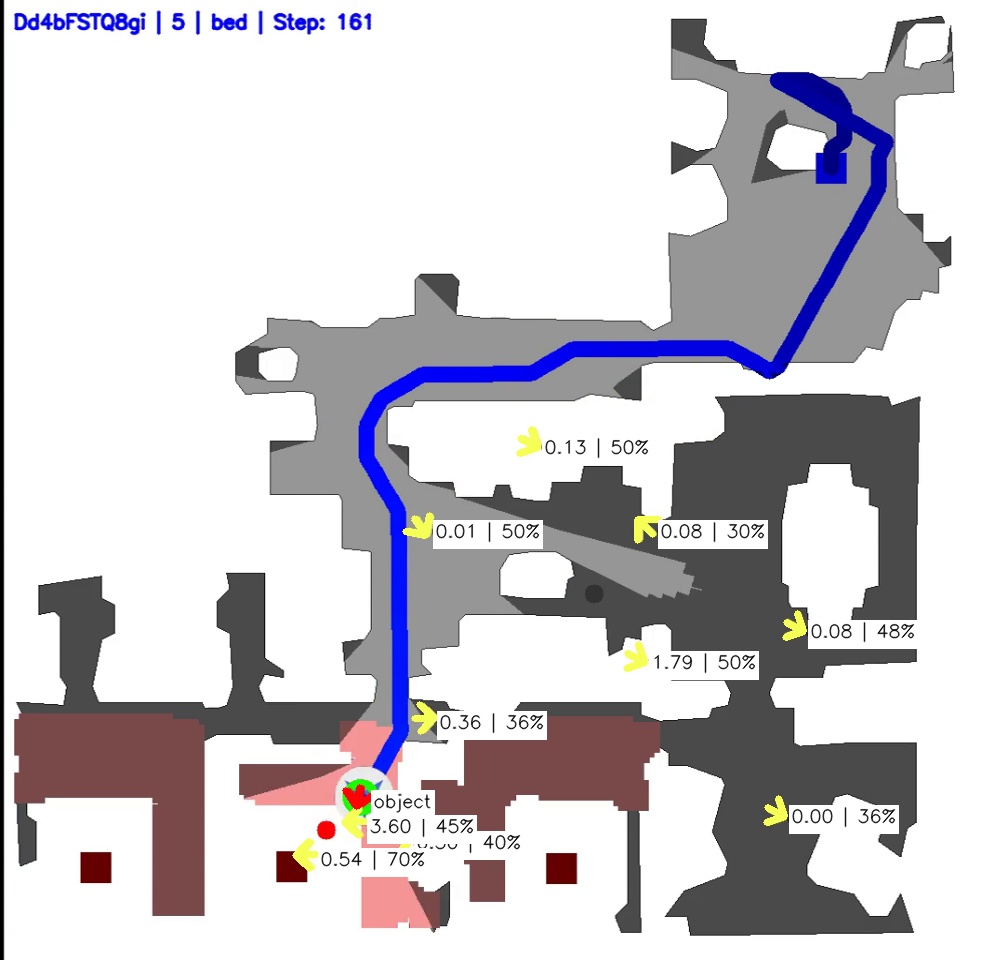}
        \caption{Scene: Dd4bFSTQ8gi, VLM: Gemma3 Target: Bed}
    \end{subfigure}
    \hfill
    \begin{subfigure}[t]{0.32\textwidth}
        \centering
        \includegraphics[width=\linewidth]{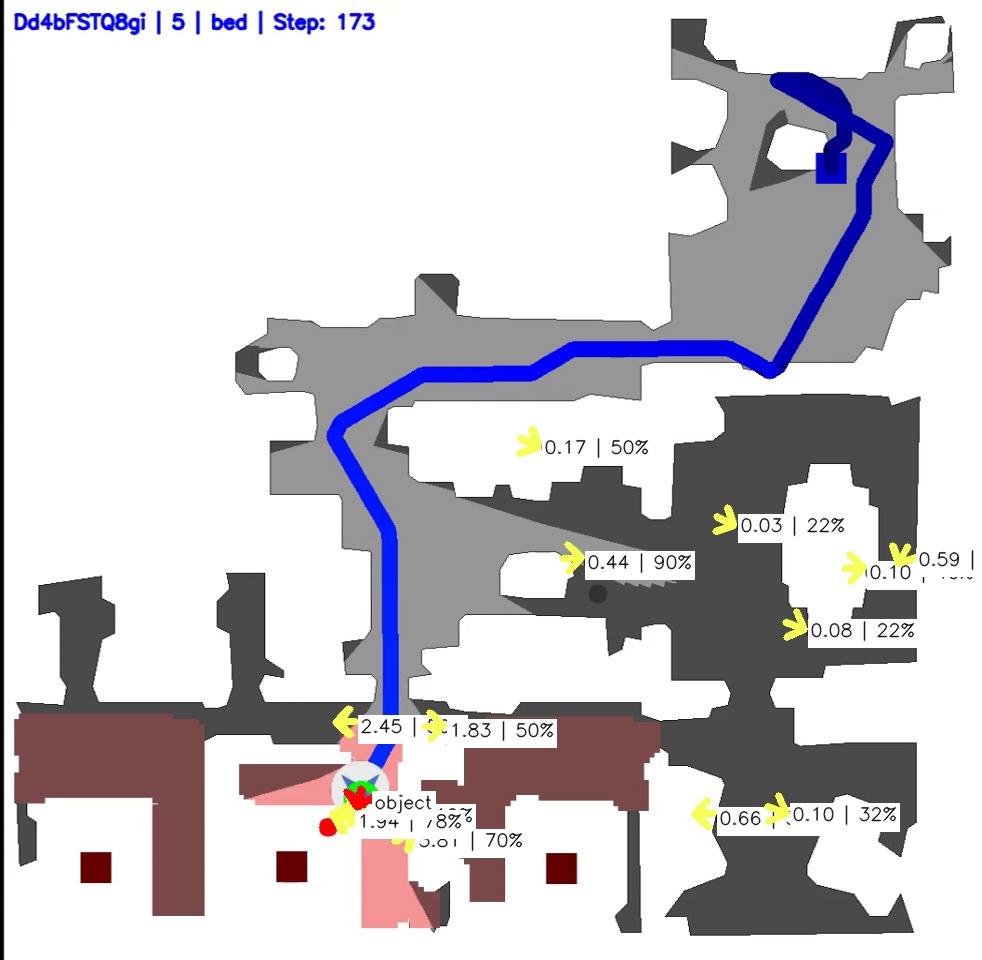}
        \caption{Scene: Dd4bFSTQ8gi, VLM: InternVL3 Target: Bed}
    \end{subfigure}

    \vspace{0.6em}

    \begin{subfigure}[t]{0.32\textwidth}
        \centering
        \includegraphics[width=\linewidth]{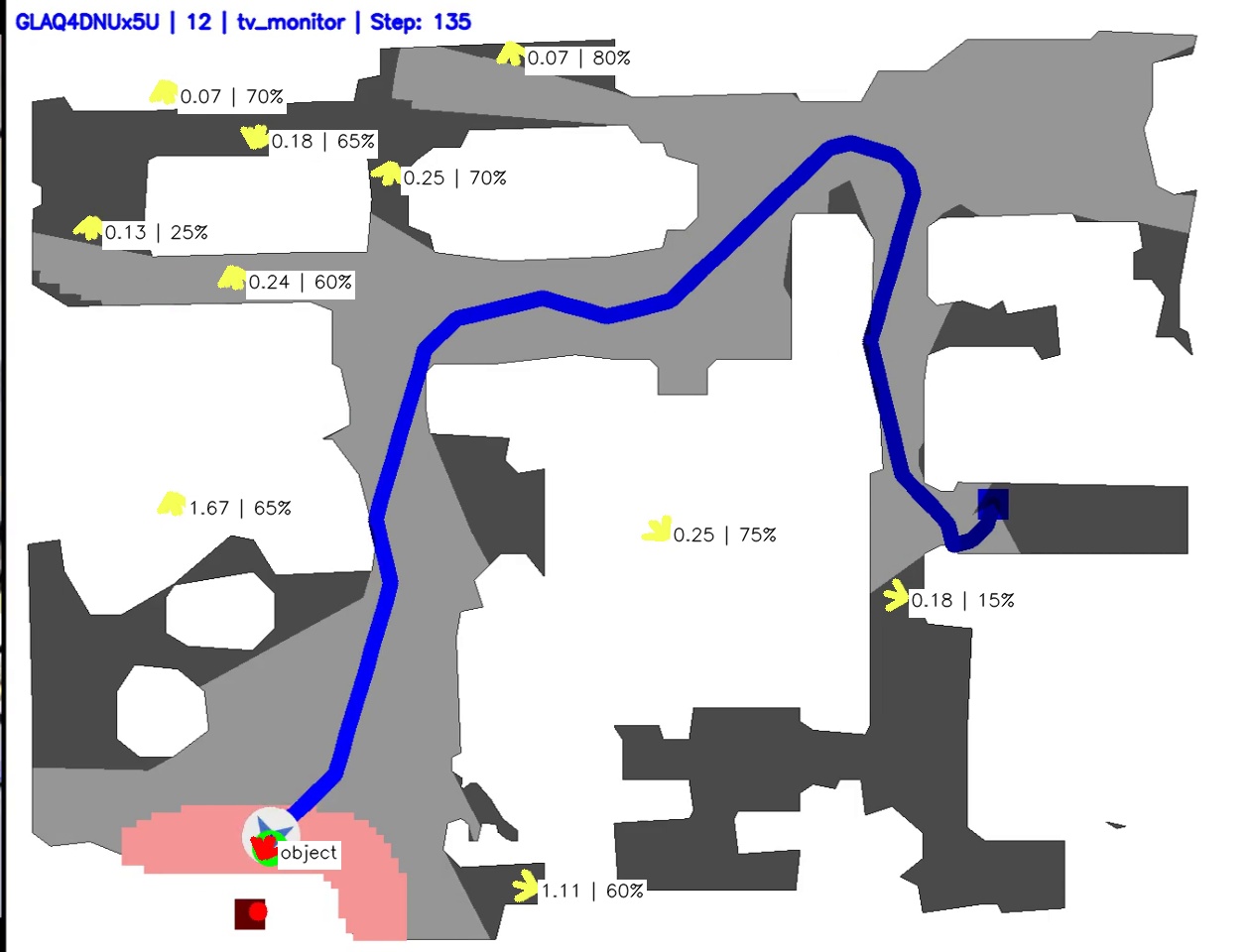}
        \caption{Scene: GLAQ4DNUx5U, VLM: Gemini, Targe: TV Monitor}
    \end{subfigure}
    \hfill
    \begin{subfigure}[t]{0.32\textwidth}
        \centering
        \includegraphics[width=\linewidth]{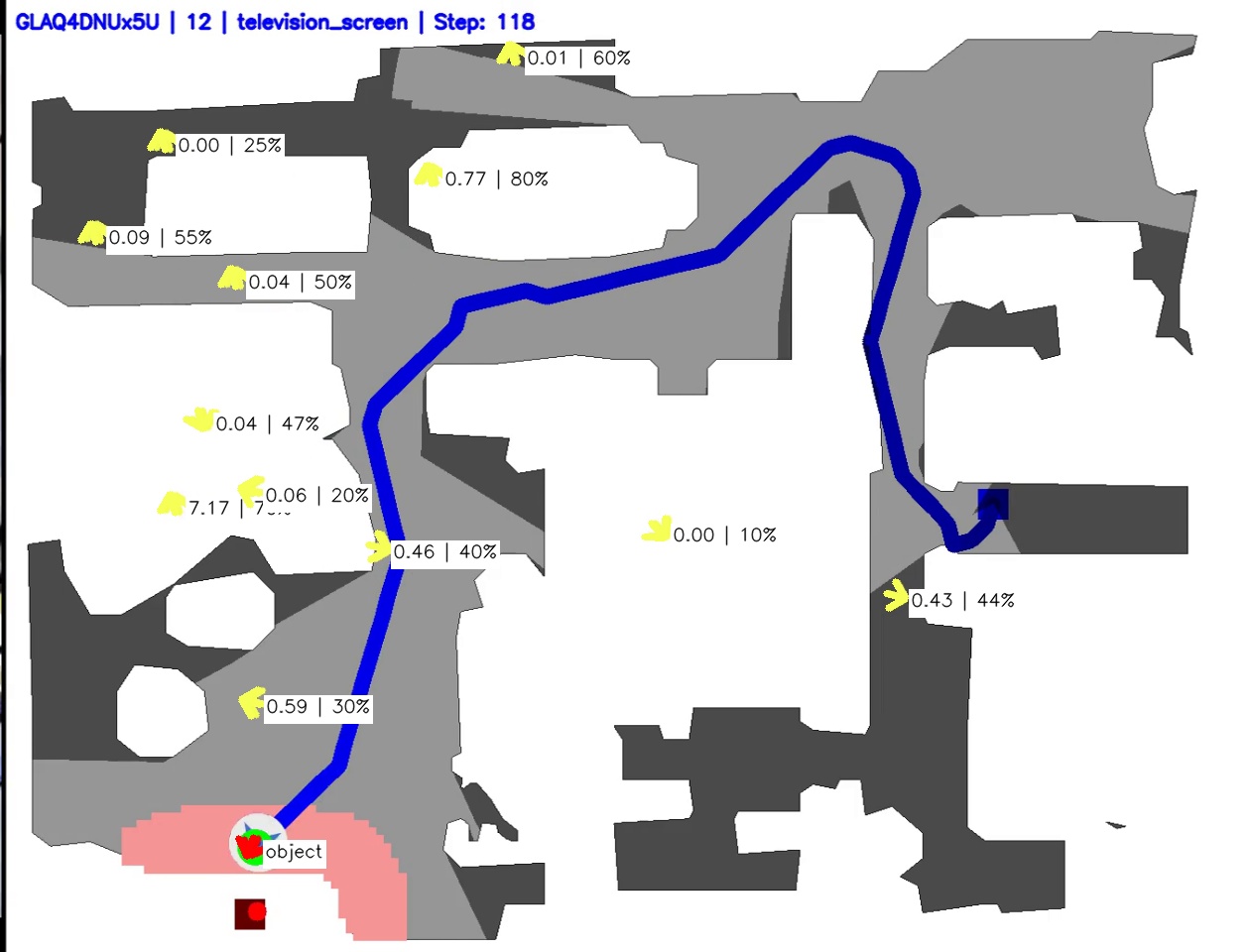}
        \caption{Scene: GLAQ4DNUx5U, VLM: Gemma3, Targe: TV Monitor}
    \end{subfigure}
    \hfill
    \begin{subfigure}[t]{0.32\textwidth}
        \centering
        \includegraphics[width=\linewidth]{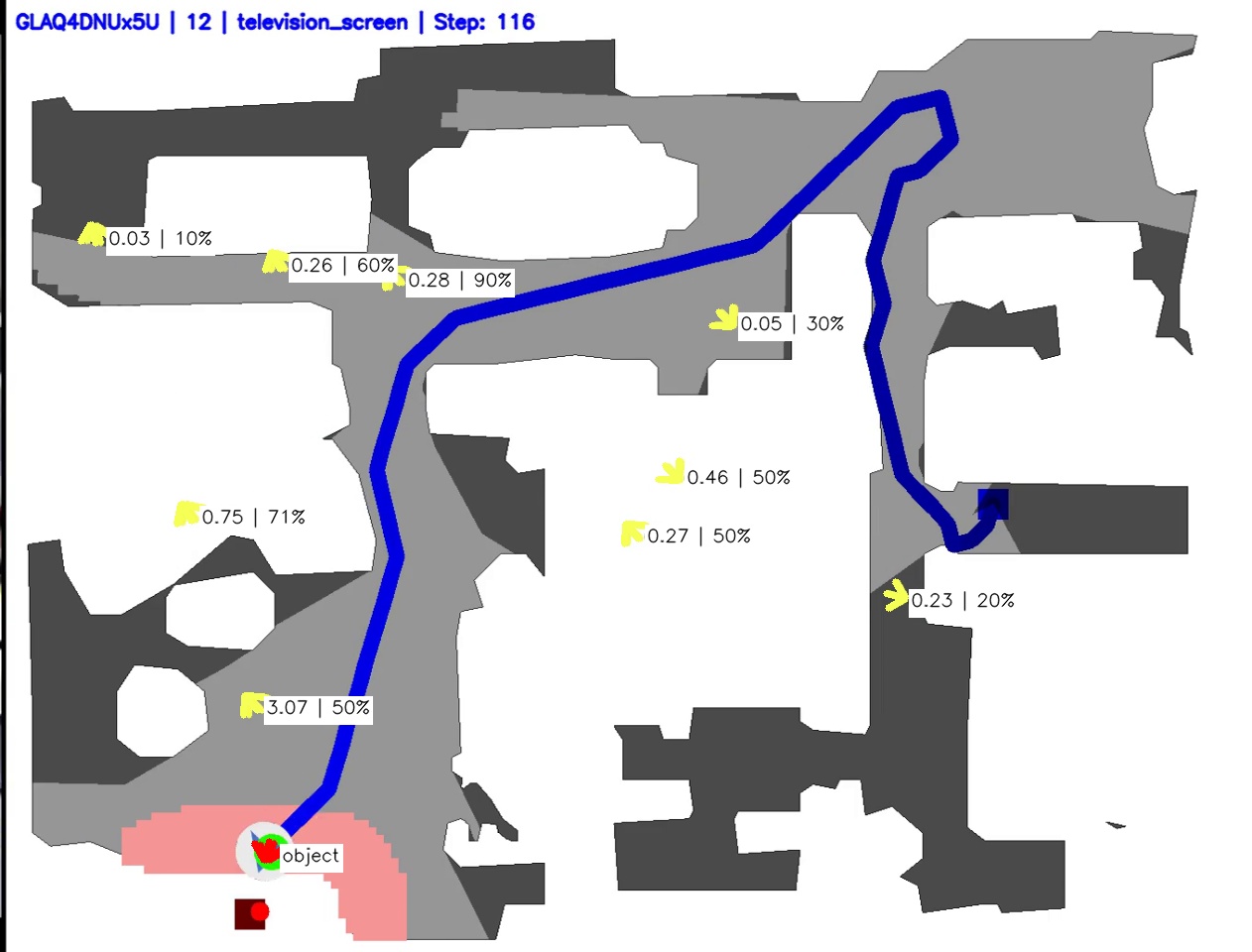}
        \caption{Scene: GLAQ4DNUx5U, VLM: InternVL3, Targe: TV Monitor}
    \end{subfigure}

    \vspace{0.6em}

    \begin{subfigure}[t]{0.32\textwidth}
        \centering
        \includegraphics[width=\linewidth]{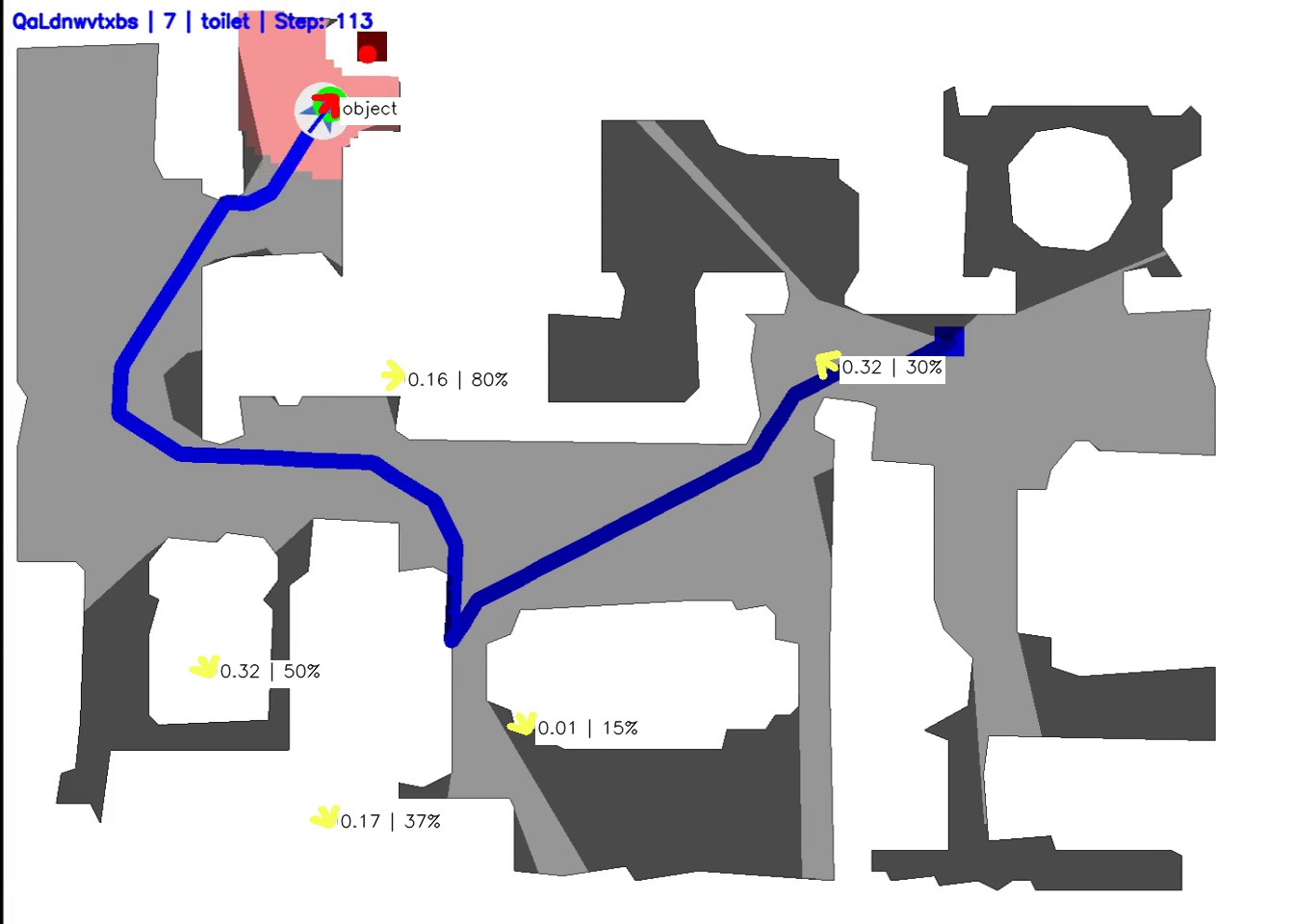}
        \caption{Scene: QaLdnwvtxbs, VLM: Gemini, Targe: Toilet}
    \end{subfigure}
    \hfill
    \begin{subfigure}[t]{0.32\textwidth}
        \centering
        \includegraphics[width=\linewidth]{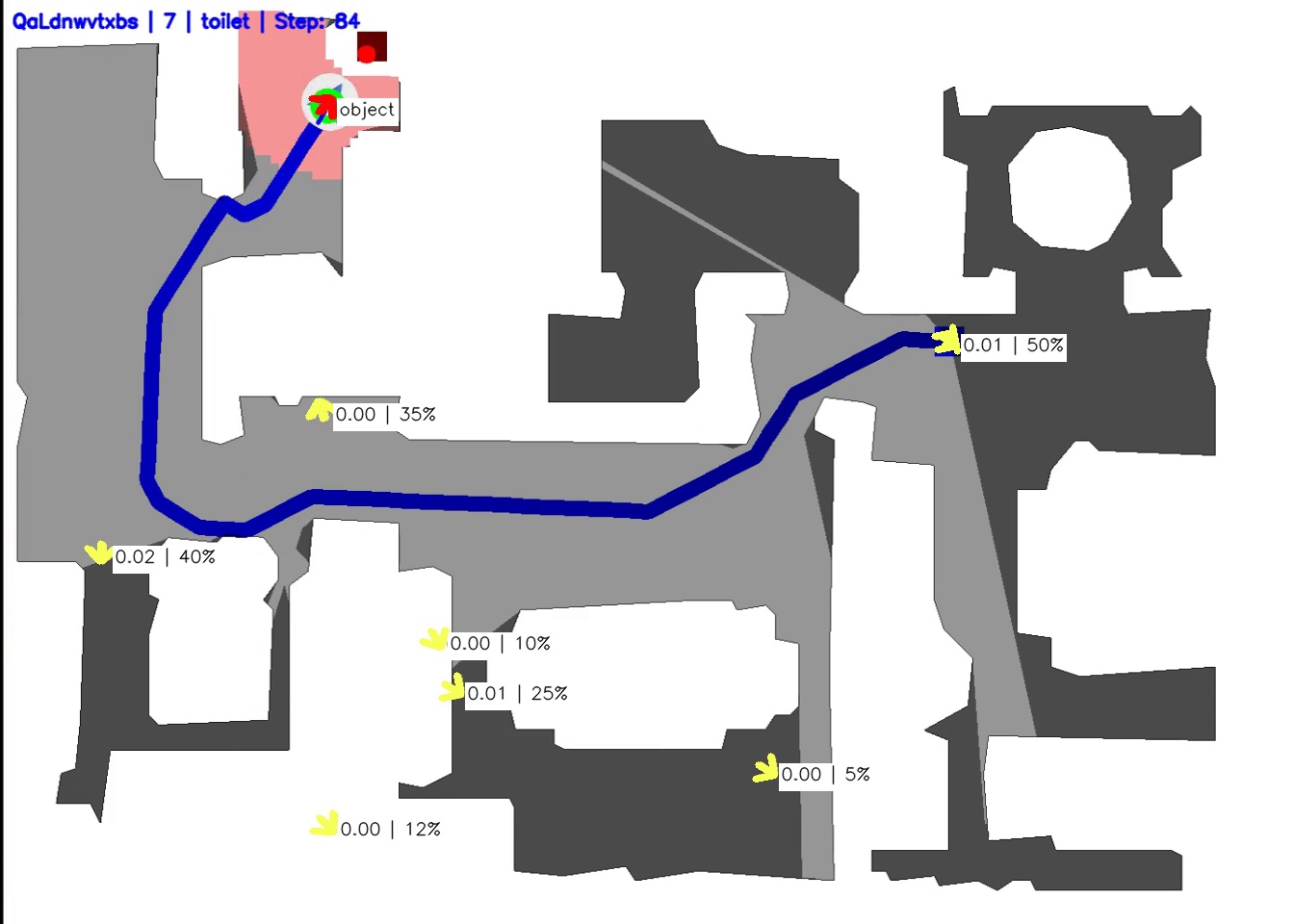}
        \caption{Scene: QaLdnwvtxbs, VLM: Gemma3, Targe: Toilet}
    \end{subfigure}
    \hfill
    \begin{subfigure}[t]{0.32\textwidth}
        \centering
        \includegraphics[width=\linewidth]{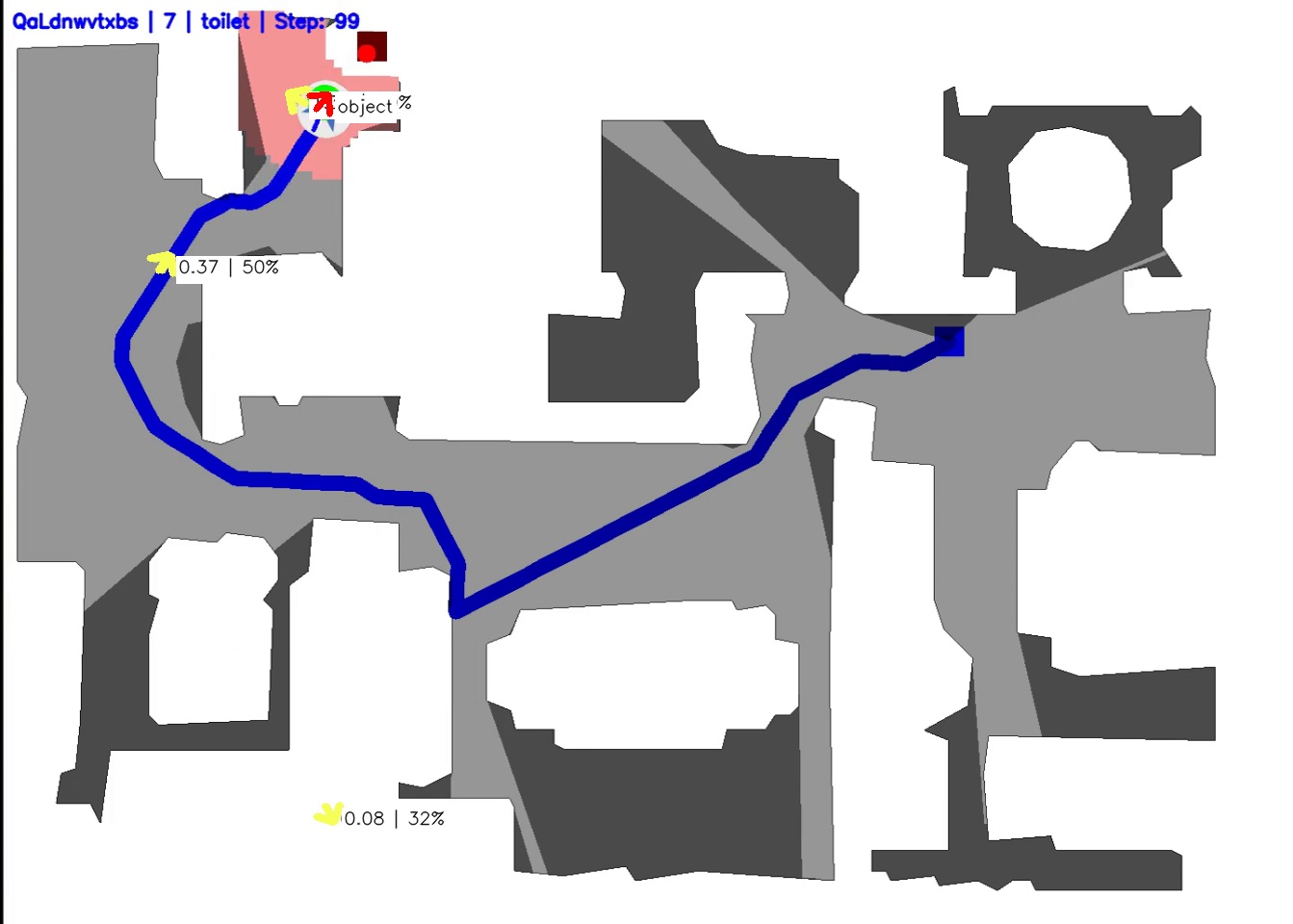}
        \caption{Scene: QaLdnwvtxbs, VLM: InternVL3, Targe: Toilet}
    \end{subfigure}

\caption{\textbf{Navigation Examples with different VLMs} in HM3D across four scenes with same target objects.}
    \label{fig:qualitative_vlms}
\end{figure*}

The best performance is achieved using Gemini-2.5-flash, same as the configuration adopted in the main experiments.
Replacing it with Gemma-3 or InternVL3 results in only marginal performance degradation, with less than 3\% drop in success rate and approximately 2\% drop in success weighted by path length.
Notably, both Gemma-3 and InternVL3 are publicly available models with relatively small model sizes: we use Gemma-3 with 4B parameters, and InternVL3 that with 8B parameters.
Fig.~\ref{fig:qualitative_vlms} presents additional qualitative comparisons of navigation trajectories produced by different VLMs.

\subsection{Real-World Navigation Results}
We evaluate OpenFrontier in a large-scale indoor environment with diverse layouts and room types.
The robot is initialized at random locations, and navigation targets are specified as objects present in the environment, including fire extinguishers, microwaves, scissor lifts, and television.
Videos illustrating the full navigation process are provided in the supplementary material.

\subsection{Failure Cases}
We present representative failure cases observed during benchmarking and group them into three categories: 
\emph{reaching the maximum step limit}, \emph{false positives}, and \emph{robot stuck/cannot reach goal}.

Both failure due to reaching the maximum step limit and robot stuck result in the same outcome-the episode terminates after exceeding the allowed number of steps without success. However they arise from different underlying causes. In the \emph{reaching the maximum step limit} cases, the robot fails to identify the target object throughout the whole episode. In contrast, in \emph{robot stuck} cases, the robot successfully detects the target object but becomes trapped in a local minimum while planning or executing a path toward it, eventually exhausting the step budget. The \emph{false positive} cases occur when the robot navigates to an object that is semantically consistent with the language description but is not annotated as the correct target in the benchmark ground truth.

\begin{figure*}[t]
    \centering

    \begin{subfigure}[t]{0.32\textwidth}
        \centering
        \includegraphics[width=\linewidth]{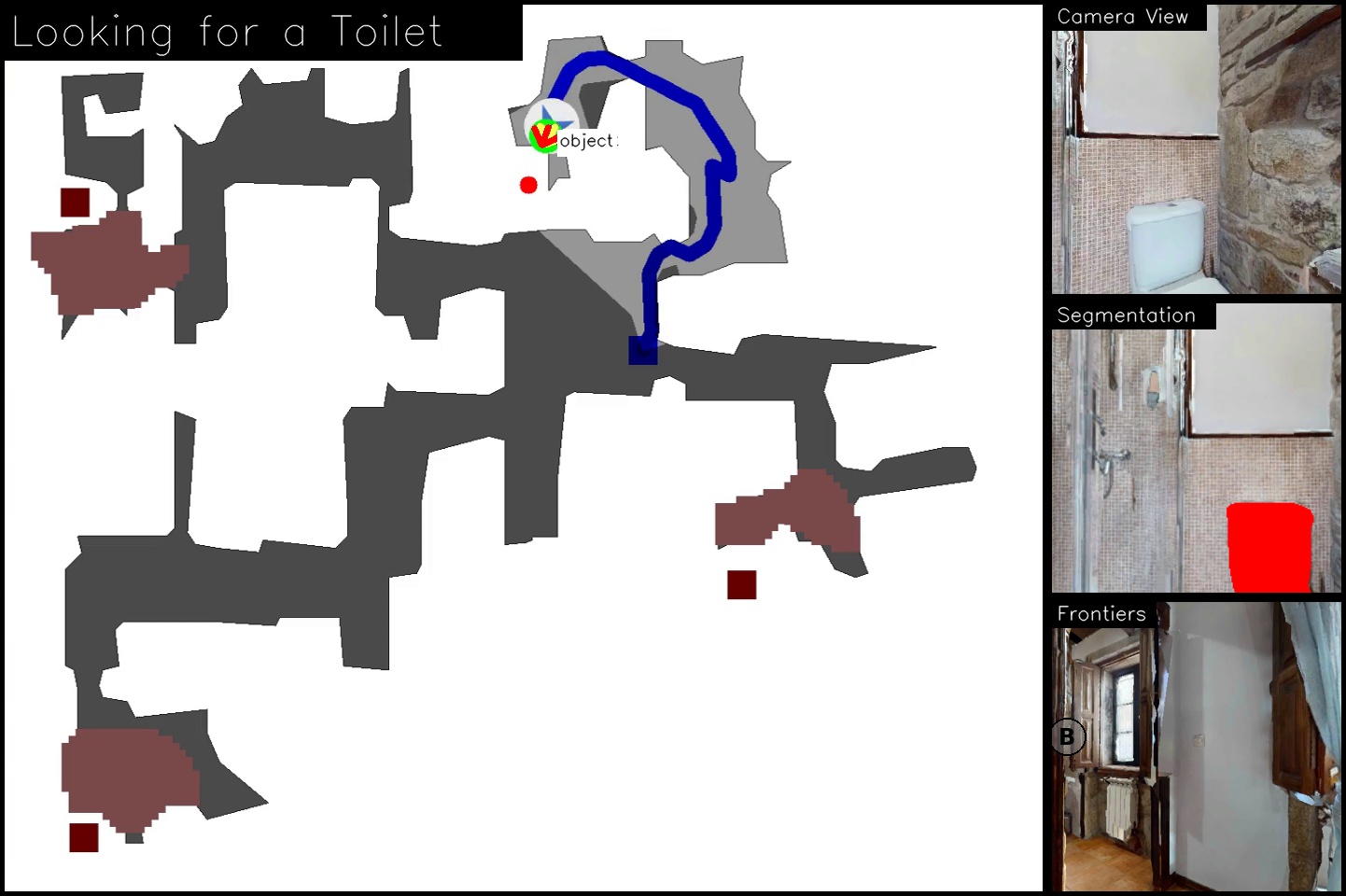}
        \caption{Scene: a8BtkwhxdRV, Target: Toilet, Failure: False Positive}
    \end{subfigure}
    \hfill
    \begin{subfigure}[t]{0.32\textwidth}
        \centering
        \includegraphics[width=\linewidth]{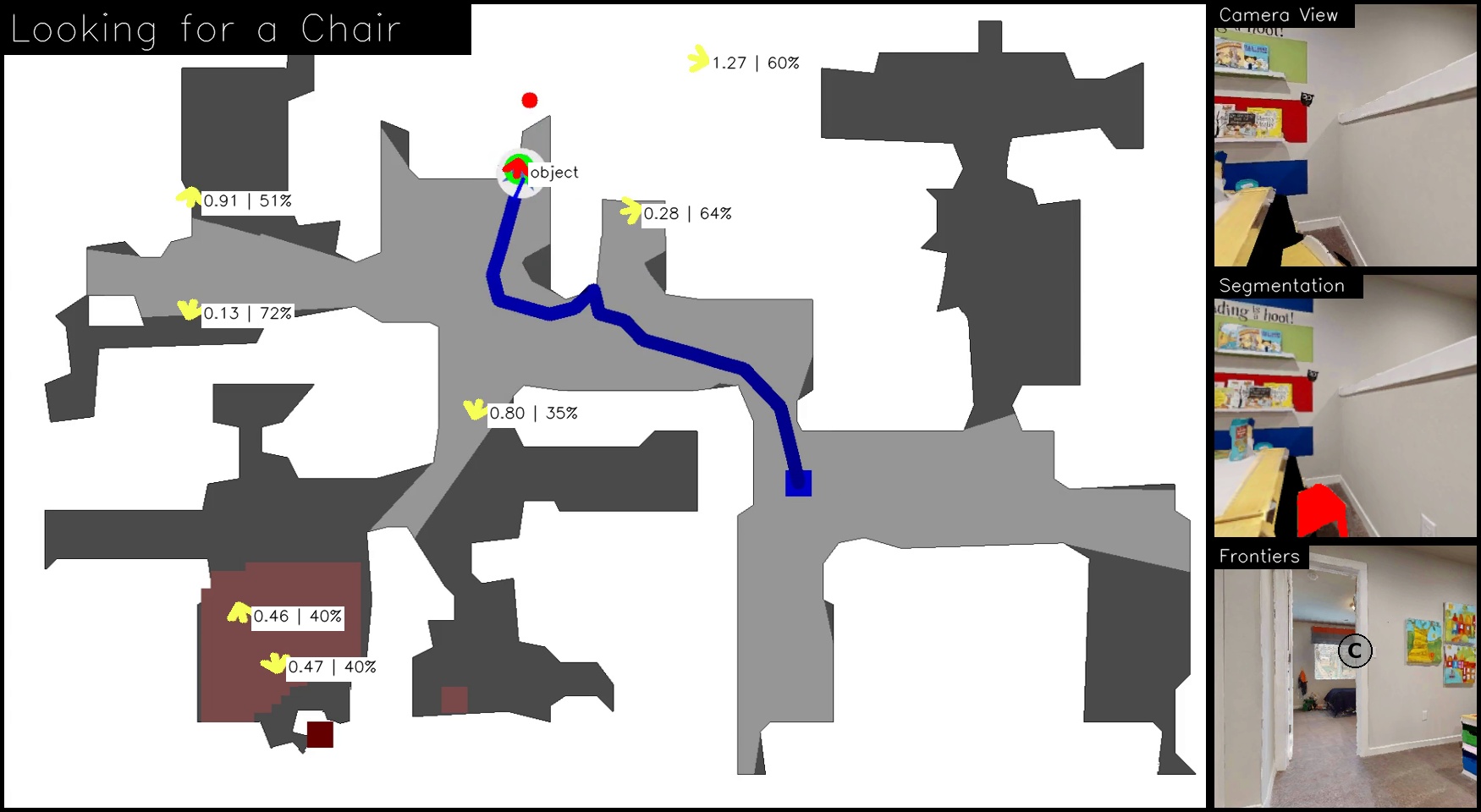}
        \caption{Scene: bxsVRursffK, Target: Chair, Failure: False Positive}
    \end{subfigure}
    \hfill
    \begin{subfigure}[t]{0.32\textwidth}
        \centering
        \includegraphics[width=\linewidth]{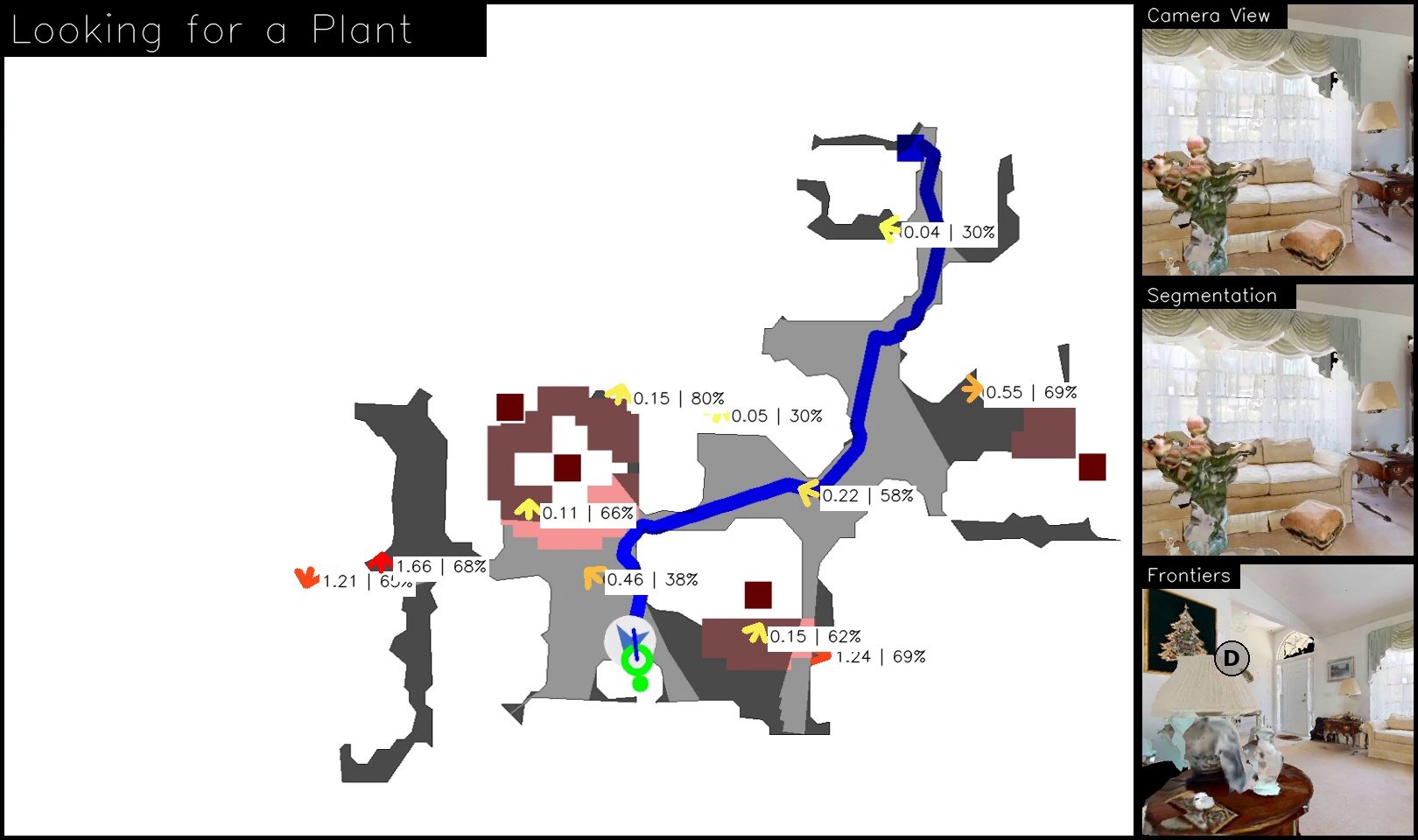}
        \caption{Scene: VBzV5z6i1WS, Target: Plant, Failure: False Positive}
    \end{subfigure}

    \vspace{0.6em}

    \begin{subfigure}[t]{0.32\textwidth}
        \centering
        \includegraphics[width=\linewidth]{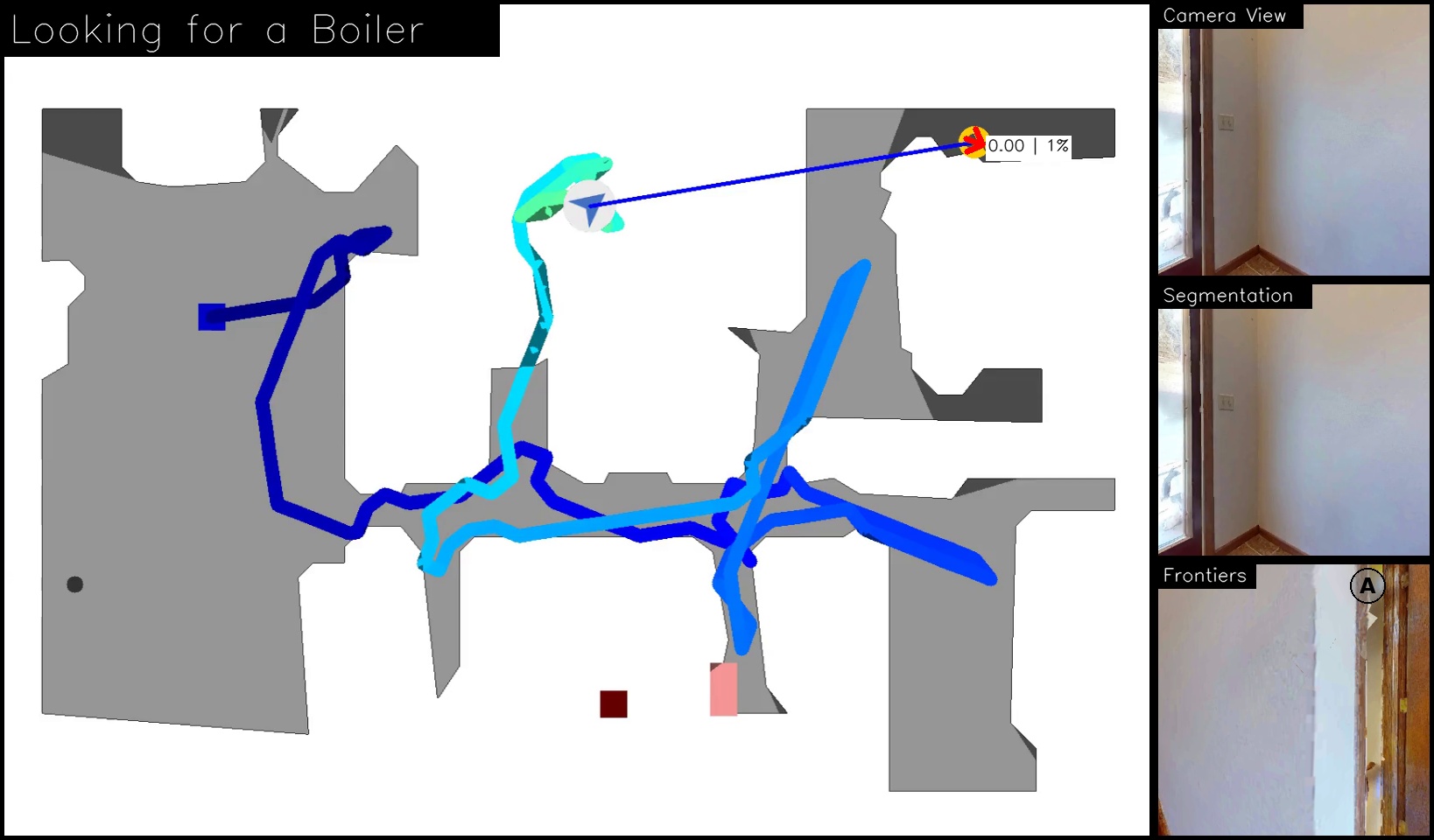}
        \caption{Scene: 4ok3usBNeis, Target: Boiler, Failure: Reach Max Steps}
    \end{subfigure}
    \hfill
    \begin{subfigure}[t]{0.32\textwidth}
        \centering
        \includegraphics[width=\linewidth]{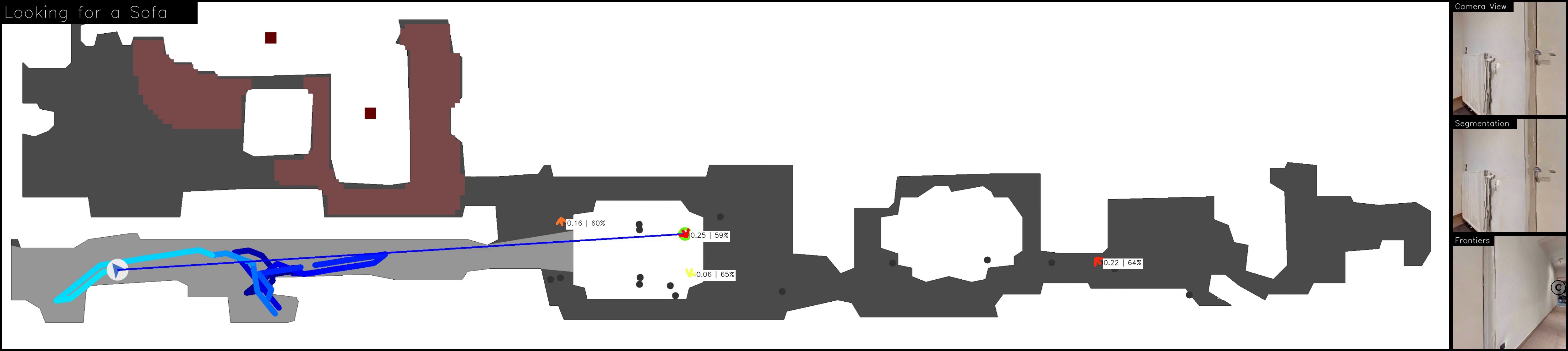}
        \caption{Scene: 6s7QHgap2fW, Target: Sofa, Failure: Reach Max Steps}
    \end{subfigure}
    \hfill
    \begin{subfigure}[t]{0.32\textwidth}
        \centering
        \includegraphics[width=\linewidth]{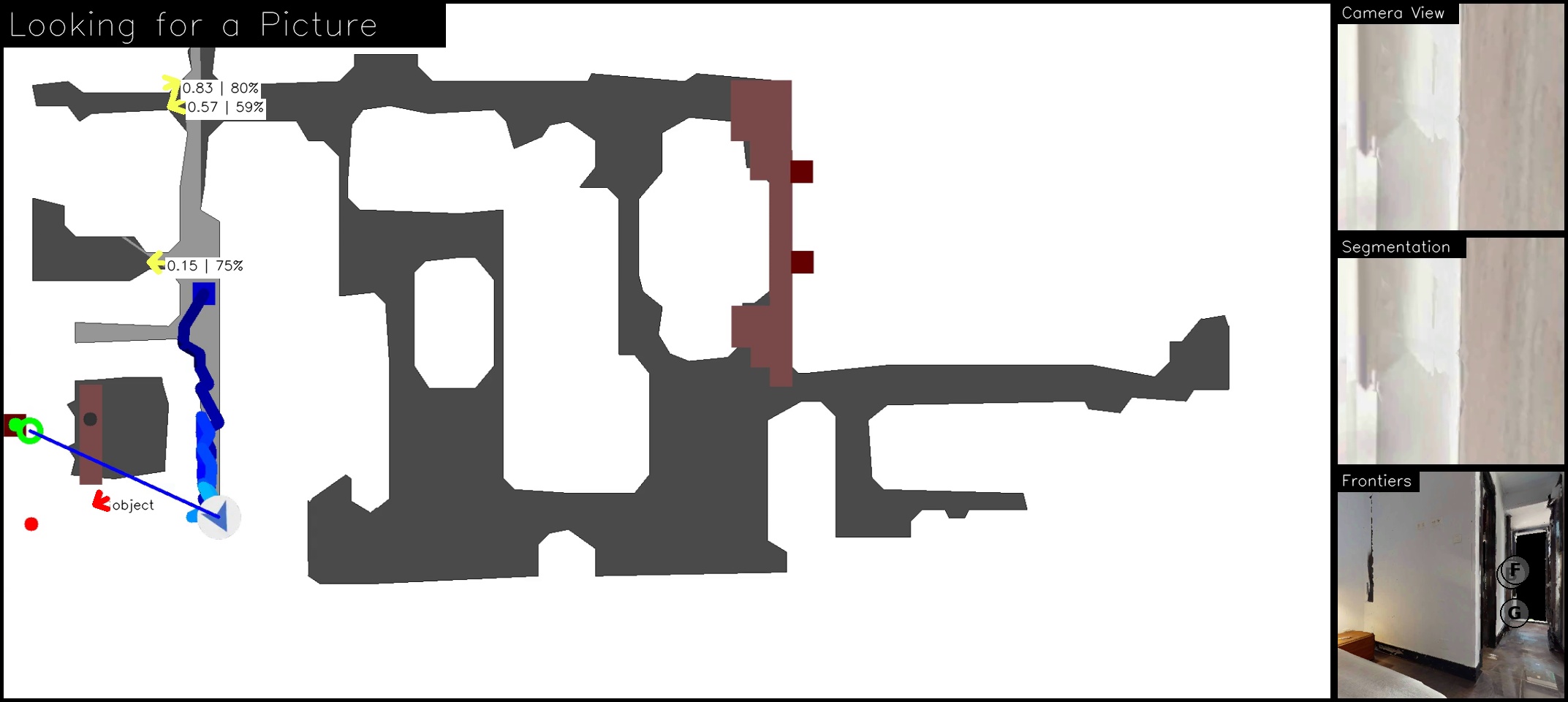}
        \caption{Scene: DYehNKdT76V, Target: Picture, Failure: Cannot Reach Goal}
    \end{subfigure}

    \vspace{0.6em}

    \begin{subfigure}[t]{0.32\textwidth}
        \centering
        \includegraphics[width=\linewidth]{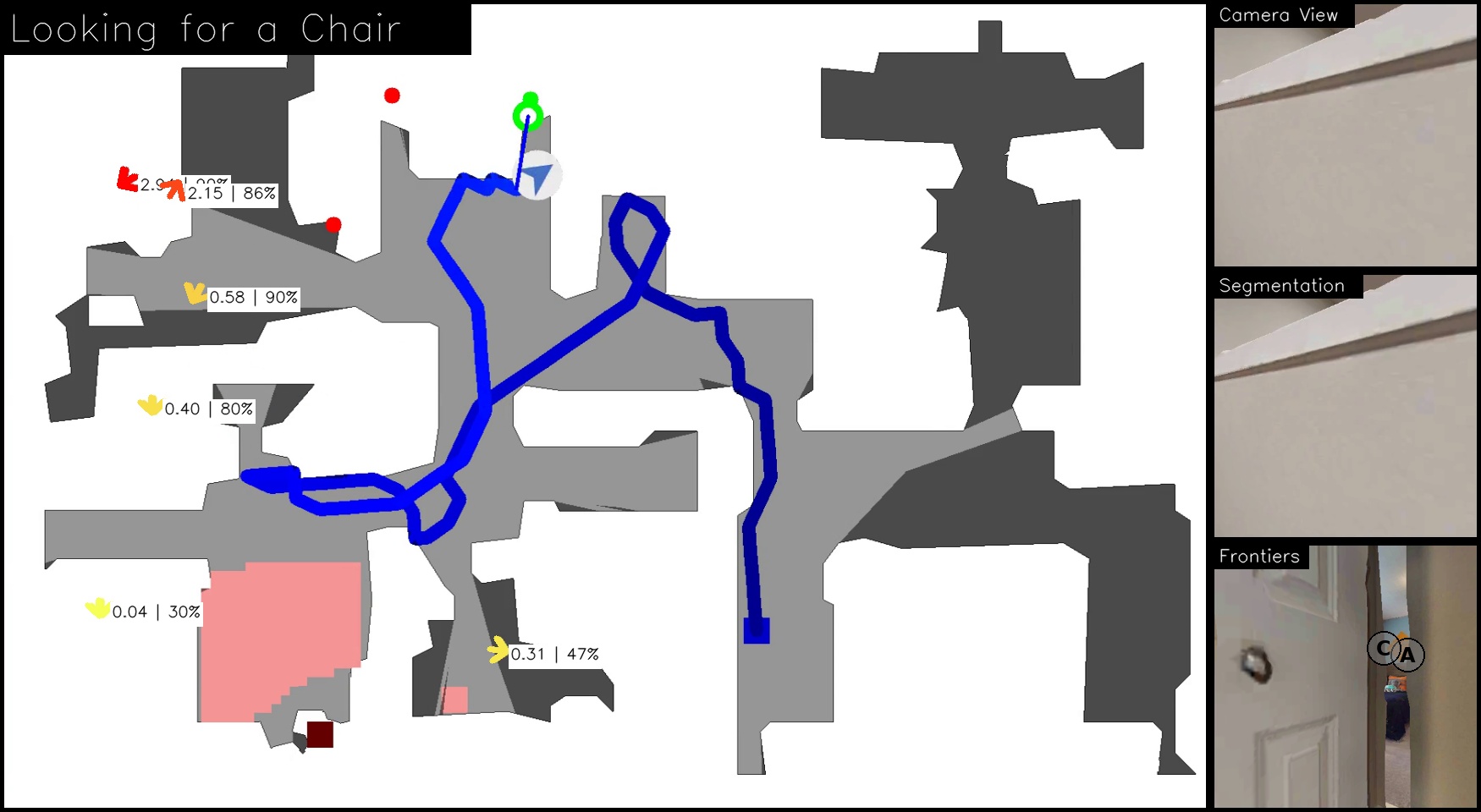}
        \caption{Scene: bxsVRursffK, Target: Chair, Failure: False Positive}
    \end{subfigure}
    \hfill
    \begin{subfigure}[t]{0.32\textwidth}
        \centering
        \includegraphics[width=\linewidth]{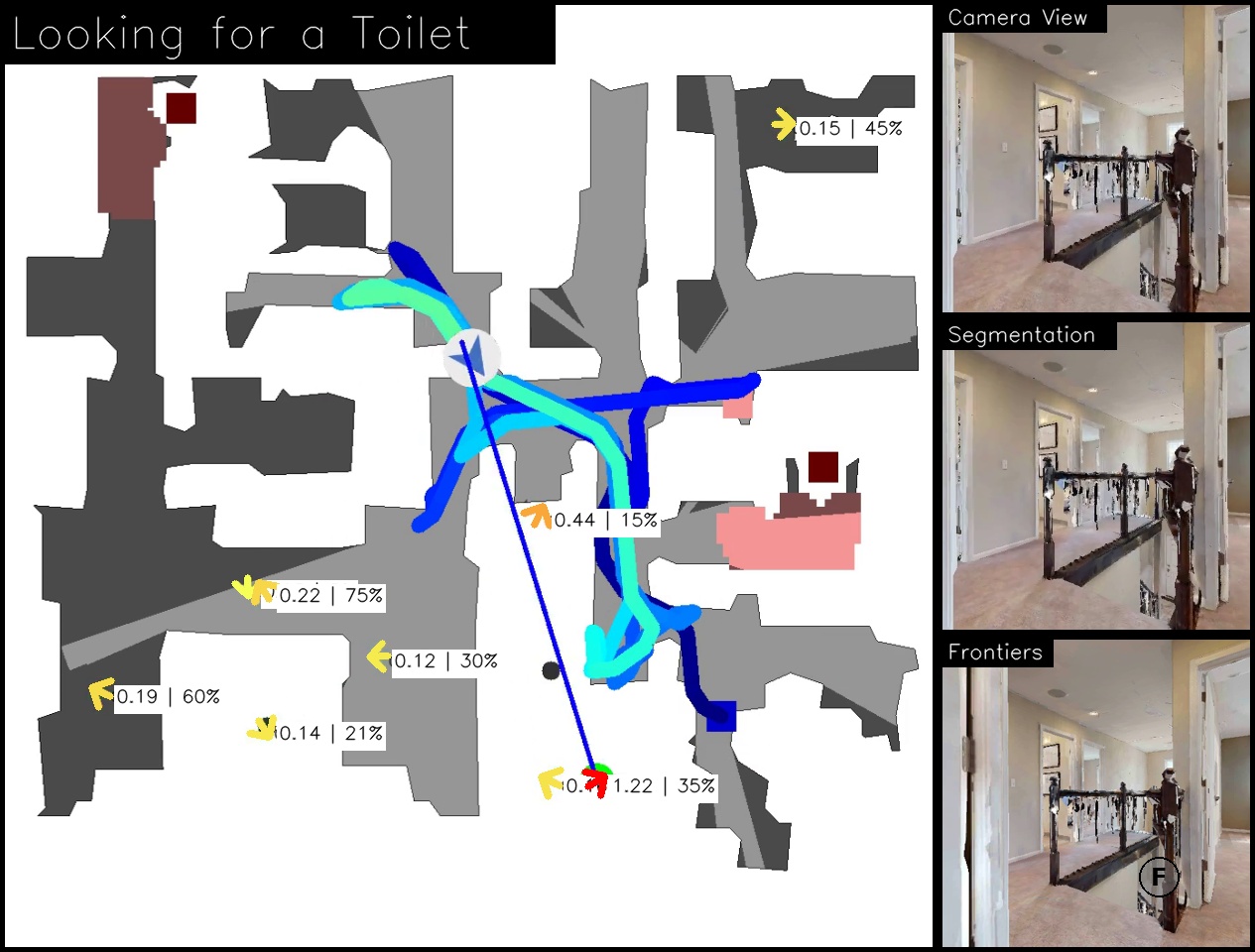}
        \caption{Scene: CrMo8WxCyVb, Target: Toilet, Failure: Reach Max Steps}
    \end{subfigure}
    \hfill
    \begin{subfigure}[t]{0.32\textwidth}
        \centering
        \includegraphics[width=\linewidth]{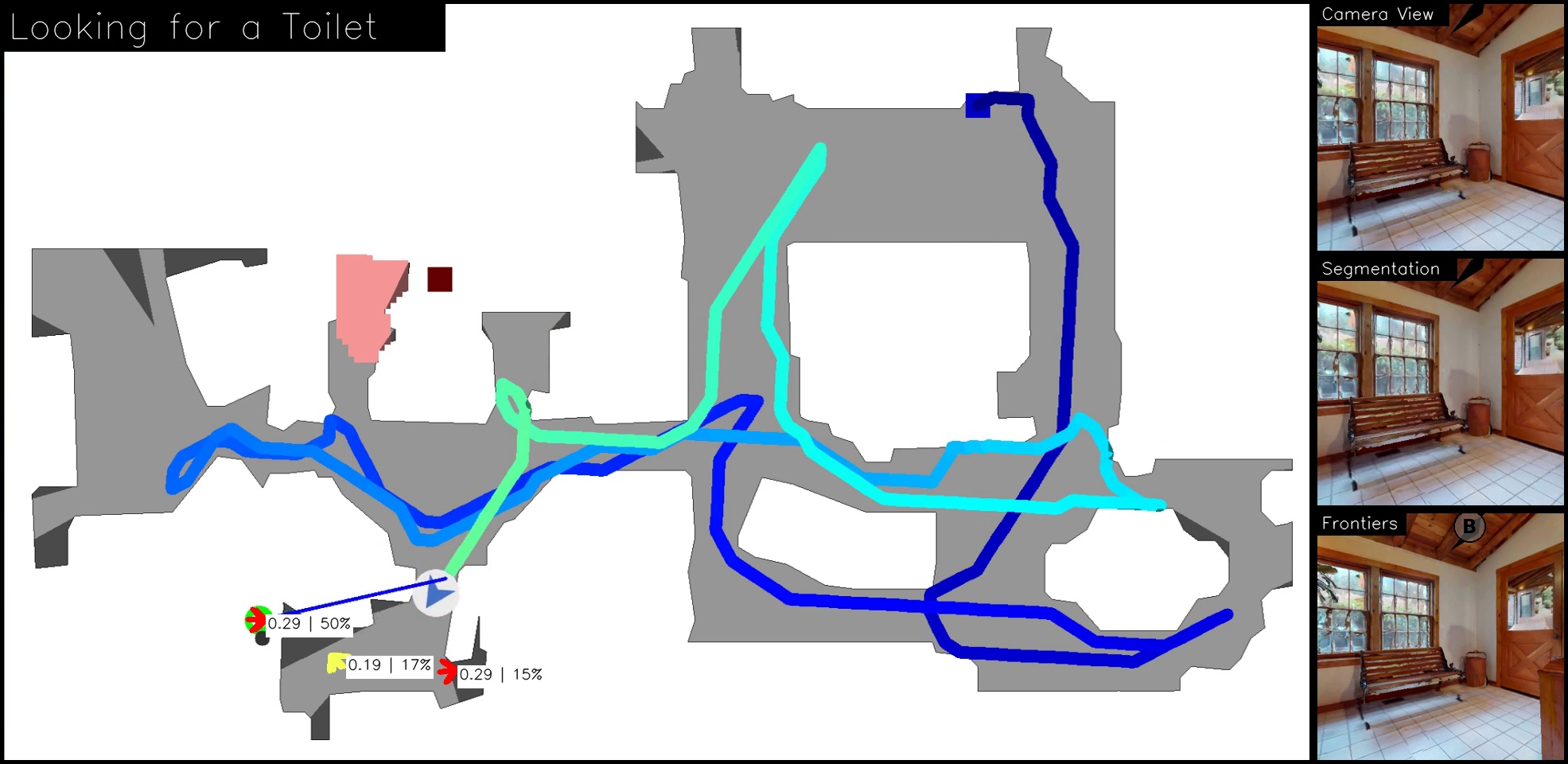}
        \caption{Scene: h1zeeAwLh9Z, Target: Toilet, Failure: Reach Max Steps}
    \end{subfigure}

    \vspace{0.6em}

\caption{\textbf{Failure Cases}. 
Each example visualizes the robot’s navigation trajectory, the final RGB observation, and the corresponding outputs of the segmentation and frontier detectors at the time the episode terminates unsuccessfully.}

    \label{fig:failure}
\end{figure*}

These failures stem from multiple sources.
Some are attributable to limitations of the Habitat benchmarks themselves.
For example, Fig. \ref{fig:failure} (a) - (c) shows a case in which the robot reaches an object that closely matches the target description but is not included in the ground-truth annotations.
Other failures are caused by limitations of the our chosen low-level PointGoal navigator (DD-PPO), such as becoming trapped in local minima or failing to escape confined regions when approaching the target, as illustrated in Fig. \ref{fig:failure} (d) - (f).
Finally, certain failures are specific to our method.
As shown in Fig. \ref{fig:failure} (g) - (i), the robot may pass near the target object but fail to register informative frontiers that would guide it closer to the goal. In some other cases, the robot detect the incorrect object as target. 

We believe that analyzing these failure cases provides valuable insights into future improvements.
In particular, more reliable frontier detection and stronger low-level navigation policies could further improve robustness.
At the same time, some cases highlight inherent ambiguities and limitations in current benchmarks,  which can affect evaluation independently of the navigation policy.

